\documentclass[a4paper,fleqn]{cas-dc}
\usepackage{caption}
\usepackage{adjustbox}

\usepackage{makecell, multirow}
\usepackage{apacite}
\usepackage[sort]{natbib}
\usepackage{filecontents}
\usepackage{lscape} 
\usepackage{soul}
\usepackage[dvipsnames]{xcolor}

\usepackage{amssymb}
\usepackage{pifont}
\usepackage[figuresright]{rotating}
\usepackage{comment}

\newcommand{\cmark}{\ding{51}}
\newcommand{\xmark}{\ding{55}}
\def\tsc#1{\csdef{#1}{\textsc{\lowercase{#1}}\xspace}}
\tsc{WGM}
\tsc{QE}
\tsc{EP}
\tsc{PMS}
\tsc{BEC}
\tsc{DE}

\defcitealias{whodash}{WHO, 2023}
\defcitealias{cdcVariant}{NCIRD, 2023}

\newcommand{\bl}[1]{#1}
\newcommand{\gl}[1]{#1}

\begin{document}
\let\WriteBookmarks\relax
\def\floatpagepagefraction{1}
\def\textpagefraction{.001}
\shorttitle{}
\shortauthors{}
\title [mode = title]{An Explainable AI Approach for Diagnosis of COVID-19 using MALDI-ToF Mass Spectrometry}                      
\author[1]{Venkata Devesh Reddy Seethi}
\ead{dseethi@niu.edu}
\author[1]{Zane LaCasse}
\ead{zlacasse1@niu.edu}
\author[1]{Prajkta Chivte}
\ead{pchivte@niu.edu}
\author[2]{Joshua Bland}
\ead{jbland3@uic.edu}
\author[2]{Shrihari S. Kadkol}
\ead{skadkol@uic.edu}
\author[1]{Elizabeth R. Gaillard}
\ead{gaillard@niu.edu}
\author[3]{Pratool Bharti}
\ead{pratool.bharti@intel.com}
\author[1]{Hamed Alhoori}
\ead{alhoori@niu.edu}
\address[a]{Northern Illinois University, 1425 W Lincoln Hwy, Dekalb, Illinois, 60115, USA}
\address[b]{University of Illinois Chicago, 840 S Wood St, Chicago, Illinois, 60612, USA}
\address[c]{Intel Corporation, 2501 NE Century Blvd, Hillsboro, Oregon, 97124, USA}

\begin{abstract}    
The severe acute respiratory syndrome coronavirus type-$2$ (SARS‐CoV‐$2$) caused a global pandemic and immensely affected the global economy. Accurate, cost-effective, and quick tests have proven substantial in identifying infected people and mitigating the spread. Recently, multiple alternative platforms for testing coronavirus disease $2019$ (COVID-$19$) have been published that show high agreement with current gold standard real-time polymerase chain reaction (RT-PCR) results. These new methods do away with nasopharyngeal (NP) swabs, eliminate the need for complicated reagents, and reduce the burden on RT-PCR test reagent supply. In the present work, we have designed an artificial intelligence-based (AI) testing method to provide confidence in the results. Current AI applications for COVID-$19$ studies often lack a biological foundation in the decision-making process, and our AI approach is one of the earliest to leverage explainable AI (X-AI) algorithms for COVID-$19$ diagnosis using mass spectrometry. Here, we have employed X-AI to explain the decision-making process on a local (per-sample) and global (all samples) basis underscored by biologically relevant features. We evaluated our technique with data extracted from human gargle samples and achieved a testing accuracy of $94.12\%$. Such techniques would strengthen the relationship between AI and clinical diagnostics by providing biomedical researchers and healthcare workers with trustworthy and, most importantly, explainable test results. 
\end{abstract}


 \maketitle

\section{INTRODUCTION}
\label{sec:introduction}
The virulent, fast-spreading, and mutating nature of SARS-CoV-$2$ caused a worldwide pandemic with more than $765$ million people infected, and $6.9$ million lives were taken in the span of $3$ years since emerging~\citepalias{whodash}. This pandemic has challenged the ability of healthcare systems to triage patients and overwhelmed capacities. Beyond healthcare, societal impacts have been apparent, and major sectors of the global economy have been affected~\citep{sohrabi2020world}. Given the severity of the pandemic and the emergence of new variants of SARS-CoV-$2$~\citepalias{cdcVariant}, curbing the infection caused by the virus has become a major concern for all nations. Several protocols have been implemented, such as vaccinations, mask mandates, and lockdowns, to reduce the spread of the virus. Frequent and rapid surveillance testing, nonetheless, remains the key critical factor in reducing the transmission of the virus~\citep{nicola2020health}.  
As COVID-$19$ exhibits a broad range of clinical manifestations, from mild fever to life-threatening conditions, it is critical to implement testing protocols for its early diagnosis in order to curb viral transmission. Molecular detection of SARS-CoV-$2$ predominantly relies on the current \bl{gold standard Real-time Polymerase Chain Reaction (RT-PCR) technique~\citep{filchakova2022review}} for viral RNA identification. Despite ongoing COVID-19 testing advancements, there is still a significant trade-off between testing specificity and complexity. Although RT-PCR protocols have achieved high specificities of $95.2\%$ 
 \citep{habibzadeh2021molecular}, complete reliance on RT-PCR exerts a strain on laboratory reagents for the testing process~\citep{carter2020assay} and the availability of skilled personnel to conduct the relatively labor-intensive process. Various other alternatives for detecting SARS-CoV-$2$ nucleic acids, such as isothermal amplification assays, amplicon-based metagenomic sequencing, hybridization using microarray and CRISPR-based technologies, are simultaneously being developed to meet the unprecedented demands for a rapid but reliable testing platform~\citep{habibzadeh2021molecular,carter2020assay,feng2020molecular,lasserre2022sars}. Approximately $400$ rapid diagnostic tests are commercially available, amongst which lateral flow devices are the most prominent ones and provide up to $70\%$ sensitivity to detect antibodies~\citep{valera2021covid}.

Matrix-Assisted Laser Desorption/Ionization Time-of-Flight (MALDI-ToF) has become one of the leading proteomic tools due to its minimal sample preparation, tolerance for matrix contaminants, and rapid analysis providing multi-dimensional information~\citep{sivanesan2022consolidating,spick2022systematic,preiano2021maldi}. Previous research reported a MALDI-ToF-based method for profiling proteins in gargle samples where both host and viral proteins are observed. This study included samples from people whose COVID-$19$ status was verified with RT-PCR testing. Utilizing straight-forward area under the curve (AUC) analysis of protein signals in the MALDI-ToF spectra, the method achieved overall above $90\%$ specificity and sensitivity (recall) rates~\citep{chivte2021maldi}. 

\begin{figure*}[ht]
	\centering
	\includegraphics[width=\linewidth]{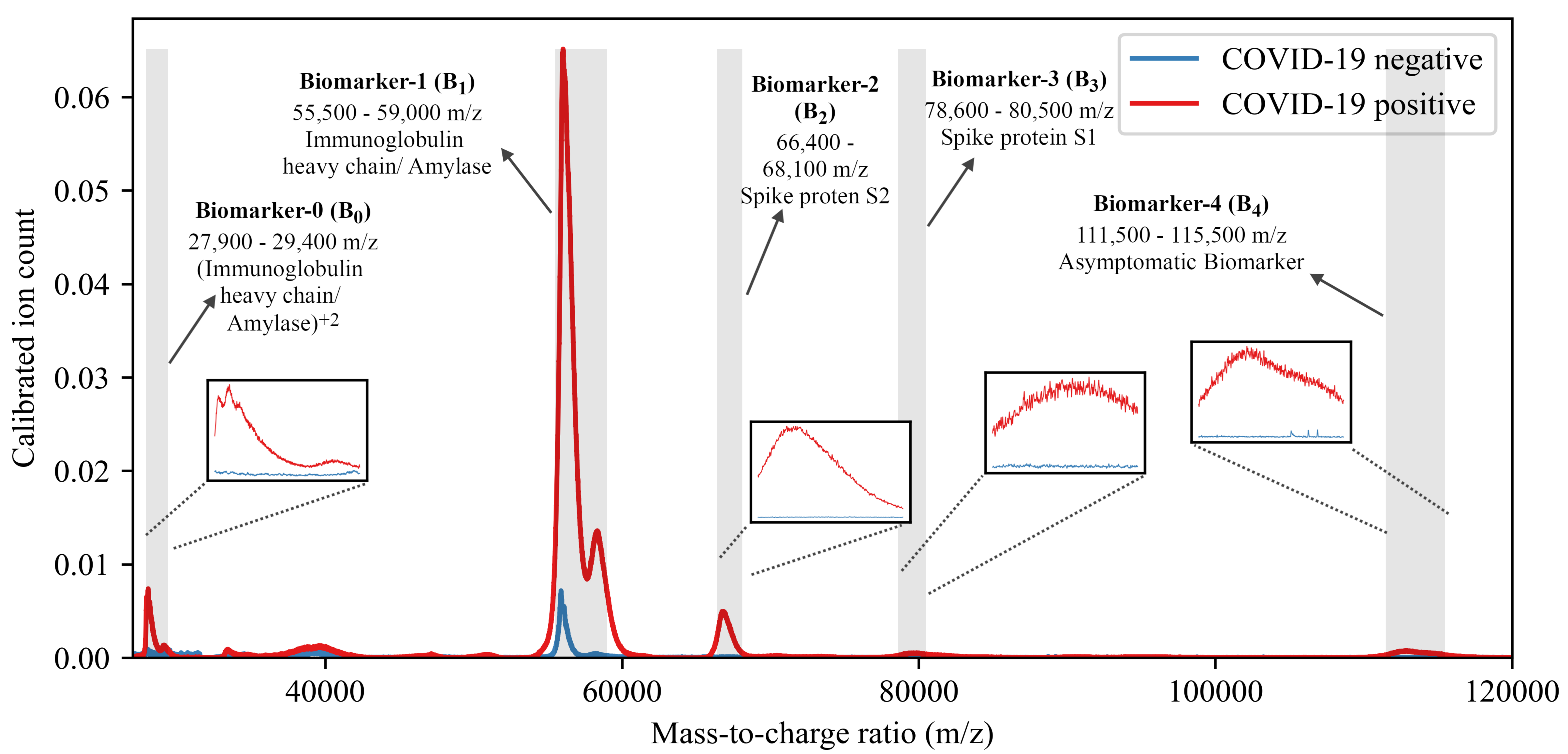}
	\caption{Comparison of COVID-$19$ negative and a COVID-$19$ positive gargle MALDI-ToF spectra after preprocessing with baseline correction and normalization. Regions of interest of different biomarkers in the spectra are highlighted with a gray hue. Zoom-in panels for each range show differences between the disease statuses.}
	\label{fig:raw}
\end{figure*}

In this study, we utilized mass spectra data for gargle samples obtained through MALDI-ToF process from~\cite{chivte2021maldi}, along with newly collected samples. An in-depth look at the spectra, as illustrated in Figure~\ref{fig:raw}, shows that they are fairly complicated, with a multitude of signals detected from both host immune response proteins and viral proteins. The amounts of these proteins change over the course of an infection's lifecycle, further confounding the data in the spectra. Thus, we now report on an alternative analysis using AI and X-AI algorithms. 

The complexity of patterns learned by AI systems often results in opaque and uninterpretable decision-making processes, creating black-box systems. In clinical settings, black-box systems pose an impossible task for domain experts and health practitioners to understand the factors contributing to a test's outcome. Therefore, we tackle this problem using X-AI techniques that document the decision-making process in transferring a mass spectral dataset into a COVID-$19$ diagnosis. Although the AI system's decision-making process is complex, the explanations are easy to understand, assisting clinicians in interpreting the rationale behind the diagnosis. This characteristic is analogous to a real-time diagnosis, where doctors explain to patients what markers lead to their final diagnosis. Therefore, ensuring model interpretability is critical, as it fosters transparency and instills trust in our AI system. 


\begin{figure*}[ht]
	\centering
	\includegraphics[width=\linewidth]{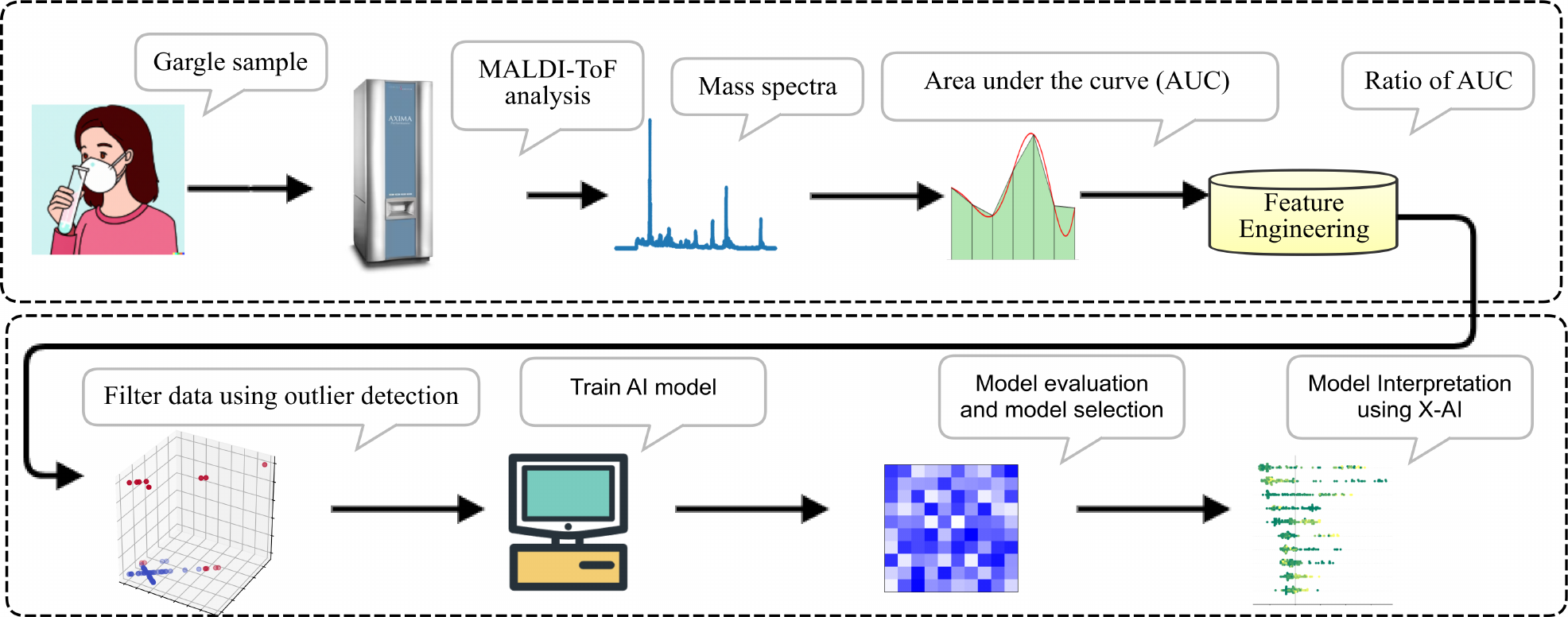}
	\caption{\gl{The experimental setup includes various stages, such as data collection, data preprocessing, data transformation, ML model training, and interpretation. Images for gargle sample adapted with~\citep{ramesh2022hierarchical} and MALDI-ToF image from~\citealp{massspec}.}}
	\label{fig:eswa_expt}
\end{figure*}

\section{LITERATURE REVIEW}
\label{sec:literaturereview}
Machine learning algorithms have become an integral part of research in COVID-$19$ diagnostic protocols~\citep{khan2021applications}. As discussed in Section~\ref{sec:introduction}, \bl{numerous protocols have investigated} different sample types, feature engineering techniques, and machine learning algorithms. In previous mass spectrometry studies that use NP swabs, \cite{nachtigall2020detection} collected $362$ samples that were analyzed by MALDI-ToF Mass Spectrometry \bl{(MS)} in the mass-to-charge range $3$ –- $15.5$ kDa. Spectral peaks were extracted by filtering local maxima, followed by correlation-based and impurity-based feature selection methods to identify 88 important peaks that discriminate between COVID-$19$ positive and negative samples. They achieved $93.9\%$ accuracy with the support vector machine (SVM) algorithms~\citep{noble2006support}. \cite{rocca2020combined} collected $311$ samples and conducted several statistical tests to select characteristic peaks, and their k-nearest neighbor classifier achieved an accuracy of approximately $90\%$. In other work, \cite{tran2021novel} collected $199$ samples, extracted peaks, and optimized their AI solution using a machine intelligence learning optimizer (MILO)~\citep{milo}. MILO builds an end-to-end pipeline that automatically selects the best features and tunes hyperparameters for algorithms~\citep{milo}. The results of MILO showed that extreme gradient boosting (XgBoost)~\citep{chen2016xgboost}, and the deep neural network (DNN)~\citep{lecun2015deep} performed best with \bl{$96.6\%$ and $98.3\%$ accuracies, respectively}. In another study, \cite{deulofeu2021detection} collected two datasets with $148$ and $66$ samples using different sample processing techniques. The authors generated $5$-$50$ principal components from the peaks dataset using principal component analysis~\citep{wold1987principal} and reported the accuracies as $70.3\%$ with XGBoost and $88.4\%$ with SVM for their two datasets, respectively.

Alternative to NP swabs, \cite{yan2021rapid} collected $298$ serum samples. In feature processing, they selected $50$ peaks where $20$, $20$, $10$ peaks were determined using LASSO~\citep{tibshirani1996regression}, partial least-squares-discriminant analysis \citep{staahle1987partial}, and recursive feature elimination with cross-validation~\citep{guyon2002gene}, respectively. As a next step, they conducted empirical analysis, picked $25$ relevant peaks from the $50$ peaks, and found that logistic regression gave the best accuracy of $99\%$. \cite{costa2022exploratory} explored the efficacy of saliva samples in COVID-$19$ testing and collected $360$ samples. They further extracted and selected nine statistically significant peaks with the Wilcoxon rank sum test and Benjamini–Hochberg correction~\citep{benjamini1995controlling}. Lastly, their SVM algorithm exhibited the best performance with accuracy ranging from $64.3\%$ -- $88.9\%$. 

\bl{While several studies have used NP-swabs, blood or blood-based specimen type coupled with X-AI to diagnose COVID-19, our study focused on saliva sampling as it is non-invasive, convenient, comfortable across all age groups, and provides an option of self-sampling. Moreover, blood is a more complex biological fluid than saliva, which increases the cost of its storage and time for processing during the detection of biomarkers \citep{esser2008sample,lasisi2019preference,abraham2012saliva}.}

Researchers have also adopted AI approaches driven by X-AI analysis~\citep{salahuddin2022transparency} in COVID-$19$ diagnosis for different datasets such as blood samples~\citep{yan2021rapid, gong2022explainable, thimoteo2022explainable,rostami2022novel, alves2021explaining}, CT scans~\citep{ye2021explainable,ullah2022explainable,pennisi2021explainable,basu2022covid}, and X-Rays~\citep{hu2022explainable, sharma2022segmentation,ozturk2020automated,ardakani2020application, wen2022acsn,vinod2021fully, li2022cov}. \cite{rahman2021multimodal} used multimodal data from X-Rays, CT scans, cough audio data, and facial recognition to create an ensemble pipeline with multiple point-of-care premise tools to diagnose COVID-$19$. Some studies also explored forecasting COVID-$19$ cases by learning the time-series relationship between historical data, such as transmission and recovery rates, to forecast case numbers or chances of getting infected with COVID-$19$~\citep{nadimi2022enhanced,leung2021explainable, nguyen2022becaked}. \cite{smith2021identifying} identified mortality factors of patients in a hospital by training a model on tabular data with features such as age, gender, and lymphocyte count and presented the features that had a high contribution towards a patient's mortality using X-AI techniques. Furthermore, explainable AI has been used in spectrometry datasets for biomarker discovery~\citep{tideman2021automated} and cell classification~\citep{xie2020single} and has also shown potential in learning age group-dependent features in COVID-$19$~\citep{liangou2021method}. 

In summary, previous X-AI studies are based on datasets that are tabular or $2$D images. Tabular data generally consist of rows and columns, while image data can be either a 2D X-ray or CT scan. By looking at previous studies in X-AI, it is apparent that SHAP~\citep{lundberg2018explainable,lundberg2020local} is the best-known method for tabular datasets and decision trees~\citep{safavian1991survey} for building transparent models with tree-based explanations~\citep{alves2021explaining}. On the other hand, for image datasets, GRAD-CAM~\citep{selvaraju2017grad} and SCORE-CAM~\citep{wang2020score} techniques are more popular. In this study, we demonstrate an X-AI framework \bl{that converts mass spectrometry data into tabular data, trains an AI model, and generates explanations using multiple X-AI algorithms, which have been proponents in making AI models study trustworthy}.

\bl{We have compiled several studies that incorporated MALDI-ToF analysis in Table~\ref{table:review} and compared the instrument, sample type, data type, sample processing time, dataset size, feature engineering technique, X-AI model, AI model, and accuracy. It is important to note that each study employs data that have different participants and different protocols. We aim to understand the machine-learning approaches used for different sample processing techniques. Other sample-related details as discussed by authors have not been discussed here \citep{spick2022systematic, sivanesan2022consolidating}.}

\section{CONTRIBUTIONS}
\label{sec:contributions}
In our current work, we incorporate one of the newest sample collection processes known for cost-effectiveness and ease of implementation. We also present a thorough evaluation of our feature processing technique as well as from the previous mass spectrometry literature. Furthermore, we demonstrate that the diagnosis becomes feasible and accurate with the integration of AI algorithms. We additionally interpret the AI algorithm to provide informed explanations of diagnosis, leverage a cost-effective and easy-to-use MALDI-ToF MS approach, and use a non-invasive specimen type (saliva). \bl{We designed our AI approach to learn the relationships between the intensity of the protein signals and the outcome of an RT-PCR test. Our experimental setup, as illustrated in Figure~\ref{fig:eswa_expt}, encompasses multiple stages: data collection, preprocessing, feature engineering, ML model training, and model interpretation. The first step in our experimental setup involves acquiring gargle samples, which are subsequently analyzed using MALDI-ToF to generate mass spectra. We then transform data by calculating the area under the curve (AUC) and employ the ratio of AUCs for feature engineering. Next, we apply anomaly detection techniques to filter the dataset to identify and filter any outliers before training the machine learning model. In the final stages, we evaluate feature importance, global and local model interpretation, and finally present an explainable AI framework that encapsulates our systematic workflow to generate the outcome along with  explanations for the outcome on unseen samples.} 

\begin{table}[h!tbp]
\centering
\begin{tabular}[h!t]{||c|c|c||}
\hline
\makecell{\textbf{Biomarker}} & \textbf{Mass Range (m/z) } & \textbf{Potential Identity}\\  \hline \hline
$B_0$ & $27900$ -- $29400$ & {\makecell{Human Protein-$1$ \\ \bl{(HP$1$)}}}\\ \hline
$B_1$ & $55500$ -- $59000$ & {\makecell{Human Protein-$2$ \\ \bl{(HP$2$)}}}\\ \hline
$B_2$ & $66400$ -- $68100$ & {\makecell{Viral Protein-$1$ \\ \bl{(VP$1$)}}} \\ \hline
$B_3$ & $78600$ -- $80500$ & {\makecell{Viral Protein-$2$ \\ \bl{(VP$2$)}}}\\ \hline
$B_4$ & $111500$ -- $115500$ & {\makecell{Human Protein-$3$ \\ \bl{(HP$3$)}}} \\ \hline
\end{tabular}
\captionof{table}{Potential biomarker ranges and their descriptions.}
\label{table:bioRanges}
\end{table} 

\section{DATASET PREPARATION}
\subsection{Data Collection}
We collected the samples from the students at Northern Illinois University in August and September $2020$. Ethical approval was issued by the Institutional Review Board. All participants provided written consent before participating in the research study. Students who consented to participate in the research study were asked to provide a water gargle sample at the same time as the NP swab sample collection. After sample collection, we processed the samples as described below while NP swabs were subjected to RT-PCR testing conducted near the University. For each gargle sample, we used the outcome of the RT-PCR test with an NP swab as ground truth in our analyses. We analyzed $152$ gargle samples, including $60$ COVID-$19$ positives and $92$ COVID-$19$ negatives.

In brief, participants gargled $10$ mL of bottled spring water for $30$ seconds and deposited the gargled sample in a $50$ mL conical centrifuge tube. We then filtered the gargled sample with a $0.45\mu$m syringe filter. Following this, we used acetone precipitation to concentrate proteins, centrifuged the sample for $30$ minutes, and resuspended the pellet in a reconstitution buffer for a further \bl{$15$ minutes}. After incubation, we spotted each sample on a target plate, air-dried for $20$ minutes, and transferred the plate to a Shimadzu AXIMA Performance mass spectrometer~\citep{massspec}, which was calibrated daily with a protein standard. On a single MALDI-ToF target plate, $394$ samples can be spotted and it takes \bl{$4$ minutes} per sample to run on the instrument. Ions were detected in a positive-ion, linear detection mode over a range of $2000$ -- $200000$ m/z. Lastly, we corrected the baseline of all the generated spectra using the Shimadzu Launchpad software obtained from the manufacturer~\citep{massspec} and normalized the ion count intensities with the parent peak of the protein standard calibrant. \bl{Overall, our process from sample collection to MALDI-ToF analysis, takes less than $1.5$ hours.}

\begin{table}[ht]
\centering
\begin{tabular}{||c|c|c||}
\hline
\textbf{Feature} & \textbf{Description} & \textbf{\bl{Potential Identity}} \\  \hline \hline
$R_0$ & $AUC_{B_0} \thickspace/\thickspace AUC_{B_1}$ & HP1/HP2\\ \hline
$R_1$ & $AUC_{B_0}\thickspace/\thickspace AUC_{B_2}$ & HP1/VP1\\ \hline
$R_2$ & $AUC_{B_0}\thickspace/\thickspace AUC_{B_3}$ & HP1/VP2\\ \hline
$R_3$ & $AUC_{B_0}\thickspace/\thickspace AUC_{B_4}$ & HP1/HP3\\ \hline
$R_4$ & $AUC_{B_1}\thickspace/\thickspace AUC_{B_2}$ & HP2/VP1\\ \hline
$R_5$ & $AUC_{B_1}\thickspace/\thickspace AUC_{B_3}$ & HP2/VP2\\ \hline
$R_6$ & $AUC_{B_1}\thickspace/\thickspace AUC_{B_4}$ & HP2/HP3\\ \hline
$R_7$ & $AUC_{B_2}\thickspace/\thickspace AUC_{B_3}$ & VP1/VP2\\ \hline
$R_8$ & $AUC_{B_2}\thickspace/\thickspace AUC_{B_4}$ & VP1/HP3\\ \hline
$R_9$ & $AUC_{B_3}\thickspace/\thickspace AUC_{B_4}$ & VP2/HP3\\ \hline
\end{tabular}
\\ \rule{0mm}{0mm}
\captionof{table}{Ratio features generated from AUC values of five potential biomarkers.}
\label{table:features} 
\end{table}

\setlength\extrarowheight{10pt}
\begin{landscape}
\begin{table}[ht!]
    \centering
\begin{adjustbox}{center}  
\begin{tabular}[h!t]{||c|c|c|c|c|c|c|c|c|c||}
\hline
\makecell{\textbf{Study}} &
\makecell{\textbf{\makecell{Instrument}}} & 
\textbf{Data Type} &
\textbf{Sample Type} &
\textbf{\makecell{Sample \\Processing \\Time }} &
\textbf{\makecell{Dataset\\ Size}} & 
\textbf{\makecell{Feature\\ Engineering}} &
\textbf{\makecell{XAI\\ Model}} &
\textbf{AI Model} &
\textbf{Accuracy} \\ \hline \hline

\cite{nachtigall2020detection} &  
MALDI-ToF &
Mass spectra & 
Nasal swabs &
---- &
$362$ &
\makecell{\makecell{Selected peaks}} &
-- &
SVM &
$93.9\%$ \\ \hline

\cite{rocca2020combined} &  
MALDI-ToF &
Mass spectra & 
Nasal swabs &
-- &
$311$ &
Selected peaks &
-- &
KNN &
$90\%$ \\ \hline

\cite{deulofeu2021detection} &  
MALDI-ToF &
Mass spectra & 
Nasal swabs &
-- &
$214$ &
\makecell{PCA on peaks} &
-- &
\makecell{XgBoost\\ SVM} &
$70.3\%$ - $88.4\%$ \\ \hline

\cite{tran2021novel} & 
MALDI-ToF & 
Mass spectra & 
Nasal swabs & 
$<1$ hour & 
$199$ & 
Selected peaks & 
-- & 
\makecell{XgBoost \\and DNN} & 
$98.3\%$ \\ \hline 

\cite{garza2021rapid} & 
\makecell{ESI-MS \\(MassSpec Pen)} & 
Mass spectra & 
Nasal swab & 
-- & 
$171$ & 
\makecell{Selected peaks} & 
-- & 
-- & 
$78.4\%$ \\ \hline 

\cite{yan2021rapid} & 
MALDI-ToF & 
Mass spectra & 
Blood serum & 
-- & 
$298$ & 
\makecell{Selected peaks} & 
-- & 
LR & 
$99\%$ \\ \hline 

\cite{chivte2021maldi} & 
MALDI-ToF  & 
Mass spectra & 
Gargle & 
$1.5$ hr & 
$60$ & 
AUC & 
-- & 
-- & 
$91.5\%$ - $96.7\%$ \\ \hline 

\cite{de2022maldi} & 
MALDI FT-ICR & 
Mass spectra & 
Saliva & 
$>5.5$ hours & 
$149$ & 
\makecell{Selected peaks} & 
-- & 
SVM & 
$86.7\%$ - $95.6\%$ \\ \hline 

\cite{thimoteo2022explainable} & 
-- & 
Clinical variables & 
Blood test & 
-- & 
$1500$ & 
\makecell{Physiological\\ measures} & 
SHAP & 
RF & 
$70\%$ - $90\%$ \\ \hline 

\cite{lazari2022maldi} & 
MALDI-ToF  & 
Mass spectra & 
Saliva & 
-- & 
232 & 
Filtered peaks & 
-- & 
SVM & 
$67.8\%$ - $88.5\%$ \\ \hline 

\cite{ullah2022explainable} & 
--  & 
\makecell{X-Ray\\CT scan} & 
Chest scan & 
-- & 
\makecell{$1646$,\\ $2481$} & 
\makecell{Contrast\\ limited adaptive\\ histogram\\ equalization} & 
GRAD-CAM & 
ensemble & 
$98.5\%$ \\ \hline 
 
\cite{xu2023improving} & 
--  & 
 CT scan & 
Chest scan & 
-- & 
$194922$ & 
\makecell{Bicubic\\ interpolation} & 
-- & 
\makecell{DenseNet121 \\with similarity \\ regularization} & 
$99.2\%$ \\ \hline 


\end{tabular}
\end{adjustbox}
\captionof{table}{Literature survey exploring various platforms coupled with AI for the diagnosis of COVID-$19$.}
\label{table:review}
\end{table}
\end{landscape}

\subsection{Potential Biomarker Selection}
\label{sec:biomarkeridentification}
The presence or absence of intensities in the collected mass spectra can be correlated to the presence or abundance of proteins within each specimen. The spectra collected represent a multitude of proteins (in this case, after acetone precipitation) that are in the range of $2,000$ -- $200,000$ m/z. With MALDI-ToF MS, we measure large molecules that have a distribution over a range of m/z values instead of a single m/z value. While we aimed to distinguish between a positive and negative sample, we selected m/z ranges identified in \cite{chivte2021maldi} that were observed to change given the COVID-$19$ status by RT-PCR. 

An overlay of mass spectra highlighting these differences for a COVID-$19$ positive and negative sample is shown in Figure~\ref{fig:raw}. Note that Figure~\ref{fig:raw} only shows a segment of the spectra where potential COVID-$19$ biomarkers are present.  These biomarker peaks were selected to potentially include both human proteins (i.e., immunoglobulin heavy chain, amylase, asymptomatic biomarker) and SARS-CoV-$2$ viral proteins (Spike protein subunits S$2$ and S$1$). For the sake of simplicity, here, these ranges were named Human Protein-$1$ (HP1), Human Protein-$2$ (HP2), Human Protein-$3$ (HP3), and Viral Protein-$1$ (VP1), Viral Protein-$2$ (VP2), respectively. We provide a summary of the biomarkers and their mass ranges in Table~\ref{table:bioRanges}.

\subsection{Feature Engineering} \label{sec:featureengineering} 
The performance of supervised machine learning (ML) algorithms depends on the quality \bl{and the dimensionality} of the features used in training. In general, an AI model becomes more robust as the number of features increases and weakens due to overfitting~\citep{mclachlan2004discriminant}. We thus conduct a holistic analysis of \bl{different} feature processing techniques such as raw spectral data, peaks in spectra, statistical features, the area under the curve (AUC), the ratio of AUC features, and statistical hand-crafted features. \bl{We evaluated only one feature processing technique at a time without combining several feature processing techniques for two reasons: to keep the dimensionality of data as small as possible and to make the feature interpretation easy to understand.} We implemented all feature engineering techniques in Python $3.8$ using libraries such as NumPy~\citep{harris2020array}, Pandas~\citep{mckinney2011pandas}, and SciPy~\citep{virtanen2020scipy}. Below, we delineate all the feature engineering techniques experimented with in our study.

\subsubsection{Raw spectral data} We had directly used the raw readings of all five chosen m/z ranges within the spectra, which resulted in $12600$ features (the combined bandwidth of all five biomarker ranges) for each sample. However, this could potentially overfit the ML models. We have also attempted binning the spectra using window sizes such as $50$ and $100$; however, we did not see any improvement in performance using binning. Therefore, we extracted features that balanced being easy to compute but also had enough information to discriminate between the positive and negative samples. 

\subsubsection{Statistical features} 
We calculated simple statistical features for each biomarker range, such as minimum, maximum, standard deviation, variance, skewness, kurtosis, and the number of peaks. Therefore, since we have five biomarkers and seven features, we generated a total of $5*7=35$ features. The statistical features extract the variables from the raw data and encapsulate the data in fewer dimensions, which reduces the burden on ML algorithms to learn patterns from the data.

\subsubsection{Area Under the Curve (AUC)}  As discussed in Section~\ref{sec:biomarkeridentification}, the presence of a viral or human protein can be measured by the intensity of the spectral peak over a narrow distribution of m/z values and cannot be localized to a single point. Hence, AUC~\citep{tallarida1987area} is more suitable over local maxima for representing an effective viral or human protein signal as it aggregates the intensities of all points in the range. Moreover, \cite{chivte2021maldi} established AUC as a reliable feature for distinguishing COVID-$19$ positive and negative samples using gargle samples analyzed via mass spectrometry. We use the biomarkers from Chivte et al. and describe them as $B_{0}$--$B_{4}$, as shown in Table~\ref{table:bioRanges}. 

\subsubsection{Ratio of the AUC values}
Although the AUC captures the intensity of a biomarker, we observed a wide variance from one sample to another. This happens mainly due to two reasons. First, each sample has potentially different amounts of human/viral protein present. Second, information regarding exposure to the virus or the time course of infection was impossible to collect. Such differences in immune response and viral load would be reflected in the biomarker intensities. We tested different arithmetic combinations such as subtraction, addition, ratio, and product for each sample with all possible combinations of different biomarker pairs. Regardless, we observed that the ratio of AUC between pairs of biomarkers showed a consistent pattern for positive and negative samples (shown in Table~\ref{table:features}).

\subsection{Data Preparation}
Initially we used stratified five fold cross validation to evaluate different machine learning algorithms for the various feature processing techniques listed above. For our final results, however, we used train-test split as we aim interpret the model using XAI techniques. partitioned $152$ samples from our MALDI-ToF dataset using a train-test-split strategy with a $70:30$ train-test-split ratio to form two disjoint and independent splits. This minimizes data leakage from classifier training into calibration and performance assessment. Within the train and test splits, we kept the ratio of the count of positive to negative samples constant to maintain class balance. Our classifier uses the train-split dataset using features in Table~\ref{table:features} (containing $41$ COVID-$19$ positive and $65$ COVID-$19$ negative) and learns the COVID-$19$ prediction task. Then, we further calibrated the classifier using the same train-split dataset. As a final step, we assessed the performance of the trained classifier on the unseen test-split dataset (containing $19$ COVID-$19$ positive and $27$ COVID-$19$ negative samples).

\section{METHODS}
\label{sec:methods} 
\begin{figure*}[ht]
	\centering
\includegraphics[width=\linewidth]{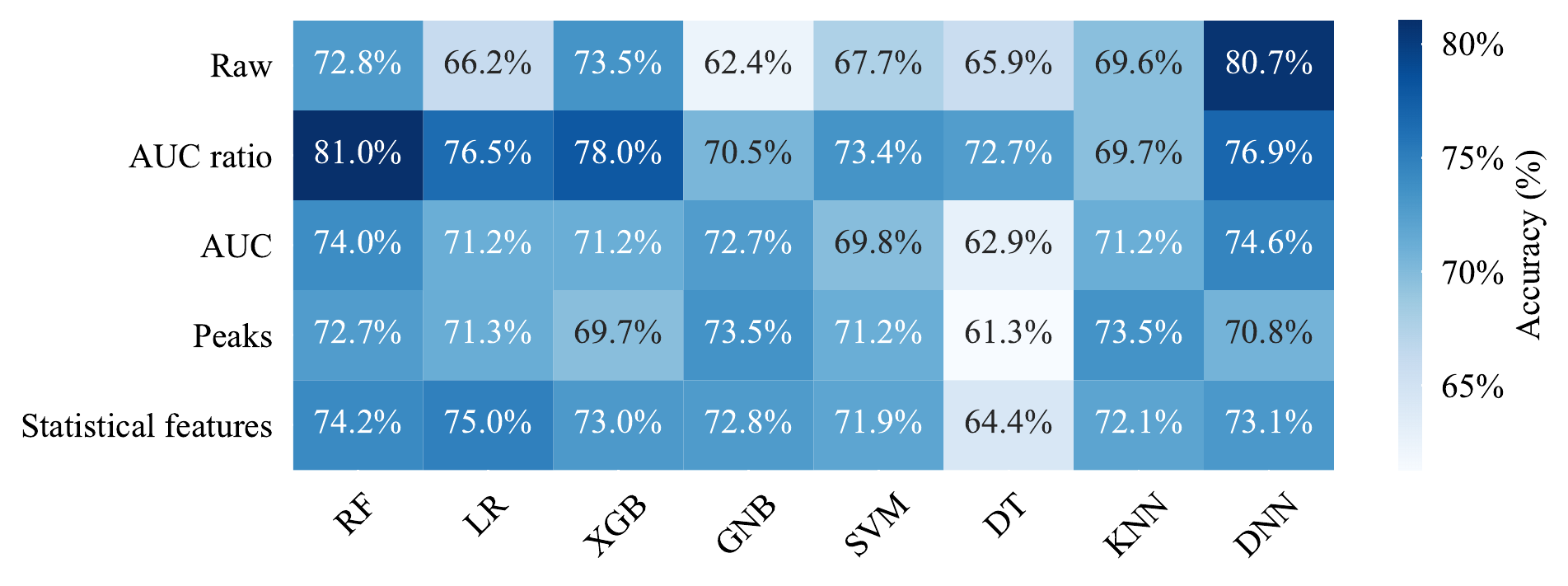}
	\caption{\gl{Five fold stratified cross-validation accuracy scores for machine learning models (shown on the x-axis) and feature engineering techniques (shown on the y-axis).}}
	\label{fig:gridsearch}
\end{figure*}

\bl{In our study, we utilized the RT-PCR outcome as the ground truth for our dataset. However, our dataset is prone to outliers for several reasons. First, the RT-PCR test results might not always be accurate. Another factor is the potential of some added noise during the sample processing stage, affecting the dataset's consistency and model performance due to outliers. To address this issue, we employed the Isolation Forest (IF)~\citep{liu2008isolation, liu2012isolation} technique to identify potential outliers. IF is an unsupervised learning algorithm that efficiently detects outliers by isolating data points in a random decision tree structure. It assigns an anomaly score to each data point based on the average path length from the root node to the terminating node, effectively distinguishing outliers from normal observations. We identified and removed $20$ outliers in our dataset using IF and then used stratified $5$-fold stratified cross-validation on various machine learning algorithms and feature engineering techniques as shown in Figure~\ref{fig:gridsearch}. Removing outliers from our dataset helped enhance the reliability of our models and ensured a more robust analysis.}

We chose the random forest (RF) classifier~\citep{breiman2001random} after assessing several machine learning algorithms \bl{such as gaussian naive bayes (GNB)~\citep{rish2001empirical}, logistic regression (LR)~\citep{hosmer2013applied}, k-nearest neighbors (KNN)~\citep{mucherino2009k}, decision tree
(DT)~\citep{safavian1991survey}, extreme gradient boosting (XGB)~\citep{chen2016xgboost}} and deep neural networks(DNN)~\citep{lecun2015deep}. \bl{Figure~\ref{fig:gridsearch} shows the accuracy of the machine learning models when tested with different feature engineering techniques. We observed that neural networks (NN) had good performance when trained on raw spectral data. However, we chose RF over NN as there is potential to overfit while using raw data of high dimensions for training on small dataset. Moreover, RF is easier to interpret, requires less training time, and is widely used in medicine-related applications~\citep{vinod2021fully, alves2021explaining,shiri2022impact}.}

 \bl{The objective of RF or any other ML} algorithm is to train a classifier function ($f$) that learns the relationships between the ratio features generated from training samples ($x_{train}$) and the outcome of the associated COVID-$19$ RT-PCR test ($y_{train}$). After training, we calibrated the RF and \bl{evaluated} on new/unseen saliva samples ($x_{test}$) to evaluate the precision, recall, \bl{and F1-score} metrics of the trained classifier's prediction ($f(x_{test})=y_{test}$) as compared to the corresponding RT-PCR result. We then leveraged X-AI algorithms to illustrate explanations for the outcomes of the RF algorithm on both a global (all-samples) and local (per-sample) basis. \bl{As a final step, we consolidated the outcomes of the RF model and explanations of X-AI algorithms into an easy-to-interpret $3$-stage framework for COVID-$19$ diagnosis.} 

 We used several Python libraries for implementing AI and X-AI algorithms. For instance,  we used Scikit-Learn~\citep{pedregosa2011scikit} for all machine learning algorithms and Tensorflow~\citep{abadi2016tensorflow} for neural networks. For X-AI algorithms, we used Scikit-learn and SHAP library~\citep{lundberg2017unified, lundberg2020local}.

\subsection{Random Forest (RF) Classifier}
\label{sec:rf}
RF is an ensemble classification algorithm that combines several tree predictors to build a forest classifier. Each tree in the forest trains from a randomly sampled vector from the input features. These random vectors are sampled independently from the same distribution in the predictor space~\citep{breiman2001random}. The outcome of a trained forest is the majority vote taken in as the mean or median of the predictions from all the independently trained trees. Additionally, in the RF algorithm, we can fine-tune the parameters such as the number of trees (also known as estimators) in the forest, methods of bootstrapping, and the properties of trees. If bootstrapping is enabled, random samples are drawn from the training set with replacements to train decision trees and take the majority vote. Furthermore, we can tune the core properties of each decision tree, such as maximum depth, the maximum number of leaves, splitting criterion, minimum samples per split, and minimum samples per leaf, to optimize the performance.  
    
Our first step in training the RF classifier was to find optimized hyperparameters by leveraging grid search with stratified $5$-fold cross-validation on the training dataset. Grid search picks hyperparameter configurations from our defined grid of hyperparameters, wherein we cataloged each configuration with its cross-validation accuracy score. Finally, the hyperparameter configuration with the best score (shown in Table~\ref{table:hyperparameters}) was selected to train the RF algorithm on complete training data, which is further evaluated and interpreted with the test-split data.

\begin{table}[h!t]
\centering
\begin{tabular}{||c|c||}
\hline
\textbf{Hyperparameter} & \textbf{Value} \\  \hline \hline
Number of estimators & $100$ \\ \hline
Bootstrap & Enabled \\ \hline
Splitting criterion & Gini impurity \\ \hline
Maximum depth of tree & $3$ \\ \hline
Minimum samples per split & $2$ \\ \hline
Minimum samples per leaf & $1$ \\ \hline
	\end{tabular}
  \\ \rule{0mm}{0mm}
\captionof{table}{Hyperparameters of our RF classifier.}
\label{table:hyperparameters}
\end{table}

\subsection{Random Forest Calibration}

 Let our trained RF model ($f$) compute the probability ($P$) for a sample ($x_i$) to test positive, then $P(f(x_i)==$ \textit{COVID-19 positive}) is computed as the fraction of trees in the forest that vote for COVID-$19$ positive. \bl{The final outcome is then predicted as COVID-$19$ positive if $P(f(x_i)>0.5$}. \bl{While the final outcome helps in making a binary decision,} the probability can help medical practitioners to understand the certainty of having the \bl{COVID-$19$ disease}. Furthermore, the binary outcome of COVID-$19$ positive or COVID-$19$ negative tests does not help diagnose early infection and recovery stages. Rather than a simple ``yes'' or ``no'' answer, knowing the probability of having COVID-$19$ is crucial as it helps to understand the intermediary stages in the disease cycle.

\begin{figure}[ht]
	\centering
	\includegraphics[width=\linewidth]{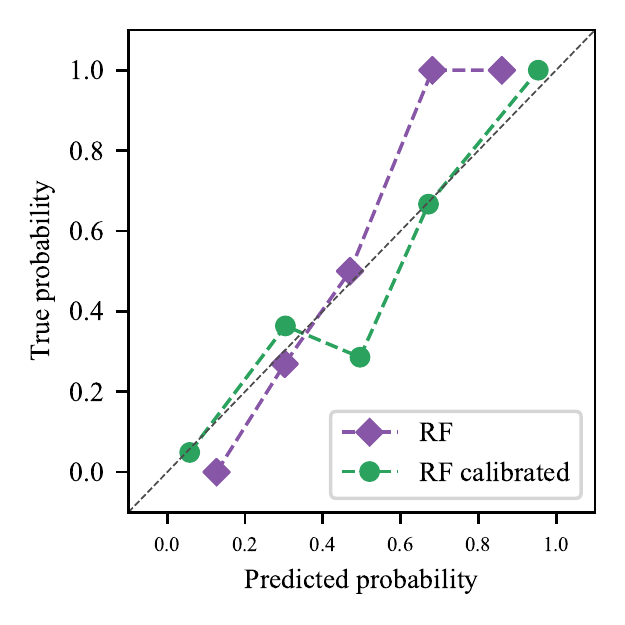}
	\caption{Calibration curves for un-calibrated and calibrated RF classifier. The gray dotted line represents an ideally calibrated classifier.}
	\label{fig:calib}
\end{figure}

Although RF achieves good performance, the predicted probabilities are sometimes out of step with their actual probabilities (i.e., the ratio of positive samples and the entire dataset). A classifier indicates ideal calibration if the predicted and actual probabilities are synchronized for the entire probability range $0-1$ (shown as a gray dotted line in Figure~\ref{fig:calib}). For example, if an ideally calibrated classifier predicts $0.8$ probability for COVID-$19$ positive, then eight of ten samples should be COVID-$19$ positive (meaning actual probability is $8/10 = 0.8$). However, in a real-life scenario, the classifiers require additional calibration to refine their probability scores.  

The most common calibration methods include isotonic regression~\citep{menon2012predicting} and Platt scaling~\citep{platt1999probabilistic}. Isotonic regression fits a step-wise regression function to the predicted probabilities and maps them to the actual probabilities, which usually works well for larger datasets. On the other hand, Platt scaling is compatible with even smaller datasets such as ours. \bl{Therefore, we used} Platt scaling \bl{that} adds a one-dimensional logistic regression~\citep{hosmer2013applied} function to the output of RF that calibrates RF's predicted probabilities with a sigmoid function~\citep{goodfellow2016deep}.  

\begin{figure*}[ht]
	\centering
    \includegraphics[width=0.475\linewidth, keepaspectratio]{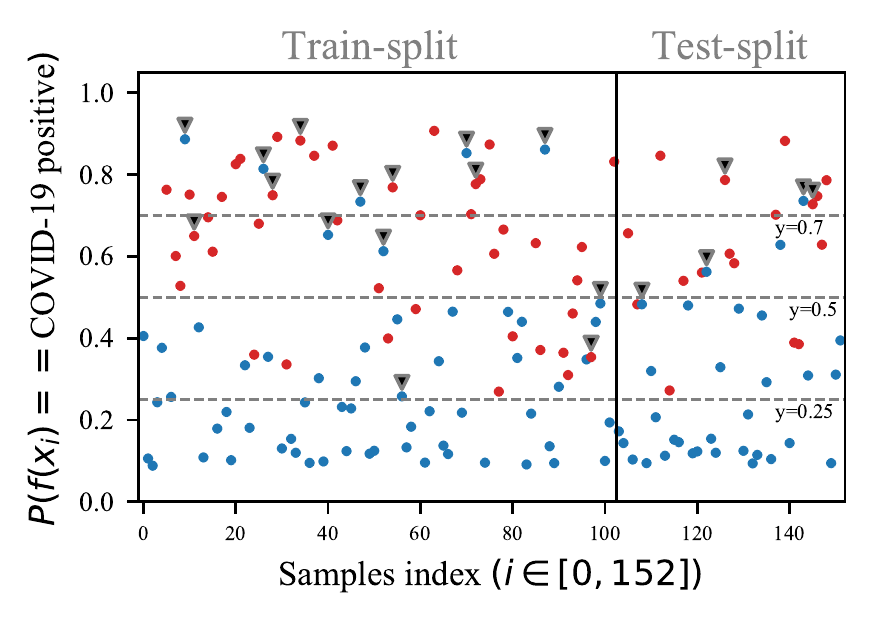}
\includegraphics[width=0.475\linewidth, keepaspectratio]{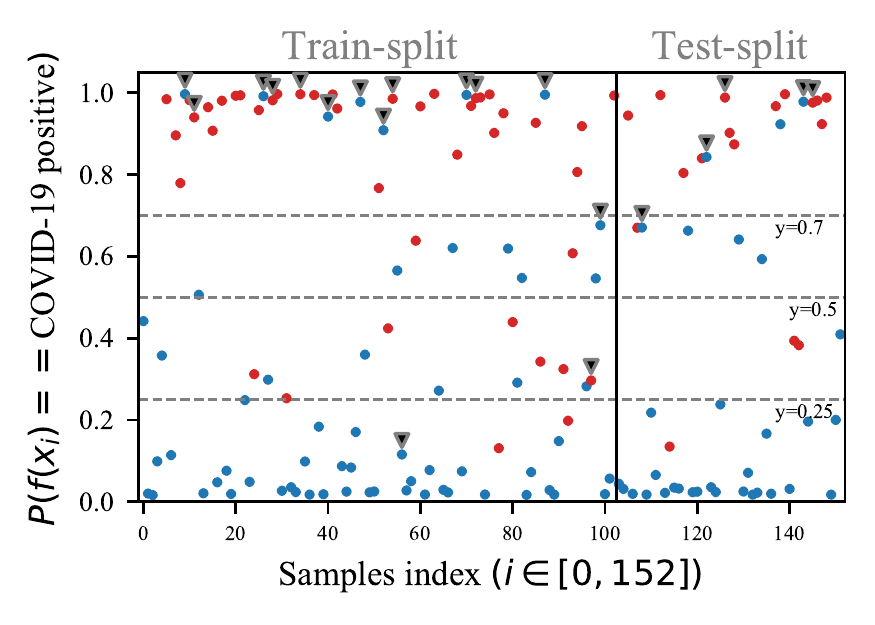}
    \caption{Probability of each sample (on the x-axis) to test COVID-$19$ positive (on the y-axis) for un-calibrated RF (on the left) and calibrated RF (on the right). The true labels for each sample are color-coded in blue and red for COVID-$19$ negative and positive tests, respectively. The grey triangles show the anomalies identified by the isolation forest.}
    \label{fig:rfcalibration}
\end{figure*}

\begin{table}
\begin{center}
\begin{tabular}{||c|c|c|c||}
\hline
\makecell{$\mathbf{P(f(x_{i})==}$ \textbf{COVID-19}\\ \textbf{positive)} $\mathbf{(P)}$}  & \textbf{Positive} & \textbf{Negative} & \textbf{Total} \\  \hline \hline
$1\geq P>0.7$ & $38$ & $1$ & $39$  \\ \hline
$0.7\geq P>0.5$  & $3$ & $9$ & $12$ \\ \hline
$0.5\geq P>0.25$  & $8$  & $8$ & $16$ \\ \hline
$0.25\geq P\geq0$  & $3$  & $62$ & $65$ \\ \hline
	\end{tabular}
  \\ \rule{0mm}{0mm}
\end{center}
\captionof{table}{Samples in each region segregated with thresholds on probability \bl{after removing outliers} (for calibrated classifier).}
\label{table:probabilitiescal}
\end{table}

After calibrating RF, we investigated the calibration quality by plotting a calibration curve (as shown in Figure~\ref{fig:calib}) using the train-split dataset. The calibration curve has predicted probabilities (on the x-axis) and actual probabilities (on the y-axis). Note that the probabilities were binned into \bl{five} bins to make the curve less noisy and more readable. We observed that the probabilities were well-calibrated for calibrated RF compared to the original RF. We also graphed the predicted probabilities for each sample in the train-split and the test-split datasets in Figure~\ref{fig:rfcalibration} to get a more in-depth view of the probability scores for each sample. In this plot, all the positive samples are colored red and negative samples with blue. All samples that were identified as anomalies/outlier by IF are marked with a grey triangle on top. 
According to the train-split dataset's probability scores, we found that setting a gatekeeper threshold for probability resulted in better performance. Hence, based on the train-split dataset, we set the threshold to predict COVID-$19$ positive when probability (P) is $1\geq P>0.7$ and $0.25 \geq P\geq0$ for COVID-$19$ negative. \bl{After setting the probability thresholds, we observed lower mispredictions and improved accuracy on the train-split dataset.} Therefore, we used these thresholds on both train-split and test-split datasets and tabulated the count of positives and negatives in each threshold in Table~\ref{table:probabilitiescal}. Using the thresholds on calibrated RF, overall, all samples had no false negatives and only two false positives. The false positives could be an experimental error in sample preparation or from the PCR analysis ground truth.  

\subsection{Random Forest Interpretation Using Explainable AI}
\label{sec:interpret}
To instill trust in the model's predictions, we interpret the RF model both globally and locally using X-AI techniques. While global interpretation explains the importance of each feature in overall model prediction, local interpretation provides reasoning on a single instance level. For global interpretation, analyzing the importance of features determined by a trained model gives information about its prediction process. In this section, we leverage multiple techniques to investigate the feature importance of our trained RF model.

\subsubsection{Impurity-based feature importance (IFI)}
\label{sec:ifi} 
First, we interpreted the trained RF classifier with a model-specific IFI technique. This technique computes the feature importance as the average decrease in Gini impurity~\citep{ceriani2012origins} from a feature node to its child nodes. Generally, except for the leaf nodes at the bottom of the tree, all other nodes have a threshold condition placed on a feature whose decision splits the node into two child nodes. The drop in impurity in this split is then calculated as the difference of Gini impurity between the node and the sum of its child nodes, weighted by the number of samples in each node. The final feature importance is the mean of drop in impurity at every prevailing feature node across the forest. Following these procedures, we calculated the IFI for all $10$ features as shown in panel A of Figure~\ref{fig:featureimportance} with the drop in Gini impurity on the y-axis and features on the x-axis. Although this method gives a general understanding of feature rankings, one disadvantage is the inflation of importance for features with high cardinality in the forest. Therefore, we investigated other feature-importance algorithms to validate the results.

\subsubsection{Permutation feature importance (PFI)}
\label{sec:pfi} 
The PFI technique~\citep{breiman2001random} 
calculates the drop in performance when the trained RF algorithm is validated with permuted feature values which essentially makes the feature a noise. This method analyzes one feature at a time by permuting their values to break the association between the feature and the outcome. Clearly, the permutation of important features results in a larger performance drop. Therefore, the feature importance of using PFI is the drop in permuted feature accuracy from the base accuracy (where every feature was used in its original form). Bias from random value assignment to feature values was prevented by permuting values and averaging the drop in accuracy across $1000$ iterations. However, this algorithm fails to capture feature interaction effects as it permutes only one feature at a time. Panel B in Figure~\ref{fig:featureimportance} shows the permutation importance scores for the RF classifier with overall feature importance printed on the y-axis and the features on the x-axis. 

\subsubsection{Shapely additive explanations (SHAP)}
\label{sec:shap}
\cite{lundberg2017unified} formalized the SHAP algorithm as an additive feature attribution technique that unifies concepts from local model-agnostic explanations (LIME)~\citep{ribeiro2016should} and Shapely values~\citep{shapley1953value}. In general, linear models are easy to interpret as the feature importance of the model is reflected by the model's coefficients. However, due to the complexity of the RF algorithm, it cannot be interpreted as a simple linear model. The output of a single sample from RF, on the other hand, could be explained by producing some neighboring samples and then fitting a linear model locally on the newly obtained neighborhood samples. The LIME algorithm follows this approach: it generates a local point of view per sample by applying perturbations to its feature values and generating some artificial data in the neighborhood of a sample. LIME then fits a surrogate model on the artificial data and presents the coefficients of the surrogate model as the model interpretation. The downside of LIME is that it generates artificial samples using a heuristic kernel function. However, artificially replicating clinical data is difficult, and the dependence on kernel function heuristics to generate artificial samples is unreliable. SHAP adopts LIME procedures and bridges its limitations with the Shapely values, which borrow principles from cooperative game theory. Shapely values rationalize the distribution of credit or payoff among the players in a cooperative game by inspecting each player's contribution individually and collaboratively as a team~\citep{shapley1953value}.

In theory, the feature attributions assigned by our trained RF classifier ($f$) to the features $R_i$ (where $i\in(0,9)$ for $10$ ratio features) in a given data sample ($x$) is computed as ($\phi_{R_i}(f,x)$) using Equation~\eqref{eq:shaplyeq}. This equation facilitates SHAP to compute a unique optimal solution for feature attribution by combining individual as well as team efforts. This can be understood in light of PFI, where we extracted feature attributions by masking each feature at a time and computing the drop in performance of the classifier. PFI captures individual contributions but fails to grasp collaborative contributions. The more intuitive approach of SHAP allows for collaborative contributions in conjunction with individual contributions to be considered by iterating over all possible subsets $S$ in our feature space having $F$ number of features excluding feature $R_i$ (where $F=10$) as $S \subseteq F /\{R_i\}$. In each subset, the marginal contribution ($C$) of feature $R_i$ is extracted as the difference in the outcome of the function $f$ with and without $R_i$ in the subset as shown in Equation~\eqref{eq:shaplyCont} where $f_{S\cup\{R_i\}}$ and $f_{S}$ are two retrained functions with and without $R_i$ respectively that are marginalized over features absent in the subset. The resultant sum of $C$ is weighted with $W$ (refer to Equation~\eqref{eq:shaplyWeight}) whose numerator is the product of the different number of ways a subset can be formed ($S!$) and the number of ways to choose features excluding the features in a subset ($(|F|-|S|-1)!$). The numerator is normalized by $F!$, the total possible ways of choosing the features. Finally, for each feature in $x$, Equation~\ref{eq:shaplyeq} computes the SHAP values through iterations over all possible subsets $S$. The magnitude of final SHAP values shows the contribution of a feature to the outcome of the classifier. In this sense, the positive or negative SHAP value indicates the feature's support towards the COVID-$19$ positive or negative decision respectively. 

\begin{equation}
C = f_{S\cup\{R_{i}\}}(x_{S\cup\{R_{i}\}})-f_{S}(x_{S})
\label{eq:shaplyCont}
\end{equation}

\begin{equation}
 W = \frac{|S|!(|F|-|S|-1)!}{|F|!}
\label{eq:shaplyWeight}
\end{equation}

\begin{equation}
\phi_{R_{i}}(f,x)=\sum_{S \subseteq F /\{R_{i}\}} W \times C
\label{eq:shaplyeq}
\end{equation}
Since the idea behind Shapely values is that there exists only one solution to allocate the attributions, the results are more reliable than previous methods such as LIME~\citep{ribeiro2016should}. In addition to this, SHAP has other desirable axiomatic properties. The local accuracy property entails that the sum of all feature attributions of a model should approximate the original model meaning $f(x)=\phi_{R_n}+\sum_{i=0}^{9}\phi_{R_i}$, where $R_n$ is the feature attribution when all features are toggled off. The missingness property ensures that the features toggled off do not impact the model's output. Lastly, the consistency property ensures for any feature $R_i$, $\phi_{R_i}$ remains consistent with respect to the impact of feature values of any retrained model. For example, in two functions $(f_1, f_2)$ trained on different subsets including the feature $R_i$, if the feature values of $R_i$ have larger impact on $f1$ than $f2$, then $\phi_{R_i}(f1, x) \geq \phi_{R_i} (f2, x)$. 

 \begin{figure}[h!t]
	\centering
	 \includegraphics[width=\linewidth]{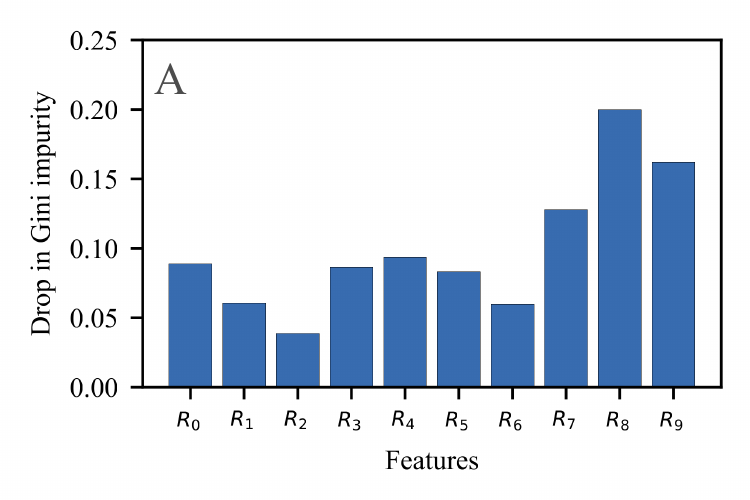}
    \includegraphics[width=\linewidth]{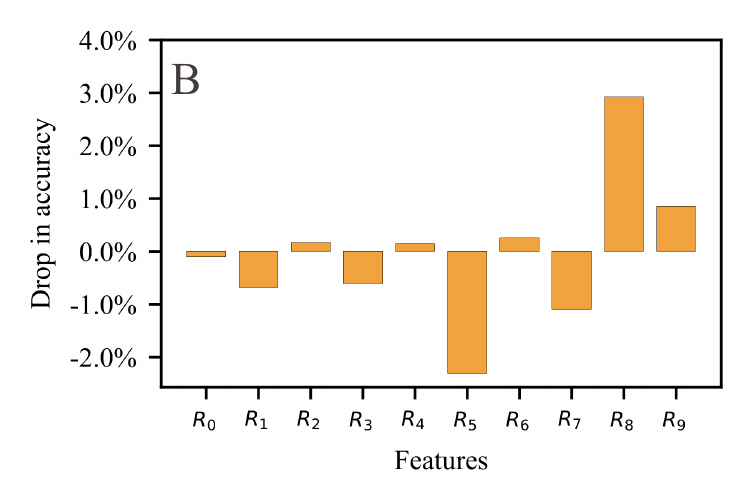}
\includegraphics[width=\linewidth,keepaspectratio]{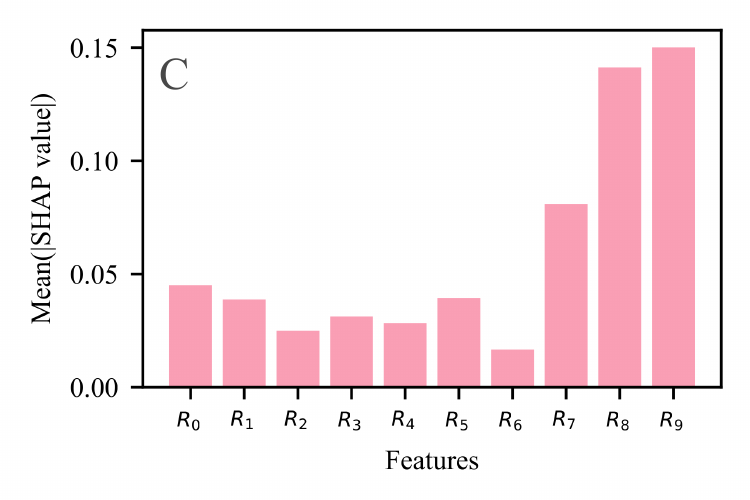}\\

    \caption{Feature importance for RF classifier using three feature interpretation techniques: Impurity-based feature importance (panel A, top), Permutation Feature Importance (panel B, middle), and Shapely Additive Explanations (panel C, bottom). The y-axis shows the feature importance for each feature shown on the x-axis.}
	\label{fig:featureimportance}
\end{figure}

 We then combine Equations~\eqref{eq:shaplyCont}, \eqref{eq:shaplyWeight}, and \eqref{eq:shaplyeq} and compute SHAP values for all $132$ data points excluding $20$ anomalies where each data point has having $10$ ratio features. The result is a SHAP value matrix of dimensions $[132\times 10]$ that will aid in the local and global model interpretation of our trained RF model. For global model interpretation, we extract the magnitude of feature importances in two steps: $1$) take the absolute SHAP values, which nullify the effect of direction, and $2$) compute the mean of absolute SHAP values across all samples. This results in condensing the $[132 \times 10]$ matrix to a single array of $10$ feature importances (as shown in panel C of Figure~\ref{fig:featureimportance}). \bl{By comparing all $3$ feature importance algorithms in Table~\ref{table:globalFI}, it is conclusive that the two most important features are $R_8$ and $R_9$ while there was some variation among the least important features.}
 In addition to global (all-sample) explanations, we present local (per-sample) explanations using the SHAP value matrix as shown in Figure~\ref{fig:shaplocal}. The x-axis represents the SHAP value per sample, and features are numbered on the y-axis. We encoded the feature values in color ranging from blue (signifying small value) to red (signifying large value) and assigned each feature to a quartile based on its value, as shown at the origin of each bar. The magnitude of each bar explains the impact of each feature on the model's output and thus allowing us to interpret them as feature importance. We provide more details on global and local explanations in sections~\ref{sec:globalinterpretation} and~\ref{sec:localinterpretation}, respectively.

\begin{table}[htbp]
 \centering
\begin{tabular}{||l|c|c|c|c||}
    \hline
    \textbf{Model} & \textbf{Accuracy} & \textbf{Precision} & \textbf{Recall} \\ \hline \hline
    RF & $84.78\%$ & $0.83$ & $0.84$ \\ \hline 
    RF + AD & $87.8\%$ & $0.84$ & $0.89$ \\ \hline 
    \makecell{Cal RF +\\AD + PT} & $94.12\%$ & $0.93$ & $0.93$ \\ \hline 
\end{tabular}
\\
AD: Anomaly detection, PT: Probability threshold \\
Cal RF: Calibrated random forest
\captionof{table}{Ablation experiment: add each component of our AI framework and observe their influence on the train-test split performance.}
\label{table:ablation}
\end{table}

\section{RESULTS}
\subsection{Results of Random Forest Classifier}
\label{sec:rfresults}
We trained the RF algorithm with the hyperparameters extracted through an exhaustive grid search, shown in Table \ref{table:hyperparameters}. Then, we \bl{removed outliers} and employed a total of $91$ samples in training and $41$ in testing the model. We then calibrated the trained RF algorithm using Platt scaling and set a threshold on predicted probability for COVID-$19$ diagnosis at $0.7$ and $0.25$ for COVID-$19$ positive and COVID-$19$ negative diagnosis, respectively. The rationale for setting a threshold on probability is that samples on the border of diagnosis (i.e., probability$==0.5$) may require special attention or re-testing. All in all, $28$ of the samples had low confidence, and the remaining $104$ had high confidence in probability. \bl{We conducted an ablation experiment by adding each component (i.e., anomaly detection, calibration, and probability threshold) to the baseline RF model. The results are shown in Table~\ref{table:ablation}. First, we observed that the baseline RF model trained and tested with the train-test split on the whole dataset had accuracy, precision, and recall as $84.78\%$, $0.93$, and $0.93$, respectively. Next, adding anomaly detection and filtering $20$ anomalous data points, we saw a slight improvement in performance with accuracy, precision, and recall to $87.8\%$, $0.84$, and $0.89$, respectively. Lastly, after calibrating the RF model and setting the probability threshold on the clean data without anomalies, the accuracy improved significantly to $94.12\%$ and precision and recall both to $0.93$. Because of this improved performance, we used the calibrated RF with anomaly detection and probability thresholds in our explainable-AI framework.}

\begin{figure}[h!t]
	\centering
\includegraphics[width=0.48\linewidth, keepaspectratio]{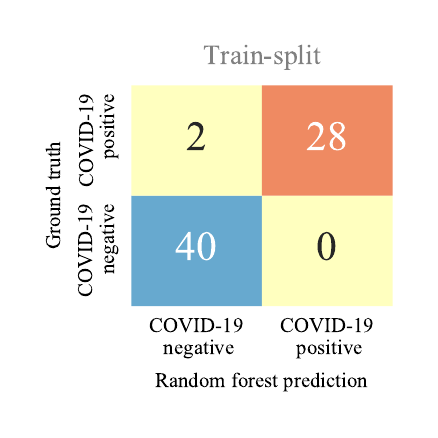}
\includegraphics[width=0.48\linewidth, keepaspectratio]{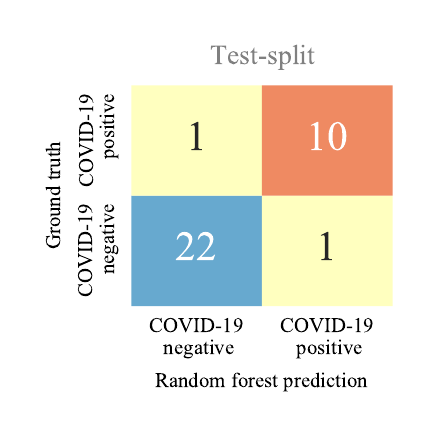}
\includegraphics[width=\linewidth, keepaspectratio]{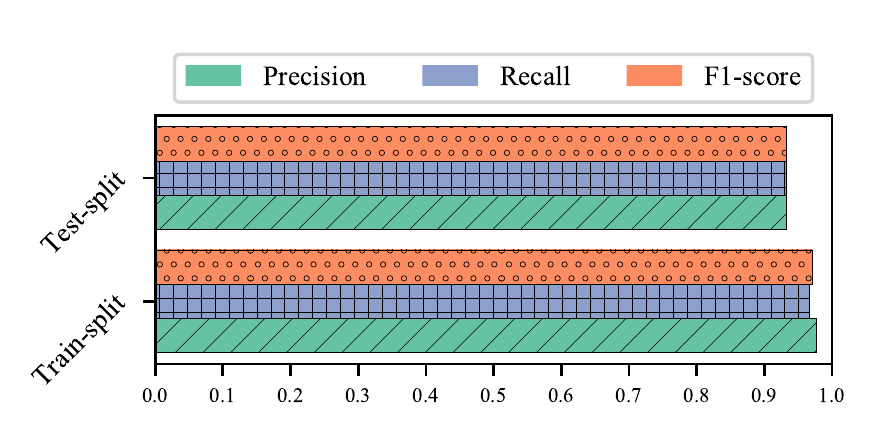}\\
    \caption{\bl{Confusion matrices for the RF classifier on the train-split and test-split datasets (on top). Precision, recall, and F$1$-scores for train-, and test-splits for the RF classifier (on bottom).}}
	\label{fig:rfreport}
\end{figure}

Since our dataset had more negative samples than positive samples, it was crucial to give equal weight to model metrics for both positive and negative categories. Therefore, we evaluated precision, recall, and F$1$ score metrics on both the test and train datasets \bl{and reported the macro average of the metrics on positive and negative COVID-$19$ samples}. As shown in the evaluation report in Figure~\ref{fig:rfreport}, the train-split had a high precision, recall, and F$1$-score of \bl{$0.98$}, \bl{$0.97$}, $0.97$, respectively. On the other hand, precision, recall, and F$1$-scores were \bl{$0.93$}, \bl{$0.93$}, and $0.93$, respectively, for the test-split. There was $1$ false positive, $1$ false negative, and $32$ true predictions for the test-split. In the train-split, there were $2$ false negatives and a total of $68$ correct predictions. Although these metrics portray the high efficiency of the trained model in COVID-$19$ testing, it still does not provide a means to understand its prediction mechanisms. This motivated us to further dissect the trained RF model and understand its prediction mechanisms using X-AI techniques.

\begin{table*}[htbp]
 \centering
\begin{tabular}{||c|c||}
\hline
Method & Order of feature importances \\  \hline \hline
Impurity-based feature importance & $R_8 > R_9 > R_7 > R_4 > R_0 > R_3 > R_5 > R_1 > R_6 > R_2$ \\ \hline
Permutation feature importance &  $R_8 > R_9 > R_6 > R_4 > R_2 > R_0 > R_3 > R_7 > R_1 > R_5$ \\ \hline
Shapely additive explanations & $R_9 > R_8 > R_7 > R_0 > R_5 > R_1 > R_3 > R_4 > R_2 > R_6$ \\ \hline
\end{tabular}
\\ \rule{0mm}{0mm}
\captionof{table}{Global feature importances shown in descending order for different feature interpretation techniques.}
\label{table:globalFI}
\end{table*}

\subsection{Model Interpretation of Random Forest Classifier}
The outcome of the RF model is more complex than a simple decision tree. In decision trees, we have a single decision path for each prediction, making the model self-interpretable. However, RF coalesces the outcome of many unique decision trees, meaning for each sample, there are a multitude of decision paths; one for each tree traversal which makes model interpretation complex. Therefore, we simplify the interpretation of RF by presenting the rationale behind predictions in the form of global (all sample basis) and local (per sample basis) interpretations. These interpretations help domain experts and clinicians handling COVID-$19$ testing understand the classifier's operations and gain rich diagnostic information. First, the idea of the global interpretation of a model goes beyond the analysis of feature importance. It can elaborate on how the distribution of each feature affects the model prediction. Second, local interpretation operates on one sample at a time and delineates each feature's value and contribution to the outcome of RF. 

\subsubsection{Global interpretation}
\label{sec:globalinterpretation}

While computing feature importance with different algorithms, we found similar feature ranking: $R_9$ and $R_8$ were consistent among the two features (as shown in Table~\ref{table:globalFI}). It is fairly common to observe slight differences in the feature importance across the three algorithms due to underlying procedural differences and the shortcomings of techniques highlighted in their respective sections. However, on the whole, these rankings reveal that most feature importances assigned by the three algorithms are fairly consistent. The consistent nature of feature attribution shows that our RF algorithm can be understood from different perspectives of the X-AI algorithms. Although these are important findings, they do not explain how the model prediction correlates with the feature values. While IFI and PFI algorithms can generate feature rankings, they do not provide enough information to explore the interactions between features. Therefore, we extended our X-AI approach to the SHAP algorithm, which can analyze feature interactions, data distribution, and the model's dependency on it.

\begin{figure}[ht]
\centering
\includegraphics[width=\linewidth,keepaspectratio]{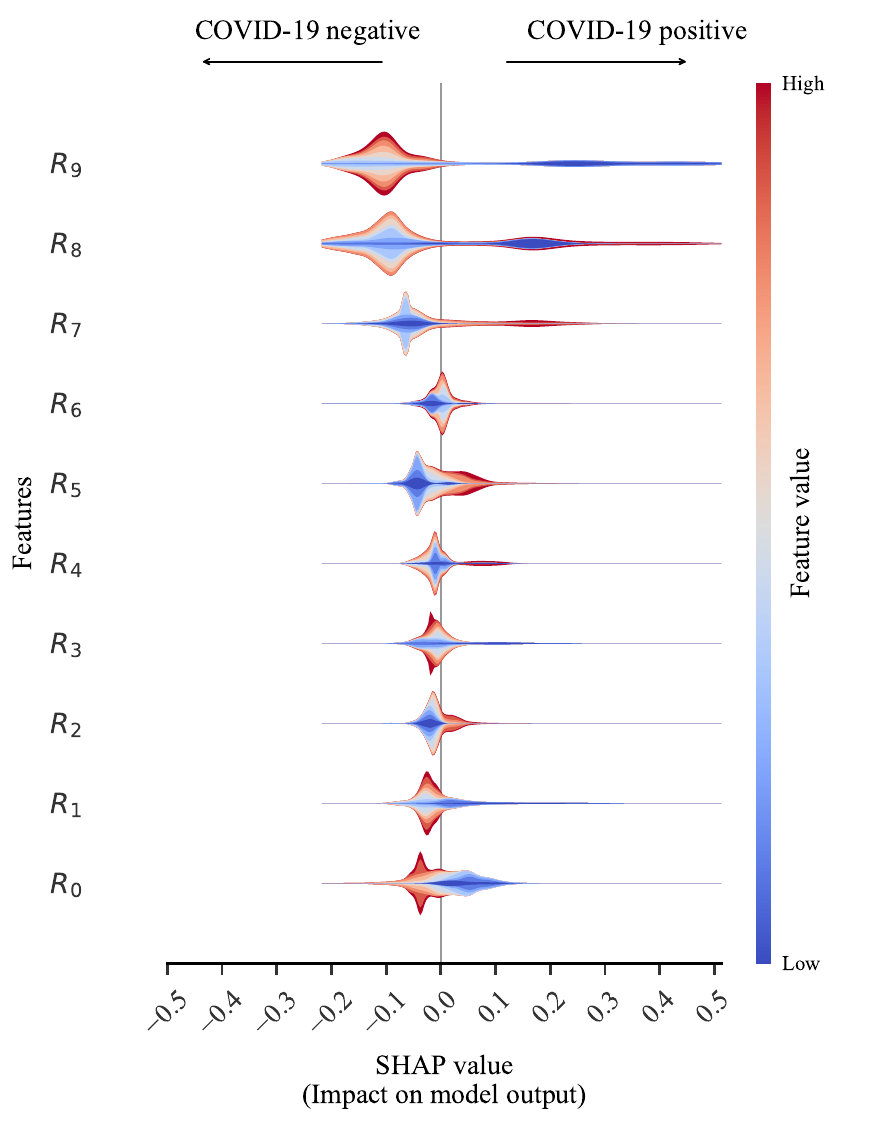}\\
\caption{Violin plot for global interpretation of calibrated RF classifier using SHAP values.}
\label{fig:shap} 
\end{figure}

We utilized the SHAP algorithm to produce a violin plot, which provides a fine-grained view of SHAP values (feature importance) and feature value distribution as shown in Figure~\ref{fig:shap}. This plot illustrates the trends of SHAP values in conjunction with feature values by plotting the SHAP value matrix (as described in section~\ref{sec:shap}) without aggregation across all test-split samples. In addition, the plot encodes the density of the samples with a certain SHAP value as the width of the line. To interpret these plots, the higher the SHAP value, the greater the importance; the positive SHAP values support positive COVID-$19$ prediction, negative values support negative output, and the feature values are represented on a blue-red color-bar where deep blue and deep red correspond to low and high feature values. For example, the feature $R_9$ has SHAP values extending up to $+0.495$ towards a positive COVID-$19$ decision, where the feature value is low. For a negative test, mostly high values of $R_9$ contributed towards a negative COVID-$19$ prediction. On the contrary, for feature $R_5$, lower and higher feature values contribute towards COVID-$19$ negative and positive tests, respectively. Features $R_6$ and $R_7$ follow similar patterns as $R_5$ while $R_0$ and $R_1$ have similar patterns of $R_9$. 

\begin{figure*}[ht]
\centering
\includegraphics[width=0.31\linewidth,keepaspectratio]{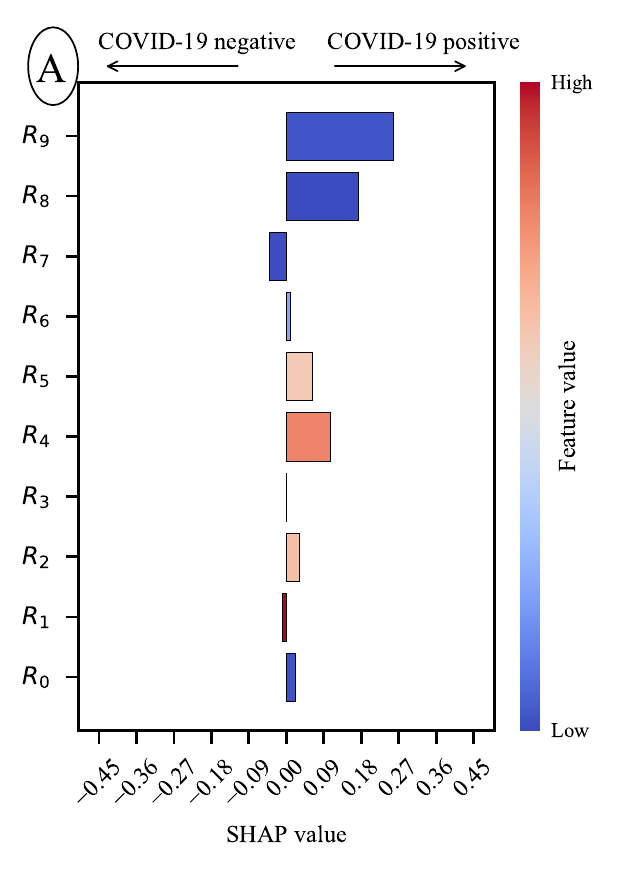}
\includegraphics[width=0.31\linewidth,keepaspectratio]{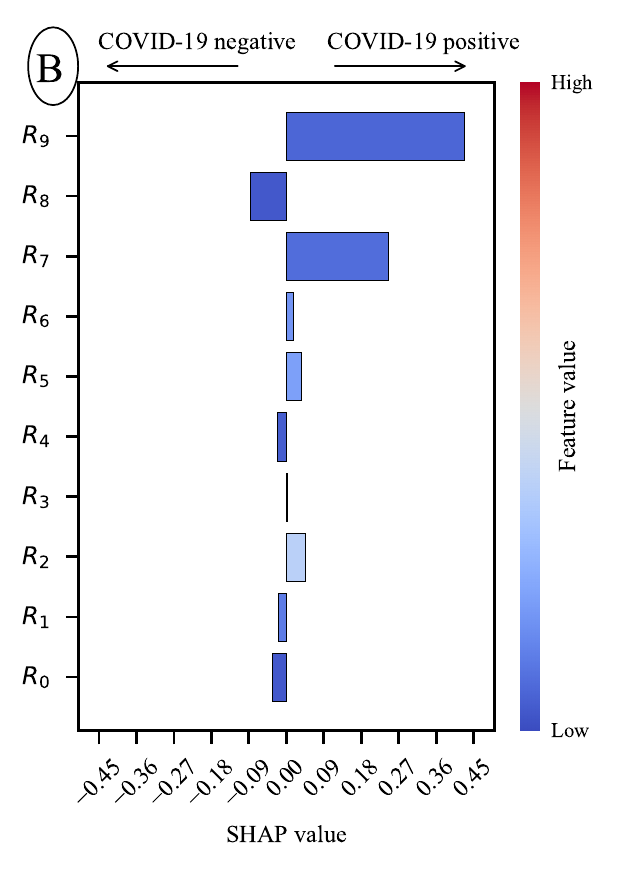}
\includegraphics[width=0.31\linewidth,keepaspectratio]{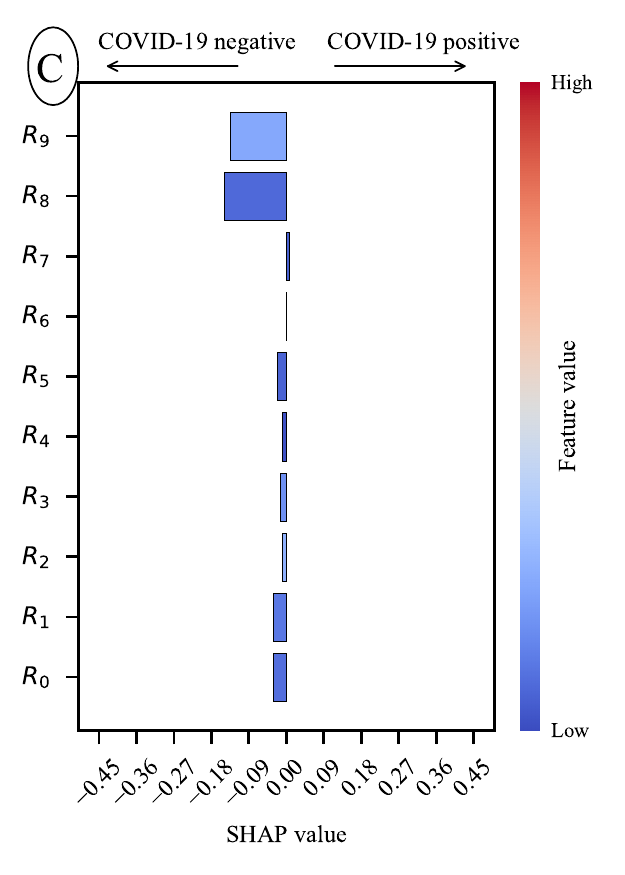}\\
\includegraphics[width=0.31\linewidth,keepaspectratio]{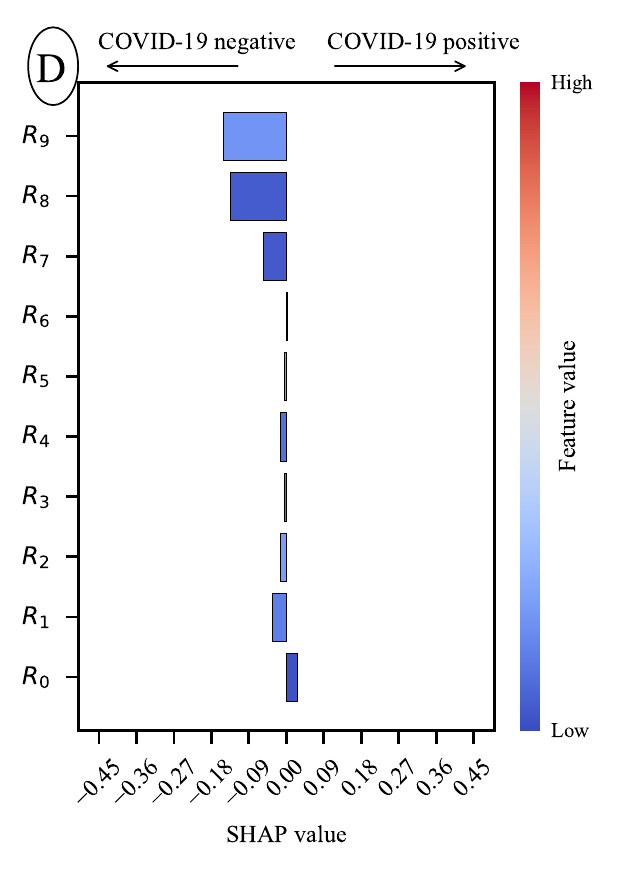}
\includegraphics[width=0.31\linewidth,keepaspectratio]{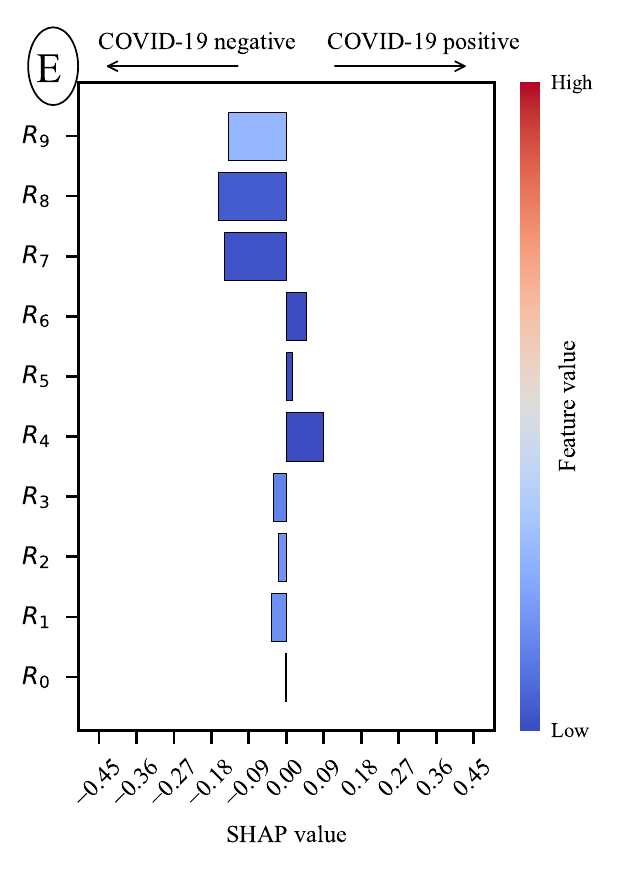} 
\includegraphics[width=0.31\linewidth,keepaspectratio]{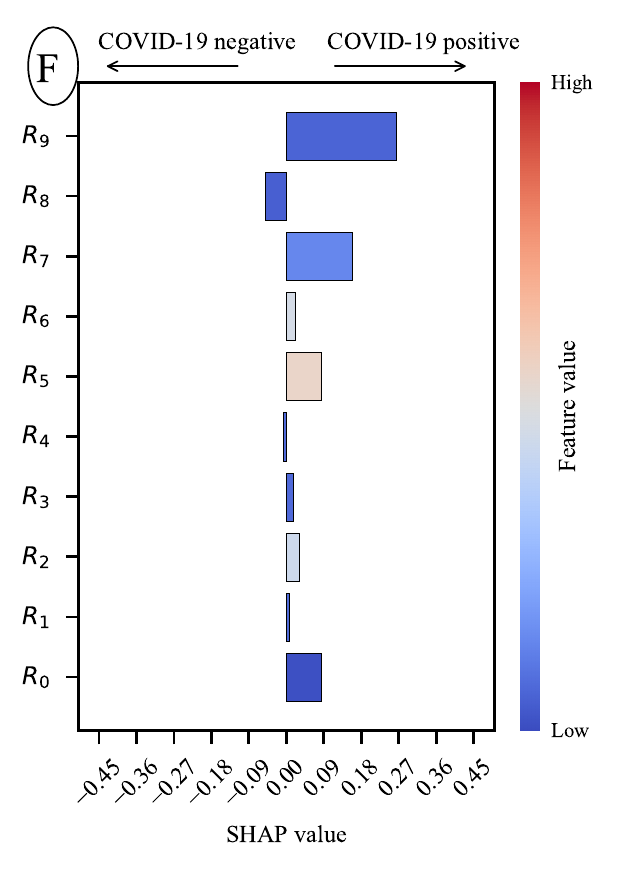}\\

\caption{Local explanations generated using SHAP algorithm for six records for COVID-$19$ positive predictions (panels A, B, and C) and COVID-$19$ negative predictions (panels D, E, and F). Panels A and D show highly confident predictions, panels B and E show low confident predictions, and panels C and F show false positive and false negative predictions, respectively.}
\label{fig:shaplocal} 
\end{figure*}

\begin{table*}[bh]
\centering
\begin{tabular}{||c|c|c|c|c|c|c|c||}
\hline
\multirow{3}{*}{\textbf{Index}} & \multirow{3}{*}{\textbf{True label}} & \textbf{Stage-1}  & \textbf{Stage-2} & \multicolumn{3}{c|}{\textbf{Stage-3}} & \textbf{Stage-4} \\ \cline{3-8}

 & &  \multirow{2}{*}{\textbf{Model prediction}} & \multirow{2}{*}{\textbf{Inspect probability}} & \multicolumn{3}{c|}{\textbf{Turn to feature interpretation}} & \multirow{2}{*}{\textbf{Final decision}} \\ \cline{5-7}
 & &  & & \textbf{Check-1} & \textbf{Check-2} & \textbf{Check-3} & \\ \hline \hline                                        
Panel-A & Positive & Positive  & $0.99$ & \cmark & \cmark & \cmark &  Valid test \\ \hline 
Panel-B & Positive & Positive & $0.67$ & \cmark & \xmark & \xmark &  Re-inspect \\ \hline 
Panel-C & Positive & Negative  & $0.13$ & \xmark & \cmark & \cmark & Re-inspect/ valid test \\ \hline 
Panel-D & Negative & Negative  & $0.07$ &  \cmark & \cmark & \cmark & Valid test \\ \hline 
Panel-E & Negative & Negative  & $0.41$ & \cmark & \xmark & \xmark & Re-inspect  \\ \hline 
Panel-F & Negative & Positive  & $0.92$ & \cmark & \cmark & \xmark & Re-inspect  \\ \hline 
     \end{tabular}
 \rule{0mm}{0mm}    
     \begin{itemize}
    \item \gl{\textbf{Check-1} Are the SHAP feature importances correlated with local feature importance? (For example, check if $R_9$, $R_8$, $R_7$, $R_5$, $R_0$ are also important in local importance.)}
    \item \gl{\textbf{Check-2} Do the feature importance and feature value on local interpretation resemble the majority in the global interpretation in Figure~\ref{fig:shap}?}
    
    \item \gl{\textbf{Check-3} Are the feature importances in local interpretation consistent in their support for COVID-19 outcome? (i.e., all features support COVID-19 negative or positive outcomes.)}
\end{itemize}

\captionof{table}{An explainable AI framework for COVID-$19$ diagnosis decision making.}
\label{table:localInterpretationFramework}
\end{table*}

\begin{figure}[ht]
	\centering
	\includegraphics[width=\linewidth]{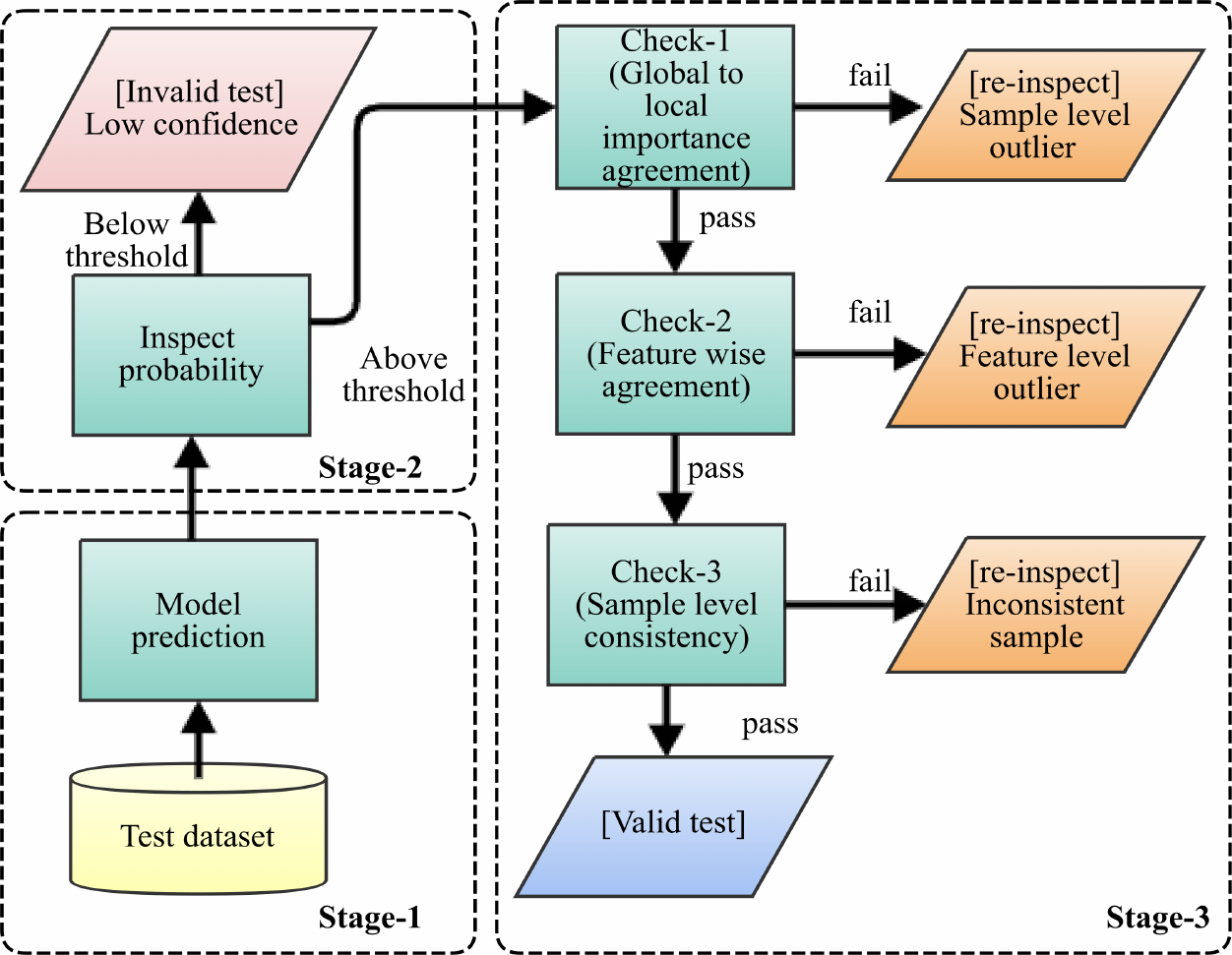}
	\caption{\gl{Our Explainable AI framework.}}
	\label{fig:eswa_data}
\end{figure}
The violin plot provides an approximate idea of any sample's model output based on the feature distribution. This can be immensely helpful to cross-examine the model if the model prediction is not following the trend of the violin plot. For example, suppose the model is predicting COVID-$19$ positive for a sample where the $R_9$ feature value is very high. In that case, we can easily identify that the model is not following the trend, and hence, it needs further validation. It also shows any outliers in the dataset, for instance, higher feature values of $R_8$ (colored in deep red) have a large SHAP value. However, the width of the line is very thin which indicates that the number of samples is potentially lacking or is the case of an outlier. This kind of information is critical to instill trust in users as it allows them to question outputs from the model.
\subsubsection{Local interpretation}
\label{sec:localinterpretation}

\gl{
We present a comprehensive analysis of the RF classifier's interpretation across various prediction scenarios, including a true positive (TP), a false positive (FP), a true negative (TN), and a false negative (FN) (as shown in Figure~\ref{fig:shaplocal} and Table~\ref{table:localInterpretationFramework}). In Figure~\ref{fig:shaplocal}, each bar is encoded with a color gradient ranging from shades of blue to red, representing the lowest and highest feature values, respectively. Similar to the global SHAP plots, the feature importance (or the SHAP values) are shown along the x-axis. A positive value signifies support for COVID-$19$ positive, while a negative for COVID-$19$ negative classification.} In addition to SHAP values, the feature values can be employed to visualize their impact on a COVID-$19$ positive or negative decision. After analyzing several local interpretation plots, we found that the feature value was easier to interpret using color bars than discrete feature values. Therefore, we infer the feature values as low-range (deep blue), high-range (deep red), and mid-range (between deep blue and deep red). The feature values help us see how each sample compares against the rest of the samples in the dataset with the aid of global interpretations in Figure~\ref{fig:shap}. For instance, the feature $R_9$ in panel A is deep blue, indicating a low feature. Now, in Figure~\ref{fig:shap}, the distributions of $R_9$ feature value at the lower end (i.e., deep blue) and the higher end of the spectrum (i.e., deep red) are relatively dense compared to the mid-region. As discussed in Section~\ref{sec:globalinterpretation}, the density of samples in the dataset at any given SHAP value can be realized by the width of the marker in the violin plot. The local interpretation enables us to comprehend how the classifier makes decisions for individual instances by accounting for the feature interactions, shedding light on the model's behavior, and enhancing our trust in its predictions.

\subsubsection{Explainable AI Framework}

\gl{In this section, we present our X-AI framework that compiles model prediction, feature importance, and global and local model interpretation in a systematic framework as illustrated in Figure~\ref{fig:eswa_data}. We organized our X-AI framework in four stages, each designed to systematically interpret the decisions made by our Random Forest (RF) classifier and reach a final conclusion. In each stage, we focus on a specific aspect of interpretation. First, we check the alignment of the local interpretation with the feature importance. Second, we assess the local interpretation and contrast it with the global interpretation to ensure that each feature is consistent with the global interpretation. Lastly, in the third stage, we evaluate the consistency of feature importance in the local interpretation. Following this approach, our framework provides clear, concise, and easy-to-understand insights into the model's predictions, facilitating a comprehensive analysis of the underlying decision-making process. We elaborate on the relevance and properties of each stage below.}

\begin{itemize}
\item \textit{Global feature importance} \gl{In this stage, we compute the feature importance scores and generate global interpretation scores. The feature importance and global model interpretation results will be used in stage-$3$-check-$1$ and stage-$3$-check$2$, respectively.}

\item \textit{Model prediction (Stage-$1$)} \gl{This stage provides the binary COVID-$19$ outcome from our RF classifier.}
\item \textit{Model probability (Stage-$2$)} \gl{Going beyond the prediction, this stage shows us the probability of the RF model. Here, if the probability is not within the threshold, then it is treated as an invalid test due to low confidence.
}

 \item \textit{Feature importance to local feature importance agreement (stage-$3$--check-$1$)}\\ 
\gl{In this stage, we investigate the local feature importance and compare it with the SHAP feature importance displayed in Figure~\ref{fig:featureimportance} panel C. From the global feature importance analysis, we identified the most important features as $R_9$, $R_8$, $R_7$, $R_5$, and $R_0$. By examining whether these features exhibit similar importance in individual samples, we can assess if the sample-based feature importance aligns with the model's feature importance derived from the training dataset.}

\item \textit{Global and local feature importance and value similarity (stage-$3$--check-$2$)}\
\gl{In this step, we analyze the similarity between global and local interpretations of feature importance and feature value. To do this, we compare individual samples from the local interpretation to the global interpretation presented in Figure~\ref{fig:shap}. Recall that the violin plot visualizes the distribution of feature values, with a wider marker indicating more samples and a narrower line representing fewer samples. Samples in narrow regions could be outliers based on feature value or receive special attention from the model, which could lead to undesirable results. For instance, a small change in feature value may result in a significant shift in model prediction. To ensure that local interpretations are consistent with global trends, we verify whether the samples lie in the denser regions of the violin plot, confirming their adherence to the majority. By comparing local and global interpretations in this manner, we can assess whether the model's behavior is consistent across different scales and better understand its decision-making process. Through this analysis, we aim to identify potential issues or inconsistencies in the model's interpretation of feature importance and feature value, which could be critical for improving model performance, addressing potential biases, and enhancing overall interpretability.
}

\item \textit{Consistency in local feature importance support for COVID-19 outcomes (stage-$3$--check-$3$)}\\
\gl{
In this check, we evaluate the consistency of feature importance in local interpretations concerning their support for COVID-$19$ outcomes, i.e., positive or negative. The objective is to determine if the contributions of features in individual samples consistently favor either a positive or negative COVID-$19$ outcome. Ensuring such consistency is vital for the interpretability and reliability of the explainable AI framework, as it helps to establish that the model's local predictions are based on coherent and meaningful feature contributions, thereby facilitating a better understanding of the model's decision-making process.}

\end{itemize}

\subsubsection{Case Study}
\bl{We first analyze a true positive prediction by the RF algorithm, with a probability of $P=0.99$ (Figure~\ref{fig:shaplocal}, panel A). This sample clearly passes stage-$2$ since the probability exceeds the threshold of $>0.7$. In stage-$3$, the sample passes check-$1$ because the most important local features, $R_9$ and $R_8$, align with the most important features in the global feature importance. In check-$2$, we compare the feature values in relation to their SHAP values and contrast them with the global model interpretation presented in the violin plot~\ref{fig:shap}. We observe that features $R_5$, $R_3$, $R_3$, and $R_2$ have large feature values, while others possess smaller values. Comparing the feature value color to the dense region of the violin plot, we find that it falls within the majority distribution, thus passing check-$2$. Finally, in check-$3$, we examine whether all SHAP values are consistent (i.e., uniformly supporting a positive or negative decision). Although one important feature, $R_7$, supports a negative decision, all other features display consistency, leading the sample to pass check-$3$ and be classified as a valid test. Next, in panel D, we show a confident prediction for a true negative case. Here, the sample passes stage-$2$, and all checks in stage-$3$ as the SHAP values align with feature importance, resemble the major distribution, and are consistent, similar to panel A.} 

\bl{Two cases with low confidence in the prediction for COVID-$19$ positive and negative (i.e., the probability is close to $0.5$) are shown in Figure~\ref{fig:shaplocal}, panels B and E. Panel B exhibits a true positive prediction similar to panel A; however, panel B has lower confidence in the prediction with a predicted probability of $P=0.67$ rendering it an invalid test. In stage-$3$, panel B passes check-$1$ as the features $R_9$, $R_8$, and $R_7$ have high SHAP values as the global feature importance. However, panel B fails check-$2$ since the features $R_2$-$R_5$ do not fall in the dense regions for positive SHAP values in the global importance. Lastly, panel B does not pass check-$3$, as well as it has four features supporting each of the positive and negative classes, and the remaining two have close to $0$ SHAP values. In our second example for low confidence prediction ($P=0.41$) in panel E, we see that it passes check-$1$ as the blue color in feature value can be found in dense regions of the violin plot. However, it fails check-$2$ and check-$3$ for the same reasons as panel B. Since both panel B and panel E failed stage $3$, they need a re-inspection or re-testing.} 

\bl{Lastly, we analyze two incorrect predictions in our test set: a false negative (panel C) with a probability of $P=0.13$ and a false positive (panel F) with a probability of $P=0.92$. In panel C, even though the sample belongs to the positive class, all SHAP values support a negative outcome. Notably, the SHAP values of panel C match those of panel D (a true negative prediction). This similarity arises because the positive sample has feature values closely resembling the negative sample, leading to a similar prediction for panel C as for panel D. Panel C does not pass check-$1$ since only features $R_9$ and $R_8$ are important, while the others have minimal contributions. The result's outcome may necessitate considering patient symptom information to determine whether a re-test is required or if the result can be deemed valid. In panel F, the feature importance magnitudes match the global feature importance; however, the most important feature, $R_8$, displays an inversion, while the other features support a positive prediction. This indicates that panel F passes check-$1$ and check-$2$ but fails check-$3$.}
           
\section{DISCUSSION}
\label{sec:discussion} 

\bl{While analyzing our results through different interpretation techniques, we observed quite remarkably that all three methods of global feature interpretation consistently resulted in the most important features as $R_8$ and $R_9$ (Table~\ref{table:globalFI}). The meaning of these features is of particular interest because of the biomarker ranges making up the features, $AUC_{B_2}/AUC_{B_4}$ for $R_8$ and $AUC_{B_3}/AUC_{B_4}$ for $R_9$. First, the features $R_8$ and $R_9$ are virus-to-host features, which may represent the interaction between host immune response and viral antigen, as described in~\cite{chivte2021maldi}. Second, the viral protein fragment corresponding to $B_3$ is tentatively suggested to be the S$1$ subunit of the viral spike protein (see Table~\ref{table:bioRanges}). S$1$ is a key viral biomarker in the mass spectra because the observed m/z ratio does not overlap significantly with other proteins or fragments from the gargle sample. Furthermore, because our model uses SHAP values, it was particularly remarkable that $R_9$ was the most important feature, as it contains $B_3$. Under a similar interpretation of our model, $R_6$ was deemed to be the least important feature, which contains $B_1$ and $B_4$, a host-to-host protein ratio. As no viral protein is represented in $R_6$, it follows that such a feature would not be useful in the diagnosis of COVID-$19$. It should be noted once again that each biomarker has been assigned a putative identity based on preliminary studies only and have yet to be unambiguously characterized. }
      

Turning to local feature interpretations, these could offer a powerful, comprehensive diagnostic view in contrast to a binary decision by RT-PCR or an algorithm with percent confidence. The local explanations generated using SHAP values show the contribution of each feature in each sample. This could allow domain experts and health officials administering such a test to make more informed decisions based on the putative host-to-viral or viral-to-host ratios for case-by-case bases, as seen in Figure~\ref{fig:shaplocal}. If the algorithm decides positive or negative, the local feature explanation will provide additional information as to \textit{why} the test output its decision. Although data augmentation techniques could enrich the dataset with more artificial samples, we avoided doing this because it could have made the dataset noisy by adding false patterns. Nonetheless, not only in this diagnostic approach is the viral aspect of COVID-$19$ considered (as is only considered in PCR-based tests), but the human immune response towards the SARS-CoV-$2$ virus is also given due consideration. The results show that machine learning algorithms (e.g., RF) coupled with the utility of X-AI interpretability could be implemented into healthcare to address the limitations of black box systems and improve the diagnostic process, from test output to informed care.

In summary, from training machine learning algorithms to explaining the trained models, each aspect may be particularly interesting for different reasons. For example, during the AI-model building stages, a closer view of calibration, probabilities, confusion matrices, precision, and recall values for the training and testing datasets can prevent overfitting and bias. Following that, the interpretation of a model on a global and local level can have several applications. First, from a machine learning perspective, feature importance can help identify spurious correlations between features and diagnoses and monitor model drift. Second, biochemistry researchers can learn and investigate the relationships between feature value to its importance in the outcome of COVID-$19$ diagnosis. Finally, in a clinical setting, the local model interpretations, along with the probability scores, can aid in determining whether a test was properly executed or if a follow-up test was required. 

\bl{In the case of a future pandemic, MALDI-ToF coupled with AI can be used as an alternative or in-addition-to testing platform for diagnosing viral infections. MALDI-ToF MS could be advantageous in clinical virology for the option of automation, a low cost-to-efficiency ratio, rapid identification, high throughput testing, high accuracy, and reliability. Even though the cost of a MALDI-ToF instrument is higher than the materials and instrumentation for RT-PCR, purchasing a MALDI-ToF mass spectrometer is a one-time investment. Furthermore, the cost-per-test for RT-PCR testing is significantly higher. It has been estimated by \cite{deulofeu2021detection} that the cost for a MALDI-ToF test is at least $75\%$ lower than an RT-PCR test. A MALDI-ToF-based testing platform could allow scientists to develop and validate novel methods for the diagnosis of diseases. As we have seen in this pandemic, the complete reliance on the gold standard testing technique limits the ability to perform early mass testing due to supply limitations. In addition, although previous research has shown some good results incorporating AI algorithms, in this study, we show adding an additional layer of X-AI and designing a three-stage outcome can improve efficiency and generate explanations from different perspectives.} 

\section{CONCLUSION}
\label{sec:conclusion}
In this work, we have designed an AI-based COVID-$19$ diagnostic method using the MALDI-ToF MS protein profiles of human gargle samples. After thoroughly validating various machine learning algorithms, we found the performance of RF to be the best among others. We evaluated the RF classifier on a $70\%$--$30\%$ train-test-split strategy where the accuracy on the test dataset was $94.12\%$. We further utilized the concepts of X-AI and interpreted the RF algorithm using SHAP and feature importance techniques such as permutation feature importance and impurity-based feature importance to find the most important features. Using these interpretation models, we showed global and local explanations of the model's decision on COVID testing. Explaining the rationale behind the decision improve your trustworthiness of our approach by enabling domain experts and medical practitioners to understand the mechanisms of the black-box AI models when used on a real-time basis. Additionally, this methodology may allow for more detailed information about the life cycle of viral infection along with the COVID-$19$ status ascertained from the MALDI-ToF MS data. 

\subsection*{Declaration of competing interest}
The authors declare that there is no conflict of interest regarding the publication of the article.
\subsection*{Acknowledgements}
\label{sec:acknowledgements}
 This work was supported by Northern Illinois University's Molecular Analysis Core which was established with support from Shimadzu Scientific Instruments.

\bibliography{cas-refs}

\begin{thebibliography}{}

\bibitem [\protect \citeauthoryear {%
Abadi%
\ \protect \BOthers {.}}{%
Abadi%
\ \protect \BOthers {.}}{%
{\protect \APACyear {2016}}%
}]{%
abadi2016tensorflow}
\APACinsertmetastar {%
abadi2016tensorflow}%
\begin{APACrefauthors}%
Abadi, M.%
, Barham, P.%
, Chen, J.%
, Chen, Z.%
, Davis, A.%
, Dean, J.%
, Devin, M.%
, Ghemawat, S.%
, Irving, G.%
, Isard, M.%
, Kudlur, M.%
, Levenberg, J.%
, Monga, R.%
, Moore, S.%
, Murray, D\BPBI G.%
, Steiner, B.%
, Tucker, P.%
, Vasudevan, V.%
, Warden, P.%
, Wicke, M.%
, Yu, Y.%
\BCBL {}\ \BBA {} Zheng, X.%
\end{APACrefauthors}%
\unskip\
\newblock
\APACrefYearMonthDay{2016}{}{}.
\newblock
{\BBOQ}\APACrefatitle {{TensorFlow}: a system for {Large-Scale} machine
  learning} {{TensorFlow}: a system for {Large-Scale} machine learning}.{\BBCQ}
\newblock
\BIn{} \APACrefbtitle {12th USENIX symposium on operating systems design and
  implementation (OSDI 16)} {12th usenix symposium on operating systems design
  and implementation (osdi 16)}\ (\BPGS\ 265--283).
\PrintBackRefs{\CurrentBib}

\bibitem [\protect \citeauthoryear {%
Abraham%
\ \protect \BOthers {.}}{%
Abraham%
\ \protect \BOthers {.}}{%
{\protect \APACyear {2012}}%
}]{%
abraham2012saliva}
\APACinsertmetastar {%
abraham2012saliva}%
\begin{APACrefauthors}%
Abraham, J\BPBI E.%
, Maranian, M\BPBI J.%
, Spiteri, I.%
, Russell, R.%
, Ingle, S.%
, Luccarini, C.%
, Earl, H\BPBI M.%
, Pharoah, P\BPBI P.%
, Dunning, A\BPBI M.%
\BCBL {}\ \BBA {} Caldas, C.%
\end{APACrefauthors}%
\unskip\
\newblock
\APACrefYearMonthDay{2012}{}{}.
\newblock
{\BBOQ}\APACrefatitle {Saliva samples are a viable alternative to blood samples
  as a source of {DNA} for high throughput genotyping} {Saliva samples are a
  viable alternative to blood samples as a source of {DNA} for high throughput
  genotyping}.{\BBCQ}
\newblock
\APACjournalVolNumPages{BMC Medical Genomics}{5}{}{1--6}.
\PrintBackRefs{\CurrentBib}

\bibitem [\protect \citeauthoryear {%
Alves%
\ \protect \BOthers {.}}{%
Alves%
\ \protect \BOthers {.}}{%
{\protect \APACyear {2021}}%
}]{%
alves2021explaining}
\APACinsertmetastar {%
alves2021explaining}%
\begin{APACrefauthors}%
Alves, M\BPBI A.%
, Castro, G\BPBI Z.%
, Oliveira, B\BPBI A\BPBI S.%
, Ferreira, L\BPBI A.%
, Ram{\'\i}rez, J\BPBI A.%
, Silva, R.%
\BCBL {}\ \BBA {} Guimar{\~a}es, F\BPBI G.%
\end{APACrefauthors}%
\unskip\
\newblock
\APACrefYearMonthDay{2021}{}{}.
\newblock
{\BBOQ}\APACrefatitle {Explaining machine learning based diagnosis of COVID-19
  from routine blood tests with decision trees and criteria graphs} {Explaining
  machine learning based diagnosis of covid-19 from routine blood tests with
  decision trees and criteria graphs}.{\BBCQ}
\newblock
\APACjournalVolNumPages{Computers in Biology and Medicine}{132}{}{104335}.
\PrintBackRefs{\CurrentBib}

\bibitem [\protect \citeauthoryear {%
Ardakani%
, Kanafi%
, Acharya%
, Khadem%
\BCBL {}\ \BBA {} Mohammadi%
}{%
Ardakani%
\ \protect \BOthers {.}}{%
{\protect \APACyear {2020}}%
}]{%
ardakani2020application}
\APACinsertmetastar {%
ardakani2020application}%
\begin{APACrefauthors}%
Ardakani, A\BPBI A.%
, Kanafi, A\BPBI R.%
, Acharya, U\BPBI R.%
, Khadem, N.%
\BCBL {}\ \BBA {} Mohammadi, A.%
\end{APACrefauthors}%
\unskip\
\newblock
\APACrefYearMonthDay{2020}{}{}.
\newblock
{\BBOQ}\APACrefatitle {Application of deep learning technique to manage
  COVID-19 in routine clinical practice using CT images: Results of 10
  convolutional neural networks} {Application of deep learning technique to
  manage covid-19 in routine clinical practice using ct images: Results of 10
  convolutional neural networks}.{\BBCQ}
\newblock
\APACjournalVolNumPages{Computers in biology and medicine}{121}{}{103795}.
\PrintBackRefs{\CurrentBib}

\bibitem [\protect \citeauthoryear {%
Basu%
, Sheikh%
, Cuevas%
\BCBL {}\ \BBA {} Sarkar%
}{%
Basu%
\ \protect \BOthers {.}}{%
{\protect \APACyear {2022}}%
}]{%
basu2022covid}
\APACinsertmetastar {%
basu2022covid}%
\begin{APACrefauthors}%
Basu, A.%
, Sheikh, K\BPBI H.%
, Cuevas, E.%
\BCBL {}\ \BBA {} Sarkar, R.%
\end{APACrefauthors}%
\unskip\
\newblock
\APACrefYearMonthDay{2022}{}{}.
\newblock
{\BBOQ}\APACrefatitle {COVID-19 detection from CT scans using a two-stage
  framework} {Covid-19 detection from ct scans using a two-stage
  framework}.{\BBCQ}
\newblock
\APACjournalVolNumPages{Expert Systems with Applications}{193}{}{116377}.
\PrintBackRefs{\CurrentBib}

\bibitem [\protect \citeauthoryear {%
Benjamini%
\ \BBA {} Hochberg%
}{%
Benjamini%
\ \BBA {} Hochberg%
}{%
{\protect \APACyear {1995}}%
}]{%
benjamini1995controlling}
\APACinsertmetastar {%
benjamini1995controlling}%
\begin{APACrefauthors}%
Benjamini, Y.%
\BCBT {}\ \BBA {} Hochberg, Y.%
\end{APACrefauthors}%
\unskip\
\newblock
\APACrefYearMonthDay{1995}{January}{}.
\newblock
{\BBOQ}\APACrefatitle {Controlling the false discovery rate: a practical and
  powerful approach to multiple testing} {Controlling the false discovery rate:
  a practical and powerful approach to multiple testing}.{\BBCQ}
\newblock
\APACjournalVolNumPages{Journal of the Royal Statistical Society: Series B
  (Methodological)}{57}{1}{289--300}.
\PrintBackRefs{\CurrentBib}

\bibitem [\protect \citeauthoryear {%
Breiman%
}{%
Breiman%
}{%
{\protect \APACyear {2001}}%
}]{%
breiman2001random}
\APACinsertmetastar {%
breiman2001random}%
\begin{APACrefauthors}%
Breiman, L.%
\end{APACrefauthors}%
\unskip\
\newblock
\APACrefYearMonthDay{2001}{October}{}.
\newblock
{\BBOQ}\APACrefatitle {Random forests} {Random forests}.{\BBCQ}
\newblock
\APACjournalVolNumPages{Machine Learning}{45}{1}{5--32}.
\PrintBackRefs{\CurrentBib}

\bibitem [\protect \citeauthoryear {%
Carter%
\ \protect \BOthers {.}}{%
Carter%
\ \protect \BOthers {.}}{%
{\protect \APACyear {2020}}%
}]{%
carter2020assay}
\APACinsertmetastar {%
carter2020assay}%
\begin{APACrefauthors}%
Carter, L\BPBI J.%
, Garner, L\BPBI V.%
, Smoot, J\BPBI W.%
, Li, Y.%
, Zhou, Q.%
, Saveson, C\BPBI J.%
, Sasso, J\BPBI M.%
, Gregg, A\BPBI C.%
, Soares, D\BPBI J.%
, Beskid, T\BPBI R.%
, Jervey, S\BPBI R.%
\BCBL {}\ \BBA {} Liu, C.%
\end{APACrefauthors}%
\unskip\
\newblock
\APACrefYearMonthDay{2020}{April}{}.
\newblock
{\BBOQ}\APACrefatitle {Assay techniques and test development for {COVID-19}
  diagnosis} {Assay techniques and test development for {COVID-19}
  diagnosis}.{\BBCQ}
\newblock
\APACjournalVolNumPages{ACS Central Science}{6}{5}{591--605}.
\PrintBackRefs{\CurrentBib}

\bibitem [\protect \citeauthoryear {%
Ceriani%
\ \BBA {} Verme%
}{%
Ceriani%
\ \BBA {} Verme%
}{%
{\protect \APACyear {2012}}%
}]{%
ceriani2012origins}
\APACinsertmetastar {%
ceriani2012origins}%
\begin{APACrefauthors}%
Ceriani, L.%
\BCBT {}\ \BBA {} Verme, P.%
\end{APACrefauthors}%
\unskip\
\newblock
\APACrefYearMonthDay{2012}{September}{}.
\newblock
{\BBOQ}\APACrefatitle {The origins of the Gini index: extracts from
  Variabilit{\`a} e Mutabilit{\`a} (1912) by Corrado Gini} {The origins of the
  gini index: extracts from variabilit{\`a} e mutabilit{\`a} (1912) by corrado
  gini}.{\BBCQ}
\newblock
\APACjournalVolNumPages{The Journal of Economic Inequality}{10}{3}{421--443}.
\PrintBackRefs{\CurrentBib}

\bibitem [\protect \citeauthoryear {%
Chen%
\ \BBA {} Guestrin%
}{%
Chen%
\ \BBA {} Guestrin%
}{%
{\protect \APACyear {2016}}%
}]{%
chen2016xgboost}
\APACinsertmetastar {%
chen2016xgboost}%
\begin{APACrefauthors}%
Chen, T.%
\BCBT {}\ \BBA {} Guestrin, C.%
\end{APACrefauthors}%
\unskip\
\newblock
\APACrefYearMonthDay{2016}{August}{}.
\newblock
{\BBOQ}\APACrefatitle {XGBoost: A scalable tree boosting system} {Xgboost: A
  scalable tree boosting system}.{\BBCQ}
\newblock
\APACjournalVolNumPages{Proceedings of the 22nd ACM SIGKDD International
  Conference on Knowledge Discovery and Data Mining}{}{}{785–-794}.
\PrintBackRefs{\CurrentBib}

\bibitem [\protect \citeauthoryear {%
Chivte%
\ \protect \BOthers {.}}{%
Chivte%
\ \protect \BOthers {.}}{%
{\protect \APACyear {2021}}%
}]{%
chivte2021maldi}
\APACinsertmetastar {%
chivte2021maldi}%
\begin{APACrefauthors}%
Chivte, P.%
, LaCasse, Z.%
, Seethi, V\BPBI D\BPBI R.%
, Bharti, P.%
, Bland, J.%
, Kadkol, S\BPBI S.%
\BCBL {}\ \BBA {} Gaillard, E\BPBI R.%
\end{APACrefauthors}%
\unskip\
\newblock
\APACrefYearMonthDay{2021}{August}{}.
\newblock
{\BBOQ}\APACrefatitle {{MALDI-ToF} protein profiling as a potential rapid
  diagnostic platform for {COVID-19}} {{MALDI-ToF} protein profiling as a
  potential rapid diagnostic platform for {COVID-19}}.{\BBCQ}
\newblock
\APACjournalVolNumPages{Journal of Mass Spectrometry and Advances in the
  Clinical Lab}{21}{}{31--41}.
\PrintBackRefs{\CurrentBib}

\bibitem [\protect \citeauthoryear {%
Costa%
\ \protect \BOthers {.}}{%
Costa%
\ \protect \BOthers {.}}{%
{\protect \APACyear {2022}}%
}]{%
costa2022exploratory}
\APACinsertmetastar {%
costa2022exploratory}%
\begin{APACrefauthors}%
Costa, M\BPBI M.%
, Martin, H.%
, Estellon, B.%
, Dup{\'e}, F\BHBI X.%
, Saby, F.%
, Benoit, N.%
, Tissot-Dupont, H.%
, Million, M.%
, Pradines, B.%
, Granjeaud, S.%
\BCBL {}\ \BBA {} Almeras, L.%
\end{APACrefauthors}%
\unskip\
\newblock
\APACrefYearMonthDay{2022}{January}{}.
\newblock
{\BBOQ}\APACrefatitle {Exploratory study on application of {MALDI-TOF-MS} to
  detect {SARS-CoV-2} infection in human saliva} {Exploratory study on
  application of {MALDI-TOF-MS} to detect {SARS-CoV-2} infection in human
  saliva}.{\BBCQ}
\newblock
\APACjournalVolNumPages{Journal of Clinical Medicine}{11}{2}{295}.
\PrintBackRefs{\CurrentBib}

\bibitem [\protect \citeauthoryear {%
de Almeida%
\ \protect \BOthers {.}}{%
de Almeida%
\ \protect \BOthers {.}}{%
{\protect \APACyear {2022}}%
}]{%
de2022maldi}
\APACinsertmetastar {%
de2022maldi}%
\begin{APACrefauthors}%
de Almeida, C\BPBI M.%
, Motta, L\BPBI C.%
, Folli, G\BPBI S.%
, Marcarini, W\BPBI D.%
, Costa, C\BPBI A.%
, Vilela, A\BPBI C.%
, Barauna, V\BPBI G.%
, Martin, F\BPBI L.%
, Singh, M\BPBI N.%
, Campos, L\BPBI C.%
, Costa, N\BPBI L.%
, Vassallo, P\BPBI F.%
, Chaves, A\BPBI R.%
, Endringer, D\BPBI C.%
, Mill, J\BPBI G.%
, Filgueiras, P\BPBI R.%
\BCBL {}\ \BBA {} Ram\~{a}o, W.%
\end{APACrefauthors}%
\unskip\
\newblock
\APACrefYearMonthDay{2022}{}{}.
\newblock
{\BBOQ}\APACrefatitle {{MALDI (+) FT-ICR Mass Spectrometry (MS)} Combined with
  Machine Learning toward Saliva-Based Diagnostic Screening for {COVID-19}}
  {{MALDI (+) FT-ICR Mass Spectrometry (MS)} combined with machine learning
  toward saliva-based diagnostic screening for {COVID-19}}.{\BBCQ}
\newblock
\APACjournalVolNumPages{Journal of Proteome Research}{21}{8}{1868--1875}.
\PrintBackRefs{\CurrentBib}

\bibitem [\protect \citeauthoryear {%
Deulofeu%
\ \protect \BOthers {.}}{%
Deulofeu%
\ \protect \BOthers {.}}{%
{\protect \APACyear {2021}}%
}]{%
deulofeu2021detection}
\APACinsertmetastar {%
deulofeu2021detection}%
\begin{APACrefauthors}%
Deulofeu, M.%
, Garc{\'\i}a-Cuesta, E.%
, Pe{\~n}a-M{\'e}ndez, E\BPBI M.%
, Conde, J\BPBI E.%
, Jim{\'e}nez-Romero, O.%
, Verd{\'u}, E.%
, Serrando, M\BPBI T.%
, Salvad{\'o}, V.%
\BCBL {}\ \BBA {} Boadas-Vaello, P.%
\end{APACrefauthors}%
\unskip\
\newblock
\APACrefYearMonthDay{2021}{April}{}.
\newblock
{\BBOQ}\APACrefatitle {Detection of {SARS-CoV-2} infection in human
  nasopharyngeal samples by combining {MALDI-ToF-MS} and artificial
  intelligence} {Detection of {SARS-CoV-2} infection in human nasopharyngeal
  samples by combining {MALDI-ToF-MS} and artificial intelligence}.{\BBCQ}
\newblock
\APACjournalVolNumPages{Frontiers in Medicine}{8}{}{398}.
\PrintBackRefs{\CurrentBib}

\bibitem [\protect \citeauthoryear {%
Esser%
\ \protect \BOthers {.}}{%
Esser%
\ \protect \BOthers {.}}{%
{\protect \APACyear {2008}}%
}]{%
esser2008sample}
\APACinsertmetastar {%
esser2008sample}%
\begin{APACrefauthors}%
Esser, D.%
, Alvarez-Llamas, G.%
, De~Vries, M\BPBI P.%
, Weening, D.%
, Vonk, R\BPBI J.%
\BCBL {}\ \BBA {} Roelofsen, H.%
\end{APACrefauthors}%
\unskip\
\newblock
\APACrefYearMonthDay{2008}{}{}.
\newblock
{\BBOQ}\APACrefatitle {Sample stability and protein composition of saliva:
  implications for its use as a diagnostic fluid} {Sample stability and protein
  composition of saliva: implications for its use as a diagnostic
  fluid}.{\BBCQ}
\newblock
\APACjournalVolNumPages{Biomarker insights}{3}{}{BMI--S607}.
\PrintBackRefs{\CurrentBib}

\bibitem [\protect \citeauthoryear {%
Feng%
\ \protect \BOthers {.}}{%
Feng%
\ \protect \BOthers {.}}{%
{\protect \APACyear {2020}}%
}]{%
feng2020molecular}
\APACinsertmetastar {%
feng2020molecular}%
\begin{APACrefauthors}%
Feng, W.%
, Newbigging, A\BPBI M.%
, Le, C.%
, Pang, B.%
, Peng, H.%
, Cao, Y.%
, Wu, J.%
, Abbas, G.%
, Song, J.%
, Wang, D\BHBI B.%
, Cui, M.%
, Tao, J.%
, Tyrrell, D\BPBI L.%
, Zhang, X\BHBI E.%
, Zhang, H.%
\BCBL {}\ \BBA {} Le, X\BPBI C.%
\end{APACrefauthors}%
\unskip\
\newblock
\APACrefYearMonthDay{2020}{June}{}.
\newblock
{\BBOQ}\APACrefatitle {Molecular diagnosis of {COVID-19}: challenges and
  research needs} {Molecular diagnosis of {COVID-19}: challenges and research
  needs}.{\BBCQ}
\newblock
\APACjournalVolNumPages{Analytical chemistry}{92}{15}{10196--10209}.
\PrintBackRefs{\CurrentBib}

\bibitem [\protect \citeauthoryear {%
Filchakova%
\ \protect \BOthers {.}}{%
Filchakova%
\ \protect \BOthers {.}}{%
{\protect \APACyear {2022}}%
}]{%
filchakova2022review}
\APACinsertmetastar {%
filchakova2022review}%
\begin{APACrefauthors}%
Filchakova, O.%
, Dossym, D.%
, Ilyas, A.%
, Kuanysheva, T.%
, Abdizhamil, A.%
\BCBL {}\ \BBA {} Bukasov, R.%
\end{APACrefauthors}%
\unskip\
\newblock
\APACrefYearMonthDay{2022}{}{}.
\newblock
{\BBOQ}\APACrefatitle {Review of {COVID-19} testing and diagnostic methods}
  {Review of {COVID-19} testing and diagnostic methods}.{\BBCQ}
\newblock
\APACjournalVolNumPages{Talanta}{}{}{123409}.
\PrintBackRefs{\CurrentBib}

\bibitem [\protect \citeauthoryear {%
Garza%
\ \protect \BOthers {.}}{%
Garza%
\ \protect \BOthers {.}}{%
{\protect \APACyear {2021}}%
}]{%
garza2021rapid}
\APACinsertmetastar {%
garza2021rapid}%
\begin{APACrefauthors}%
Garza, K\BPBI Y.%
, Silva, A\BPBI A\BPBI R.%
, Rosa, J\BPBI R.%
, Keating, M\BPBI F.%
, Povilaitis, S\BPBI C.%
, Spradlin, M.%
, Sanches, P\BPBI H\BPBI G.%
, Var\~{a}o Moura, A.%
, Marrero~Gutierrez, J.%
, Lin, J\BPBI Q.%
, Zhang, J.%
, DeHoog, R\BPBI J.%
, Bensussan, A.%
, Badal, S.%
, Cardoso~de Oliveira, D.%
, Dias~Garcia, P\BPBI H.%
, Dias~de Oliveira~Negrini, L.%
, Antonio, M\BPBI A.%
, Canevari, T\BPBI C.%
, Eberlin, M\BPBI N.%
, Tibshirani, R.%
, Eberlin, L\BPBI S.%
\BCBL {}\ \BBA {} Porcari, A\BPBI M.%
\end{APACrefauthors}%
\unskip\
\newblock
\APACrefYearMonthDay{2021}{}{}.
\newblock
{\BBOQ}\APACrefatitle {Rapid screening of COVID-19 directly from clinical
  nasopharyngeal swabs using the MasSpec Pen} {Rapid screening of covid-19
  directly from clinical nasopharyngeal swabs using the masspec pen}.{\BBCQ}
\newblock
\APACjournalVolNumPages{Analytical chemistry}{93}{37}{12582--12593}.
\PrintBackRefs{\CurrentBib}

\bibitem [\protect \citeauthoryear {%
Gong%
, Wang%
, Zhang%
, Elahe%
\BCBL {}\ \BBA {} Jin%
}{%
Gong%
\ \protect \BOthers {.}}{%
{\protect \APACyear {2022}}%
}]{%
gong2022explainable}
\APACinsertmetastar {%
gong2022explainable}%
\begin{APACrefauthors}%
Gong, H.%
, Wang, M.%
, Zhang, H.%
, Elahe, M\BPBI F.%
\BCBL {}\ \BBA {} Jin, M.%
\end{APACrefauthors}%
\unskip\
\newblock
\APACrefYearMonthDay{2022}{June}{}.
\newblock
{\BBOQ}\APACrefatitle {An Explainable AI Approach for the Rapid Diagnosis of
  {COVID-19} Using Ensemble Learning Algorithms} {An explainable ai approach
  for the rapid diagnosis of {COVID-19} using ensemble learning
  algorithms}.{\BBCQ}
\newblock
\APACjournalVolNumPages{Frontiers in Public Health}{10}{}{874455}.
\PrintBackRefs{\CurrentBib}

\bibitem [\protect \citeauthoryear {%
Goodfellow%
, Bengio%
\BCBL {}\ \BBA {} Courville%
}{%
Goodfellow%
\ \protect \BOthers {.}}{%
{\protect \APACyear {2016}}%
}]{%
goodfellow2016deep}
\APACinsertmetastar {%
goodfellow2016deep}%
\begin{APACrefauthors}%
Goodfellow, I.%
, Bengio, Y.%
\BCBL {}\ \BBA {} Courville, A.%
\end{APACrefauthors}%
\unskip\
\newblock
\APACrefYear{2016}.
\newblock
\APACrefbtitle {Deep learning} {Deep learning}.
\newblock
\APACaddressPublisher{}{MIT press}.
\PrintBackRefs{\CurrentBib}

\bibitem [\protect \citeauthoryear {%
Guyon%
, Weston%
, Barnhill%
\BCBL {}\ \BBA {} Vapnik%
}{%
Guyon%
\ \protect \BOthers {.}}{%
{\protect \APACyear {2002}}%
}]{%
guyon2002gene}
\APACinsertmetastar {%
guyon2002gene}%
\begin{APACrefauthors}%
Guyon, I.%
, Weston, J.%
, Barnhill, S.%
\BCBL {}\ \BBA {} Vapnik, V.%
\end{APACrefauthors}%
\unskip\
\newblock
\APACrefYearMonthDay{2002}{January}{}.
\newblock
{\BBOQ}\APACrefatitle {Gene selection for cancer classification using support
  vector machines} {Gene selection for cancer classification using support
  vector machines}.{\BBCQ}
\newblock
\APACjournalVolNumPages{Machine Learning}{46}{1}{389--422}.
\PrintBackRefs{\CurrentBib}

\bibitem [\protect \citeauthoryear {%
Habibzadeh%
, Mofatteh%
, Silawi%
, Ghavami%
\BCBL {}\ \BBA {} Faghihi%
}{%
Habibzadeh%
\ \protect \BOthers {.}}{%
{\protect \APACyear {2021}}%
}]{%
habibzadeh2021molecular}
\APACinsertmetastar {%
habibzadeh2021molecular}%
\begin{APACrefauthors}%
Habibzadeh, P.%
, Mofatteh, M.%
, Silawi, M.%
, Ghavami, S.%
\BCBL {}\ \BBA {} Faghihi, M\BPBI A.%
\end{APACrefauthors}%
\unskip\
\newblock
\APACrefYearMonthDay{2021}{}{}.
\newblock
{\BBOQ}\APACrefatitle {Molecular diagnostic assays for {COVID-19}: an overview}
  {Molecular diagnostic assays for {COVID-19}: an overview}.{\BBCQ}
\newblock
\APACjournalVolNumPages{Critical reviews in clinical laboratory
  sciences}{58}{6}{385--398}.
\PrintBackRefs{\CurrentBib}

\bibitem [\protect \citeauthoryear {%
Harris%
\ \protect \BOthers {.}}{%
Harris%
\ \protect \BOthers {.}}{%
{\protect \APACyear {2020}}%
}]{%
harris2020array}
\APACinsertmetastar {%
harris2020array}%
\begin{APACrefauthors}%
Harris, C\BPBI R.%
, Millman, K\BPBI J.%
, Van Der~Walt, S\BPBI J.%
, Gommers, R.%
, Virtanen, P.%
, Cournapeau, D.%
, Wieser, E.%
, Taylor, J.%
, Berg, S.%
, Smith, N\BPBI J.%
, Kern, R.%
, Picus, M.%
, Hoyer, S.%
, Van~Kerkwijk, M\BPBI H.%
, Brett, M.%
, Haldane, A.%
, del R{\'i}o, J\BPBI F.%
, Wiebe, M.%
, Peterson, P.%
, G{\'e}rard-Marchant, P.%
, Sheppard, K.%
, Reddy, T.%
, Weckesser, W.%
, Abbasi, H.%
, Gohlke, C.%
\BCBL {}\ \BBA {} Oliphant, T\BPBI E.%
\end{APACrefauthors}%
\unskip\
\newblock
\APACrefYearMonthDay{2020}{}{}.
\newblock
{\BBOQ}\APACrefatitle {Array programming with NumPy} {Array programming with
  numpy}.{\BBCQ}
\newblock
\APACjournalVolNumPages{Nature}{585}{7825}{357--362}.
\PrintBackRefs{\CurrentBib}

\bibitem [\protect \citeauthoryear {%
Hosmer%
, Lemeshow%
\BCBL {}\ \BBA {} Sturdivant%
}{%
Hosmer%
\ \protect \BOthers {.}}{%
{\protect \APACyear {2013}}%
}]{%
hosmer2013applied}
\APACinsertmetastar {%
hosmer2013applied}%
\begin{APACrefauthors}%
Hosmer, D\BPBI W.%
, Lemeshow, S.%
\BCBL {}\ \BBA {} Sturdivant, R\BPBI X.%
\end{APACrefauthors}%
\unskip\
\newblock
\APACrefYearMonthDay{2013}{}{}.
\newblock
\APACrefbtitle {Applied logistic regression. Hoboken.} {Applied logistic
  regression. hoboken.}
\newblock
\APACaddressPublisher{}{NJ: wiley}.
\PrintBackRefs{\CurrentBib}

\bibitem [\protect \citeauthoryear {%
Hu%
\ \protect \BOthers {.}}{%
Hu%
\ \protect \BOthers {.}}{%
{\protect \APACyear {2022}}%
}]{%
hu2022explainable}
\APACinsertmetastar {%
hu2022explainable}%
\begin{APACrefauthors}%
Hu, Q.%
, Gois, F\BPBI N\BPBI B.%
, Costa, R.%
, Zhang, L.%
, Yin, L.%
, Magaia, N.%
\BCBL {}\ \BBA {} de Albuquerque, V\BPBI H\BPBI C.%
\end{APACrefauthors}%
\unskip\
\newblock
\APACrefYearMonthDay{2022}{}{}.
\newblock
{\BBOQ}\APACrefatitle {Explainable artificial intelligence-based edge fuzzy
  images for COVID-19 detection and identification} {Explainable artificial
  intelligence-based edge fuzzy images for covid-19 detection and
  identification}.{\BBCQ}
\newblock
\APACjournalVolNumPages{Applied Soft Computing}{123}{}{108966}.
\PrintBackRefs{\CurrentBib}

\bibitem [\protect \citeauthoryear {%
Khan%
\ \protect \BOthers {.}}{%
Khan%
\ \protect \BOthers {.}}{%
{\protect \APACyear {2021}}%
}]{%
khan2021applications}
\APACinsertmetastar {%
khan2021applications}%
\begin{APACrefauthors}%
Khan, M.%
, Mehran, M\BPBI T.%
, Haq, Z\BPBI U.%
, Ullah, Z.%
, Naqvi, S\BPBI R.%
, Ihsan, M.%
\BCBL {}\ \BBA {} Abbass, H.%
\end{APACrefauthors}%
\unskip\
\newblock
\APACrefYearMonthDay{2021}{}{}.
\newblock
{\BBOQ}\APACrefatitle {Applications of artificial intelligence in COVID-19
  pandemic: A comprehensive review} {Applications of artificial intelligence in
  covid-19 pandemic: A comprehensive review}.{\BBCQ}
\newblock
\APACjournalVolNumPages{Expert systems with applications}{185}{}{115695}.
\PrintBackRefs{\CurrentBib}

\bibitem [\protect \citeauthoryear {%
Lasisi%
\ \BBA {} Lawal%
}{%
Lasisi%
\ \BBA {} Lawal%
}{%
{\protect \APACyear {2019}}%
}]{%
lasisi2019preference}
\APACinsertmetastar {%
lasisi2019preference}%
\begin{APACrefauthors}%
Lasisi, T\BPBI J.%
\BCBT {}\ \BBA {} Lawal, F\BPBI B.%
\end{APACrefauthors}%
\unskip\
\newblock
\APACrefYearMonthDay{2019}{}{}.
\newblock
{\BBOQ}\APACrefatitle {Preference of saliva over other body fluids as samples
  for clinical and laboratory investigations among healthcare workers in
  Ibadan, Nigeria} {Preference of saliva over other body fluids as samples for
  clinical and laboratory investigations among healthcare workers in ibadan,
  nigeria}.{\BBCQ}
\newblock
\APACjournalVolNumPages{The Pan African Medical Journal}{34}{}{}.
\PrintBackRefs{\CurrentBib}

\bibitem [\protect \citeauthoryear {%
Lasserre%
\ \protect \BOthers {.}}{%
Lasserre%
\ \protect \BOthers {.}}{%
{\protect \APACyear {2022}}%
}]{%
lasserre2022sars}
\APACinsertmetastar {%
lasserre2022sars}%
\begin{APACrefauthors}%
Lasserre, P.%
, Balansethupathy, B.%
, Vezza, V\BPBI J.%
, Butterworth, A.%
, Macdonald, A.%
, Blair, E\BPBI O.%
, McAteer, L.%
, Hannah, S.%
, Ward, A\BPBI C.%
, Hoskisson, P\BPBI A.%
, Longmuir, A.%
, Setford, S.%
, Farmer, E\BPBI C\BPBI W.%
, Murphy, M\BPBI E.%
, Flynn, H.%
\BCBL {}\ \BBA {} Corrigan, D\BPBI K.%
\end{APACrefauthors}%
\unskip\
\newblock
\APACrefYearMonthDay{2022}{}{}.
\newblock
{\BBOQ}\APACrefatitle {{SARS-CoV-2} Aptasensors Based on Electrochemical
  Impedance Spectroscopy and Low-Cost Gold Electrode Substrates} {{SARS-CoV-2}
  aptasensors based on electrochemical impedance spectroscopy and low-cost gold
  electrode substrates}.{\BBCQ}
\newblock
\APACjournalVolNumPages{Analytical chemistry}{94}{4}{2126--2133}.
\PrintBackRefs{\CurrentBib}

\bibitem [\protect \citeauthoryear {%
Lazari%
\ \protect \BOthers {.}}{%
Lazari%
\ \protect \BOthers {.}}{%
{\protect \APACyear {2022}}%
}]{%
lazari2022maldi}
\APACinsertmetastar {%
lazari2022maldi}%
\begin{APACrefauthors}%
Lazari, L\BPBI C.%
, Zerbinati, R\BPBI M.%
, Rosa-Fernandes, L.%
, Santiago, V\BPBI F.%
, Rosa, K\BPBI F.%
, Angeli, C\BPBI B.%
, Schwab, G.%
, Palmieri, M.%
, Sarmento, D\BPBI J.%
, Marinho, C\BPBI R\BPBI F.%
, Almeida, J\BPBI D.%
, To, K.%
, Giannecchini, S.%
, Wrenger, C.%
, Sabino, E\BPBI C.%
, Martinho, H.%
, Lindoso, J\BPBI A\BPBI L.%
, Durigon, E\BPBI L.%
, Braz-Silva, P\BPBI H.%
\BCBL {}\ \BBA {} Palmisano, G.%
\end{APACrefauthors}%
\unskip\
\newblock
\APACrefYearMonthDay{2022}{}{}.
\newblock
{\BBOQ}\APACrefatitle {{MALDI-TOF} mass spectrometry of saliva samples as a
  prognostic tool for {COVID-19}} {{MALDI-TOF} mass spectrometry of saliva
  samples as a prognostic tool for {COVID-19}}.{\BBCQ}
\newblock
\APACjournalVolNumPages{Journal of Oral Microbiology}{14}{1}{2043651}.
\PrintBackRefs{\CurrentBib}

\bibitem [\protect \citeauthoryear {%
LeCun%
, Bengio%
\BCBL {}\ \BBA {} Hinton%
}{%
LeCun%
\ \protect \BOthers {.}}{%
{\protect \APACyear {2015}}%
}]{%
lecun2015deep}
\APACinsertmetastar {%
lecun2015deep}%
\begin{APACrefauthors}%
LeCun, Y.%
, Bengio, Y.%
\BCBL {}\ \BBA {} Hinton, G.%
\end{APACrefauthors}%
\unskip\
\newblock
\APACrefYearMonthDay{2015}{May}{}.
\newblock
{\BBOQ}\APACrefatitle {Deep learning} {Deep learning}.{\BBCQ}
\newblock
\APACjournalVolNumPages{Nature}{521}{7553}{436--444}.
\PrintBackRefs{\CurrentBib}

\bibitem [\protect \citeauthoryear {%
Leung%
\ \protect \BOthers {.}}{%
Leung%
\ \protect \BOthers {.}}{%
{\protect \APACyear {2021}}%
}]{%
leung2021explainable}
\APACinsertmetastar {%
leung2021explainable}%
\begin{APACrefauthors}%
Leung, C\BPBI K.%
, Fung, D\BPBI L.%
, Mai, D.%
, Wen, Q.%
, Tran, J.%
\BCBL {}\ \BBA {} Souza, J.%
\end{APACrefauthors}%
\unskip\
\newblock
\APACrefYearMonthDay{2021}{}{}.
\newblock
{\BBOQ}\APACrefatitle {Explainable data analytics for disease and healthcare
  informatics} {Explainable data analytics for disease and healthcare
  informatics}.{\BBCQ}
\newblock
\BIn{} \APACrefbtitle {25th International Database Engineering \& Applications
  Symposium} {25th international database engineering \& applications
  symposium}\ (\BPGS\ 65--74).
\PrintBackRefs{\CurrentBib}

\bibitem [\protect \citeauthoryear {%
Li%
, Zeng%
, Wu%
\BCBL {}\ \BBA {} Clawson%
}{%
Li%
\ \protect \BOthers {.}}{%
{\protect \APACyear {2022}}%
}]{%
li2022cov}
\APACinsertmetastar {%
li2022cov}%
\begin{APACrefauthors}%
Li, H.%
, Zeng, N.%
, Wu, P.%
\BCBL {}\ \BBA {} Clawson, K.%
\end{APACrefauthors}%
\unskip\
\newblock
\APACrefYearMonthDay{2022}{}{}.
\newblock
{\BBOQ}\APACrefatitle {Cov-Net: A computer-aided diagnosis method for
  recognizing COVID-19 from chest X-ray images via machine vision} {Cov-net: A
  computer-aided diagnosis method for recognizing covid-19 from chest x-ray
  images via machine vision}.{\BBCQ}
\newblock
\APACjournalVolNumPages{Expert Systems with Applications}{207}{}{118029}.
\PrintBackRefs{\CurrentBib}

\bibitem [\protect \citeauthoryear {%
Liangou%
\ \protect \BOthers {.}}{%
Liangou%
\ \protect \BOthers {.}}{%
{\protect \APACyear {2021}}%
}]{%
liangou2021method}
\APACinsertmetastar {%
liangou2021method}%
\begin{APACrefauthors}%
Liangou, A.%
, Tasoglou, A.%
, Huber, H\BPBI J.%
, Wistrom, C.%
, Brody, K.%
, Menon, P\BPBI G.%
, Bebekoski, T.%
, Menschel, K.%
, Davidson-Fiedler, M.%
, DeMarco, K.%
, Salphale, H.%
, Wistrom, J.%
, Wistrom, S.%
\BCBL {}\ \BBA {} Lee, R\BPBI J.%
\end{APACrefauthors}%
\unskip\
\newblock
\APACrefYearMonthDay{2021}{}{}.
\newblock
{\BBOQ}\APACrefatitle {A method for the identification of COVID-19 biomarkers
  in human breath using Proton Transfer Reaction Time-of-Flight Mass
  Spectrometry} {A method for the identification of covid-19 biomarkers in
  human breath using proton transfer reaction time-of-flight mass
  spectrometry}.{\BBCQ}
\newblock
\APACjournalVolNumPages{EClinicalMedicine}{42}{}{101207}.
\PrintBackRefs{\CurrentBib}

\bibitem [\protect \citeauthoryear {%
Liu%
, Ting%
\BCBL {}\ \BBA {} Zhou%
}{%
Liu%
\ \protect \BOthers {.}}{%
{\protect \APACyear {2008}}%
}]{%
liu2008isolation}
\APACinsertmetastar {%
liu2008isolation}%
\begin{APACrefauthors}%
Liu, F\BPBI T.%
, Ting, K\BPBI M.%
\BCBL {}\ \BBA {} Zhou, Z\BHBI H.%
\end{APACrefauthors}%
\unskip\
\newblock
\APACrefYearMonthDay{2008}{}{}.
\newblock
{\BBOQ}\APACrefatitle {Isolation forest} {Isolation forest}.{\BBCQ}
\newblock
\BIn{} \APACrefbtitle {2008 Eighth IEEE international conference on data
  mining} {2008 eighth ieee international conference on data mining}\ (\BPGS\
  413--422).
\PrintBackRefs{\CurrentBib}

\bibitem [\protect \citeauthoryear {%
Liu%
, Ting%
\BCBL {}\ \BBA {} Zhou%
}{%
Liu%
\ \protect \BOthers {.}}{%
{\protect \APACyear {2012}}%
}]{%
liu2012isolation}
\APACinsertmetastar {%
liu2012isolation}%
\begin{APACrefauthors}%
Liu, F\BPBI T.%
, Ting, K\BPBI M.%
\BCBL {}\ \BBA {} Zhou, Z\BHBI H.%
\end{APACrefauthors}%
\unskip\
\newblock
\APACrefYearMonthDay{2012}{}{}.
\newblock
{\BBOQ}\APACrefatitle {Isolation-based anomaly detection} {Isolation-based
  anomaly detection}.{\BBCQ}
\newblock
\APACjournalVolNumPages{ACM Transactions on Knowledge Discovery from Data
  (TKDD)}{6}{1}{1--39}.
\PrintBackRefs{\CurrentBib}

\bibitem [\protect \citeauthoryear {%
Lundberg%
\ \protect \BOthers {.}}{%
Lundberg%
\ \protect \BOthers {.}}{%
{\protect \APACyear {2020}}%
}]{%
lundberg2020local}
\APACinsertmetastar {%
lundberg2020local}%
\begin{APACrefauthors}%
Lundberg, S\BPBI M.%
, Erion, G.%
, Chen, H.%
, DeGrave, A.%
, Prutkin, J\BPBI M.%
, Nair, B.%
, Katz, R.%
, Himmelfarb, J.%
, Bansal, N.%
\BCBL {}\ \BBA {} Lee, S\BHBI I.%
\end{APACrefauthors}%
\unskip\
\newblock
\APACrefYearMonthDay{2020}{January}{}.
\newblock
{\BBOQ}\APACrefatitle {From local explanations to global understanding with
  explainable {AI} for trees} {From local explanations to global understanding
  with explainable {AI} for trees}.{\BBCQ}
\newblock
\APACjournalVolNumPages{Nature Machine Intelligence}{2}{1}{56--67}.
\PrintBackRefs{\CurrentBib}

\bibitem [\protect \citeauthoryear {%
Lundberg%
\ \BBA {} Lee%
}{%
Lundberg%
\ \BBA {} Lee%
}{%
{\protect \APACyear {2017}}%
}]{%
lundberg2017unified}
\APACinsertmetastar {%
lundberg2017unified}%
\begin{APACrefauthors}%
Lundberg, S\BPBI M.%
\BCBT {}\ \BBA {} Lee, S\BHBI I.%
\end{APACrefauthors}%
\unskip\
\newblock
\APACrefYearMonthDay{2017}{December}{}.
\newblock
{\BBOQ}\APACrefatitle {A unified approach to interpreting model predictions} {A
  unified approach to interpreting model predictions}.{\BBCQ}
\newblock
\APACjournalVolNumPages{Proceedings of the 31st International Conference on
  Neural Information Processing Systems}{}{}{4768–-4777}.
\PrintBackRefs{\CurrentBib}

\bibitem [\protect \citeauthoryear {%
Lundberg%
\ \protect \BOthers {.}}{%
Lundberg%
\ \protect \BOthers {.}}{%
{\protect \APACyear {2018}}%
}]{%
lundberg2018explainable}
\APACinsertmetastar {%
lundberg2018explainable}%
\begin{APACrefauthors}%
Lundberg, S\BPBI M.%
, Nair, B.%
, Vavilala, M\BPBI S.%
, Horibe, M.%
, Eisses, M\BPBI J.%
, Adams, T.%
, Liston, D\BPBI E.%
, Low, D\BPBI K\BHBI W.%
, Newman, S\BHBI F.%
, Kim, J.%
\BCBL {}\ \BBA {} Lee, S\BHBI I.%
\end{APACrefauthors}%
\unskip\
\newblock
\APACrefYearMonthDay{2018}{October}{}.
\newblock
{\BBOQ}\APACrefatitle {Explainable machine-learning predictions for the
  prevention of hypoxaemia during surgery} {Explainable machine-learning
  predictions for the prevention of hypoxaemia during surgery}.{\BBCQ}
\newblock
\APACjournalVolNumPages{Nature Biomedical Engineering}{2}{10}{749--760}.
\PrintBackRefs{\CurrentBib}

\bibitem [\protect \citeauthoryear {%
McKinney%
}{%
McKinney%
}{%
{\protect \APACyear {2011}}%
}]{%
mckinney2011pandas}
\APACinsertmetastar {%
mckinney2011pandas}%
\begin{APACrefauthors}%
McKinney, W.%
\end{APACrefauthors}%
\unskip\
\newblock
\APACrefYearMonthDay{2011}{}{}.
\newblock
{\BBOQ}\APACrefatitle {pandas: a foundational Python library for data analysis
  and statistics} {pandas: a foundational python library for data analysis and
  statistics}.{\BBCQ}
\newblock
\APACjournalVolNumPages{Python for high performance and scientific
  computing}{14}{9}{1--9}.
\PrintBackRefs{\CurrentBib}

\bibitem [\protect \citeauthoryear {%
McLachlan%
}{%
McLachlan%
}{%
{\protect \APACyear {2004}}%
}]{%
mclachlan2004discriminant}
\APACinsertmetastar {%
mclachlan2004discriminant}%
\begin{APACrefauthors}%
McLachlan, G\BPBI J.%
\end{APACrefauthors}%
\unskip\
\newblock
\APACrefYear{2004}.
\newblock
\APACrefbtitle {Discriminant analysis and statistical pattern recognition}
  {Discriminant analysis and statistical pattern recognition}\ (\BVOL~544).
\newblock
\APACaddressPublisher{}{John Wiley and Sons}.
\PrintBackRefs{\CurrentBib}

\bibitem [\protect \citeauthoryear {%
Menon%
, Jiang%
, Vembu%
, Elkan%
\BCBL {}\ \BBA {} Ohno-Machado%
}{%
Menon%
\ \protect \BOthers {.}}{%
{\protect \APACyear {2012}}%
}]{%
menon2012predicting}
\APACinsertmetastar {%
menon2012predicting}%
\begin{APACrefauthors}%
Menon, A\BPBI K.%
, Jiang, X\BPBI J.%
, Vembu, S.%
, Elkan, C.%
\BCBL {}\ \BBA {} Ohno-Machado, L.%
\end{APACrefauthors}%
\unskip\
\newblock
\APACrefYearMonthDay{2012}{}{}.
\newblock
{\BBOQ}\APACrefatitle {Predicting accurate probabilities with a ranking loss}
  {Predicting accurate probabilities with a ranking loss}.{\BBCQ}
\newblock
\APACjournalVolNumPages{Proceedings of the International Conference on Machine
  Learning}{2012}{}{703}.
\PrintBackRefs{\CurrentBib}

\bibitem [\protect \citeauthoryear {%
Mucherino%
, Papajorgji%
\BCBL {}\ \BBA {} Pardalos%
}{%
Mucherino%
\ \protect \BOthers {.}}{%
{\protect \APACyear {2009}}%
}]{%
mucherino2009k}
\APACinsertmetastar {%
mucherino2009k}%
\begin{APACrefauthors}%
Mucherino, A.%
, Papajorgji, P\BPBI J.%
\BCBL {}\ \BBA {} Pardalos, P\BPBI M.%
\end{APACrefauthors}%
\unskip\
\newblock
\APACrefYearMonthDay{2009}{}{}.
\newblock
{\BBOQ}\APACrefatitle {K-nearest neighbor classification} {K-nearest neighbor
  classification}.{\BBCQ}
\newblock
\BIn{} \APACrefbtitle {Data mining in agriculture} {Data mining in
  agriculture}\ (\BPGS\ 83--106).
\newblock
\APACaddressPublisher{}{Springer}.
\PrintBackRefs{\CurrentBib}

\bibitem [\protect \citeauthoryear {%
Nachtigall%
, Pereira%
, Trofymchuk%
\BCBL {}\ \BBA {} Santos%
}{%
Nachtigall%
\ \protect \BOthers {.}}{%
{\protect \APACyear {2020}}%
}]{%
nachtigall2020detection}
\APACinsertmetastar {%
nachtigall2020detection}%
\begin{APACrefauthors}%
Nachtigall, F\BPBI M.%
, Pereira, A.%
, Trofymchuk, O\BPBI S.%
\BCBL {}\ \BBA {} Santos, L\BPBI S.%
\end{APACrefauthors}%
\unskip\
\newblock
\APACrefYearMonthDay{2020}{July}{}.
\newblock
{\BBOQ}\APACrefatitle {Detection of {SARS-CoV-2} in nasal swabs using
  {MALDI-MS}} {Detection of {SARS-CoV-2} in nasal swabs using
  {MALDI-MS}}.{\BBCQ}
\newblock
\APACjournalVolNumPages{Nature Biotechnology}{38}{10}{1168--1173}.
\PrintBackRefs{\CurrentBib}

\bibitem [\protect \citeauthoryear {%
Nadimi-Shahraki%
, Zamani%
\BCBL {}\ \BBA {} Mirjalili%
}{%
Nadimi-Shahraki%
\ \protect \BOthers {.}}{%
{\protect \APACyear {2022}}%
}]{%
nadimi2022enhanced}
\APACinsertmetastar {%
nadimi2022enhanced}%
\begin{APACrefauthors}%
Nadimi-Shahraki, M\BPBI H.%
, Zamani, H.%
\BCBL {}\ \BBA {} Mirjalili, S.%
\end{APACrefauthors}%
\unskip\
\newblock
\APACrefYearMonthDay{2022}{}{}.
\newblock
{\BBOQ}\APACrefatitle {Enhanced whale optimization algorithm for medical
  feature selection: A COVID-19 case study} {Enhanced whale optimization
  algorithm for medical feature selection: A covid-19 case study}.{\BBCQ}
\newblock
\APACjournalVolNumPages{Computers in Biology and Medicine}{148}{}{105858}.
\PrintBackRefs{\CurrentBib}

\bibitem [\protect \citeauthoryear {%
{(NCIRD) National Center for Immunization and Respiratory Diseases , Division
  of Viral Diseases}%
}{%
{(NCIRD) National Center for Immunization and Respiratory Diseases , Division
  of Viral Diseases}%
}{%
{\protect \APACyear {2023}}%
}]{%
cdcVariant}
\APACinsertmetastar {%
cdcVariant}%
\begin{APACrefauthors}%
{(NCIRD) National Center for Immunization and Respiratory Diseases , Division
  of Viral Diseases}.%
\end{APACrefauthors}%
\unskip\
\newblock
\APACrefYearMonthDay{2023}{Accessed: March}{}.
\newblock
\APACrefbtitle {{SARS-CoV-2} Variant Classifications and Definitions.}
  {{SARS-CoV-2} variant classifications and definitions.}
\newblock
\begin{APACrefURL} \url{https://www.cdc.gov/coronavirus/2019-ncov/variants/}
  \end{APACrefURL}
\PrintBackRefs{\CurrentBib}

\bibitem [\protect \citeauthoryear {%
Nguyen%
\ \protect \BOthers {.}}{%
Nguyen%
\ \protect \BOthers {.}}{%
{\protect \APACyear {2022}}%
}]{%
nguyen2022becaked}
\APACinsertmetastar {%
nguyen2022becaked}%
\begin{APACrefauthors}%
Nguyen, D\BPBI Q.%
, Vo, N\BPBI Q.%
, Nguyen, T\BPBI T.%
, Nguyen-An, K.%
, Nguyen, Q\BPBI H.%
, Tran, D\BPBI N.%
\BCBL {}\ \BBA {} Quan, T\BPBI T.%
\end{APACrefauthors}%
\unskip\
\newblock
\APACrefYearMonthDay{2022}{}{}.
\newblock
{\BBOQ}\APACrefatitle {{BeCaked}: An Explainable Artificial Intelligence Model
  for {COVID-19} Forecasting} {{BeCaked}: An explainable artificial
  intelligence model for {COVID-19} forecasting}.{\BBCQ}
\newblock
\APACjournalVolNumPages{Scientific Reports}{12}{1}{1--26}.
\PrintBackRefs{\CurrentBib}

\bibitem [\protect \citeauthoryear {%
Nicola%
\ \protect \BOthers {.}}{%
Nicola%
\ \protect \BOthers {.}}{%
{\protect \APACyear {2020}}%
}]{%
nicola2020health}
\APACinsertmetastar {%
nicola2020health}%
\begin{APACrefauthors}%
Nicola, M.%
, Sohrabi, C.%
, Mathew, G.%
, Kerwan, A.%
, Al-Jabir, A.%
, Griffin, M.%
, Agha, M.%
\BCBL {}\ \BBA {} Agha, R.%
\end{APACrefauthors}%
\unskip\
\newblock
\APACrefYearMonthDay{2020}{September}{}.
\newblock
{\BBOQ}\APACrefatitle {Health policy and leadership models during the
  {COVID-19} pandemic-review article} {Health policy and leadership models
  during the {COVID-19} pandemic-review article}.{\BBCQ}
\newblock
\APACjournalVolNumPages{International Journal of Surgery}{81}{}{122–129}.
\PrintBackRefs{\CurrentBib}

\bibitem [\protect \citeauthoryear {%
Noble%
}{%
Noble%
}{%
{\protect \APACyear {2006}}%
}]{%
noble2006support}
\APACinsertmetastar {%
noble2006support}%
\begin{APACrefauthors}%
Noble, W\BPBI S.%
\end{APACrefauthors}%
\unskip\
\newblock
\APACrefYearMonthDay{2006}{December}{}.
\newblock
{\BBOQ}\APACrefatitle {What is a support vector machine?} {What is a support
  vector machine?}{\BBCQ}
\newblock
\APACjournalVolNumPages{Nature Biotechnology}{24}{12}{1565--1567}.
\PrintBackRefs{\CurrentBib}

\bibitem [\protect \citeauthoryear {%
Ozturk%
\ \protect \BOthers {.}}{%
Ozturk%
\ \protect \BOthers {.}}{%
{\protect \APACyear {2020}}%
}]{%
ozturk2020automated}
\APACinsertmetastar {%
ozturk2020automated}%
\begin{APACrefauthors}%
Ozturk, T.%
, Talo, M.%
, Yildirim, E\BPBI A.%
, Baloglu, U\BPBI B.%
, Yildirim, O.%
\BCBL {}\ \BBA {} Acharya, U\BPBI R.%
\end{APACrefauthors}%
\unskip\
\newblock
\APACrefYearMonthDay{2020}{}{}.
\newblock
{\BBOQ}\APACrefatitle {Automated detection of COVID-19 cases using deep neural
  networks with X-ray images} {Automated detection of covid-19 cases using deep
  neural networks with x-ray images}.{\BBCQ}
\newblock
\APACjournalVolNumPages{Computers in biology and medicine}{121}{}{103792}.
\PrintBackRefs{\CurrentBib}

\bibitem [\protect \citeauthoryear {%
Pedregosa%
\ \protect \BOthers {.}}{%
Pedregosa%
\ \protect \BOthers {.}}{%
{\protect \APACyear {2011}}%
}]{%
pedregosa2011scikit}
\APACinsertmetastar {%
pedregosa2011scikit}%
\begin{APACrefauthors}%
Pedregosa, F.%
, Varoquaux, G.%
, Gramfort, A.%
, Michel, V.%
, Thirion, B.%
, Grisel, O.%
, Blondel, M.%
, Prettenhofer, P.%
, Weiss, R.%
, Dubourg, V.%
, Vanderplas, D.%
, Passos, A.%
, Cournapeau, D.%
, Brucher, M.%
\BCBL {}\ \BBA {} Duchsnay, M\BPBI P\BPBI {\'E}.%
\end{APACrefauthors}%
\unskip\
\newblock
\APACrefYearMonthDay{2011}{}{}.
\newblock
{\BBOQ}\APACrefatitle {Scikit-learn: Machine learning in Python} {Scikit-learn:
  Machine learning in python}.{\BBCQ}
\newblock
\APACjournalVolNumPages{The Journal of Machine Learning
  Research}{12}{}{2825--2830}.
\PrintBackRefs{\CurrentBib}

\bibitem [\protect \citeauthoryear {%
Pennisi%
\ \protect \BOthers {.}}{%
Pennisi%
\ \protect \BOthers {.}}{%
{\protect \APACyear {2021}}%
}]{%
pennisi2021explainable}
\APACinsertmetastar {%
pennisi2021explainable}%
\begin{APACrefauthors}%
Pennisi, M.%
, Kavasidis, I.%
, Spampinato, C.%
, Schinina, V.%
, Palazzo, S.%
, Salanitri, F\BPBI P.%
, Bellitto, G.%
, Rundo, F.%
, Aldinucci, M.%
, Cristofaro, M.%
, Campioni, P.%
, Elisa, P.%
, Stefano, F\BPBI D.%
, Petrone, A.%
, Albarello, F.%
, Ippolito, G.%
, Cuzzocrea, S.%
\BCBL {}\ \BBA {} Conocie, S.%
\end{APACrefauthors}%
\unskip\
\newblock
\APACrefYearMonthDay{2021}{}{}.
\newblock
{\BBOQ}\APACrefatitle {An explainable AI system for automated COVID-19
  assessment and lesion categorization from CT-scans} {An explainable ai system
  for automated covid-19 assessment and lesion categorization from
  ct-scans}.{\BBCQ}
\newblock
\APACjournalVolNumPages{Artificial intelligence in medicine}{118}{}{102114}.
\PrintBackRefs{\CurrentBib}

\bibitem [\protect \citeauthoryear {%
Platt%
}{%
Platt%
}{%
{\protect \APACyear {1999}}%
}]{%
platt1999probabilistic}
\APACinsertmetastar {%
platt1999probabilistic}%
\begin{APACrefauthors}%
Platt, J\BPBI C.%
\end{APACrefauthors}%
\unskip\
\newblock
\APACrefYearMonthDay{1999}{March}{}.
\newblock
{\BBOQ}\APACrefatitle {Probabilistic outputs for support vector machines and
  comparisons to regularized likelihood methods} {Probabilistic outputs for
  support vector machines and comparisons to regularized likelihood
  methods}.{\BBCQ}
\newblock
\APACjournalVolNumPages{Advances in Large Margin Classifiers}{10}{3}{61--74}.
\PrintBackRefs{\CurrentBib}

\bibitem [\protect \citeauthoryear {%
Preian{\`o}%
, Correnti%
, Pelaia%
, Savino%
\BCBL {}\ \BBA {} Terracciano%
}{%
Preian{\`o}%
\ \protect \BOthers {.}}{%
{\protect \APACyear {2021}}%
}]{%
preiano2021maldi}
\APACinsertmetastar {%
preiano2021maldi}%
\begin{APACrefauthors}%
Preian{\`o}, M.%
, Correnti, S.%
, Pelaia, C.%
, Savino, R.%
\BCBL {}\ \BBA {} Terracciano, R.%
\end{APACrefauthors}%
\unskip\
\newblock
\APACrefYearMonthDay{2021}{October}{}.
\newblock
{\BBOQ}\APACrefatitle {{MALDI MS-based investigations for SARS-CoV-2
  detection}} {{MALDI MS-based investigations for SARS-CoV-2
  detection}}.{\BBCQ}
\newblock
\APACjournalVolNumPages{BioChem}{1}{3}{250--278}.
\PrintBackRefs{\CurrentBib}

\bibitem [\protect \citeauthoryear {%
Rahman%
, Hossain%
, Alrajeh%
\BCBL {}\ \BBA {} Gupta%
}{%
Rahman%
\ \protect \BOthers {.}}{%
{\protect \APACyear {2021}}%
}]{%
rahman2021multimodal}
\APACinsertmetastar {%
rahman2021multimodal}%
\begin{APACrefauthors}%
Rahman, M\BPBI A.%
, Hossain, M\BPBI S.%
, Alrajeh, N\BPBI A.%
\BCBL {}\ \BBA {} Gupta, B.%
\end{APACrefauthors}%
\unskip\
\newblock
\APACrefYearMonthDay{2021}{}{}.
\newblock
{\BBOQ}\APACrefatitle {A multimodal, multimedia point-of-care deep learning
  framework for {COVID-19} diagnosis} {A multimodal, multimedia point-of-care
  deep learning framework for {COVID-19} diagnosis}.{\BBCQ}
\newblock
\APACjournalVolNumPages{ACM Transactions on Multimedia Computing Communications
  and Applications}{17}{1s}{1--24}.
\PrintBackRefs{\CurrentBib}

\bibitem [\protect \citeauthoryear {%
Ramesh%
, Dhariwal%
, Nichol%
, Chu%
\BCBL {}\ \BBA {} Chen%
}{%
Ramesh%
\ \protect \BOthers {.}}{%
{\protect \APACyear {2022}}%
}]{%
ramesh2022hierarchical}
\APACinsertmetastar {%
ramesh2022hierarchical}%
\begin{APACrefauthors}%
Ramesh, A.%
, Dhariwal, P.%
, Nichol, A.%
, Chu, C.%
\BCBL {}\ \BBA {} Chen, M.%
\end{APACrefauthors}%
\unskip\
\newblock
\APACrefYearMonthDay{2022}{}{}.
\newblock
{\BBOQ}\APACrefatitle {Hierarchical text-conditional image generation with clip
  latents} {Hierarchical text-conditional image generation with clip
  latents}.{\BBCQ}
\newblock
\APACjournalVolNumPages{arXiv preprint arXiv:2204.06125}{}{}{}.
\PrintBackRefs{\CurrentBib}

\bibitem [\protect \citeauthoryear {%
Rashidi%
, Tran%
\BCBL {}\ \BBA {} Albahra%
}{%
Rashidi%
\ \protect \BOthers {.}}{%
{\protect \APACyear {2022}}%
}]{%
milo}
\APACinsertmetastar {%
milo}%
\begin{APACrefauthors}%
Rashidi, H.%
, Tran, N.%
\BCBL {}\ \BBA {} Albahra, S.%
\end{APACrefauthors}%
\unskip\
\newblock
\APACrefYearMonthDay{2022}{Accessed : March}{}.
\newblock
\APACrefbtitle {Machine Intelligence Learning Optimizer ({MILO}).} {Machine
  intelligence learning optimizer ({MILO}).}
\newblock
\APAChowpublished {"\url{https://milo-ml.com/}"}.
\PrintBackRefs{\CurrentBib}

\bibitem [\protect \citeauthoryear {%
Ribeiro%
, Singh%
\BCBL {}\ \BBA {} Guestrin%
}{%
Ribeiro%
\ \protect \BOthers {.}}{%
{\protect \APACyear {2016}}%
}]{%
ribeiro2016should}
\APACinsertmetastar {%
ribeiro2016should}%
\begin{APACrefauthors}%
Ribeiro, M\BPBI T.%
, Singh, S.%
\BCBL {}\ \BBA {} Guestrin, C.%
\end{APACrefauthors}%
\unskip\
\newblock
\APACrefYearMonthDay{2016}{August}{}.
\newblock
{\BBOQ}\APACrefatitle {"Why Should I Trust You?": Explaining the predictions of
  any classifier} {"why should i trust you?": Explaining the predictions of any
  classifier}.{\BBCQ}
\newblock
\APACjournalVolNumPages{Proceedings of the 22nd ACM SIGKDD International
  Conference on Knowledge Discovery and Data Mining}{}{}{1135-–1144}.
\PrintBackRefs{\CurrentBib}

\bibitem [\protect \citeauthoryear {%
Rish%
}{%
Rish%
}{%
{\protect \APACyear {2001}}%
}]{%
rish2001empirical}
\APACinsertmetastar {%
rish2001empirical}%
\begin{APACrefauthors}%
Rish, I.%
\end{APACrefauthors}%
\unskip\
\newblock
\APACrefYearMonthDay{2001}{}{}.
\newblock
{\BBOQ}\APACrefatitle {An empirical study of the naive Bayes classifier} {An
  empirical study of the naive bayes classifier}.{\BBCQ}
\newblock
\BIn{} \APACrefbtitle {IJCAI 2001 workshop on empirical methods in artificial
  intelligence} {Ijcai 2001 workshop on empirical methods in artificial
  intelligence}\ (\BVOL~3, \BPGS\ 41--46).
\PrintBackRefs{\CurrentBib}

\bibitem [\protect \citeauthoryear {%
Rocca%
\ \protect \BOthers {.}}{%
Rocca%
\ \protect \BOthers {.}}{%
{\protect \APACyear {2020}}%
}]{%
rocca2020combined}
\APACinsertmetastar {%
rocca2020combined}%
\begin{APACrefauthors}%
Rocca, M\BPBI F.%
, Zintgraff, J\BPBI C.%
, Dattero, M\BPBI E.%
, Santos, L\BPBI S.%
, Ledesma, M.%
, Vay, C.%
, Prieto, M.%
, Benedetti, E.%
, Avaro, M.%
, Russo, M.%
, Nachtigall, F\BPBI M.%
\BCBL {}\ \BBA {} Baumeister, E.%
\end{APACrefauthors}%
\unskip\
\newblock
\APACrefYearMonthDay{2020}{December}{}.
\newblock
{\BBOQ}\APACrefatitle {A combined approach of {MALDI-ToF} mass spectrometry and
  multivariate analysis as a potential tool for the detection of {SARS-CoV-2}
  virus in nasopharyngeal swabs} {A combined approach of {MALDI-ToF} mass
  spectrometry and multivariate analysis as a potential tool for the detection
  of {SARS-CoV-2} virus in nasopharyngeal swabs}.{\BBCQ}
\newblock
\APACjournalVolNumPages{Journal of Virological Methods}{286}{}{113991}.
\PrintBackRefs{\CurrentBib}

\bibitem [\protect \citeauthoryear {%
Rostami%
\ \BBA {} Oussalah%
}{%
Rostami%
\ \BBA {} Oussalah%
}{%
{\protect \APACyear {2022}}%
}]{%
rostami2022novel}
\APACinsertmetastar {%
rostami2022novel}%
\begin{APACrefauthors}%
Rostami, M.%
\BCBT {}\ \BBA {} Oussalah, M.%
\end{APACrefauthors}%
\unskip\
\newblock
\APACrefYearMonthDay{2022}{}{}.
\newblock
{\BBOQ}\APACrefatitle {A novel explainable {COVID-19} diagnosis method by
  integration of feature selection with random forest} {A novel explainable
  {COVID-19} diagnosis method by integration of feature selection with random
  forest}.{\BBCQ}
\newblock
\APACjournalVolNumPages{Informatics in Medicine Unlocked}{30}{}{100941}.
\PrintBackRefs{\CurrentBib}

\bibitem [\protect \citeauthoryear {%
Safavian%
\ \BBA {} Landgrebe%
}{%
Safavian%
\ \BBA {} Landgrebe%
}{%
{\protect \APACyear {1991}}%
}]{%
safavian1991survey}
\APACinsertmetastar {%
safavian1991survey}%
\begin{APACrefauthors}%
Safavian, S\BPBI R.%
\BCBT {}\ \BBA {} Landgrebe, D.%
\end{APACrefauthors}%
\unskip\
\newblock
\APACrefYearMonthDay{1991}{May}{}.
\newblock
{\BBOQ}\APACrefatitle {A survey of decision tree classifier methodology} {A
  survey of decision tree classifier methodology}.{\BBCQ}
\newblock
\APACjournalVolNumPages{IEEE Transactions on Systems, Man, and
  Cybernetics}{21}{3}{660--674}.
\PrintBackRefs{\CurrentBib}

\bibitem [\protect \citeauthoryear {%
Salahuddin%
, Woodruff%
, Chatterjee%
\BCBL {}\ \BBA {} Lambin%
}{%
Salahuddin%
\ \protect \BOthers {.}}{%
{\protect \APACyear {2022}}%
}]{%
salahuddin2022transparency}
\APACinsertmetastar {%
salahuddin2022transparency}%
\begin{APACrefauthors}%
Salahuddin, Z.%
, Woodruff, H\BPBI C.%
, Chatterjee, A.%
\BCBL {}\ \BBA {} Lambin, P.%
\end{APACrefauthors}%
\unskip\
\newblock
\APACrefYearMonthDay{2022}{}{}.
\newblock
{\BBOQ}\APACrefatitle {Transparency of deep neural networks for medical image
  analysis: A review of interpretability methods} {Transparency of deep neural
  networks for medical image analysis: A review of interpretability
  methods}.{\BBCQ}
\newblock
\APACjournalVolNumPages{Computers in Biology and Medicine}{140}{}{105111}.
\PrintBackRefs{\CurrentBib}

\bibitem [\protect \citeauthoryear {%
Selvaraju%
\ \protect \BOthers {.}}{%
Selvaraju%
\ \protect \BOthers {.}}{%
{\protect \APACyear {2017}}%
}]{%
selvaraju2017grad}
\APACinsertmetastar {%
selvaraju2017grad}%
\begin{APACrefauthors}%
Selvaraju, R\BPBI R.%
, Cogswell, M.%
, Das, A.%
, Vedantam, R.%
, Parikh, D.%
\BCBL {}\ \BBA {} Batra, D.%
\end{APACrefauthors}%
\unskip\
\newblock
\APACrefYearMonthDay{2017}{}{}.
\newblock
{\BBOQ}\APACrefatitle {Grad-cam: Visual explanations from deep networks via
  gradient-based localization} {Grad-cam: Visual explanations from deep
  networks via gradient-based localization}.{\BBCQ}
\newblock
\BIn{} \APACrefbtitle {Proceedings of the IEEE international conference on
  computer vision} {Proceedings of the ieee international conference on
  computer vision}\ (\BPGS\ 618--626).
\PrintBackRefs{\CurrentBib}

\bibitem [\protect \citeauthoryear {%
Shapley%
}{%
Shapley%
}{%
{\protect \APACyear {1953}}%
}]{%
shapley1953value}
\APACinsertmetastar {%
shapley1953value}%
\begin{APACrefauthors}%
Shapley, L\BPBI S.%
\end{APACrefauthors}%
\unskip\
\newblock
\APACrefYearMonthDay{1953}{March}{}.
\newblock
{\BBOQ}\APACrefatitle {A value for n-person games} {A value for n-person
  games}.{\BBCQ}
\newblock
\APACjournalVolNumPages{Contributions to the Theory of Games
  2.28}{}{28}{307--317}.
\PrintBackRefs{\CurrentBib}

\bibitem [\protect \citeauthoryear {%
Sharma%
\ \protect \BOthers {.}}{%
Sharma%
\ \protect \BOthers {.}}{%
{\protect \APACyear {2022}}%
}]{%
sharma2022segmentation}
\APACinsertmetastar {%
sharma2022segmentation}%
\begin{APACrefauthors}%
Sharma, N.%
, Saba, L.%
, Khanna, N\BPBI N.%
, Kalra, M\BPBI K.%
, Fouda, M\BPBI M.%
\BCBL {}\ \BBA {} Suri, J\BPBI S.%
\end{APACrefauthors}%
\unskip\
\newblock
\APACrefYearMonthDay{2022}{}{}.
\newblock
{\BBOQ}\APACrefatitle {Segmentation-Based Classification Deep Learning Model
  Embedded with Explainable AI for COVID-19 Detection in Chest X-ray Scans}
  {Segmentation-based classification deep learning model embedded with
  explainable ai for covid-19 detection in chest x-ray scans}.{\BBCQ}
\newblock
\APACjournalVolNumPages{Diagnostics}{12}{9}{2132}.
\PrintBackRefs{\CurrentBib}

\bibitem [\protect \citeauthoryear {%
Shimadzu%
}{%
Shimadzu%
}{%
{\protect \APACyear {2022}}%
}]{%
massspec}
\APACinsertmetastar {%
massspec}%
\begin{APACrefauthors}%
Shimadzu.%
\end{APACrefauthors}%
\unskip\
\newblock
\APACrefYearMonthDay{2022}{Accessed: March}{}.
\newblock
\APACrefbtitle {AXIMA Performance - a highly flexible research-grade Mass
  Spectrometer.} {Axima performance - a highly flexible research-grade mass
  spectrometer.}
\newblock
\APAChowpublished
  {\url{https://www.shimadzu.com/an/products/maldi/ms/axima-performance/index.html}}.
\PrintBackRefs{\CurrentBib}

\bibitem [\protect \citeauthoryear {%
Shiri%
\ \protect \BOthers {.}}{%
Shiri%
\ \protect \BOthers {.}}{%
{\protect \APACyear {2022}}%
}]{%
shiri2022impact}
\APACinsertmetastar {%
shiri2022impact}%
\begin{APACrefauthors}%
Shiri, I.%
, Amini, M.%
, Nazari, M.%
, Hajianfar, G.%
, Avval, A\BPBI H.%
, Abdollahi, H.%
, Oveisi, M.%
, Arabi, H.%
, Rahmim, A.%
\BCBL {}\ \BBA {} Zaidi, H.%
\end{APACrefauthors}%
\unskip\
\newblock
\APACrefYearMonthDay{2022}{}{}.
\newblock
{\BBOQ}\APACrefatitle {Impact of feature harmonization on radiogenomics
  analysis: Prediction of EGFR and KRAS mutations from non-small cell lung
  cancer PET/CT images} {Impact of feature harmonization on radiogenomics
  analysis: Prediction of egfr and kras mutations from non-small cell lung
  cancer pet/ct images}.{\BBCQ}
\newblock
\APACjournalVolNumPages{Computers in biology and medicine}{142}{}{105230}.
\PrintBackRefs{\CurrentBib}

\bibitem [\protect \citeauthoryear {%
Sivanesan%
\ \protect \BOthers {.}}{%
Sivanesan%
\ \protect \BOthers {.}}{%
{\protect \APACyear {2022}}%
}]{%
sivanesan2022consolidating}
\APACinsertmetastar {%
sivanesan2022consolidating}%
\begin{APACrefauthors}%
Sivanesan, I.%
, Gopal, J.%
, Vinay, R\BPBI S.%
, Hanna, L\BPBI E.%
, Oh, J\BHBI W.%
\BCBL {}\ \BBA {} Muthu, M.%
\end{APACrefauthors}%
\unskip\
\newblock
\APACrefYearMonthDay{2022}{May}{}.
\newblock
{\BBOQ}\APACrefatitle {Consolidating the potency of Matrix-assisted laser
  desorption/ionization-time of flight mass spectrometry ({MALDI-TOF MS}) in
  viral diagnosis: extrapolating its applicability for {COVID} diagnosis?}
  {Consolidating the potency of matrix-assisted laser
  desorption/ionization-time of flight mass spectrometry ({MALDI-TOF MS}) in
  viral diagnosis: extrapolating its applicability for {COVID}
  diagnosis?}{\BBCQ}
\newblock
\APACjournalVolNumPages{TrAC Trends in Analytical Chemistry}{150}{}{116569}.
\PrintBackRefs{\CurrentBib}

\bibitem [\protect \citeauthoryear {%
Smith%
\ \BBA {} Alvarez%
}{%
Smith%
\ \BBA {} Alvarez%
}{%
{\protect \APACyear {2021}}%
}]{%
smith2021identifying}
\APACinsertmetastar {%
smith2021identifying}%
\begin{APACrefauthors}%
Smith, M.%
\BCBT {}\ \BBA {} Alvarez, F.%
\end{APACrefauthors}%
\unskip\
\newblock
\APACrefYearMonthDay{2021}{}{}.
\newblock
{\BBOQ}\APACrefatitle {Identifying mortality factors from Machine Learning
  using Shapley values--a case of COVID19} {Identifying mortality factors from
  machine learning using shapley values--a case of covid19}.{\BBCQ}
\newblock
\APACjournalVolNumPages{Expert Systems with Applications}{176}{}{114832}.
\PrintBackRefs{\CurrentBib}

\bibitem [\protect \citeauthoryear {%
Sohrabi%
\ \protect \BOthers {.}}{%
Sohrabi%
\ \protect \BOthers {.}}{%
{\protect \APACyear {2020}}%
}]{%
sohrabi2020world}
\APACinsertmetastar {%
sohrabi2020world}%
\begin{APACrefauthors}%
Sohrabi, C.%
, Alsafi, Z.%
, O'neill, N.%
, Khan, M.%
, Kerwan, A.%
, Al-Jabir, A.%
, Iosifidis, C.%
\BCBL {}\ \BBA {} Agha, R.%
\end{APACrefauthors}%
\unskip\
\newblock
\APACrefYearMonthDay{2020}{April}{}.
\newblock
{\BBOQ}\APACrefatitle {World Health Organization declares global emergency: A
  review of the 2019 novel coronavirus ({COVID-19})} {World health organization
  declares global emergency: A review of the 2019 novel coronavirus
  ({COVID-19})}.{\BBCQ}
\newblock
\APACjournalVolNumPages{International Journal of Surgery}{76}{}{71--76}.
\PrintBackRefs{\CurrentBib}

\bibitem [\protect \citeauthoryear {%
Spick%
\ \protect \BOthers {.}}{%
Spick%
\ \protect \BOthers {.}}{%
{\protect \APACyear {2022}}%
}]{%
spick2022systematic}
\APACinsertmetastar {%
spick2022systematic}%
\begin{APACrefauthors}%
Spick, M.%
, Lewis, H\BPBI M.%
, Wilde, M\BPBI J.%
, Hopley, C.%
, Huggett, J.%
\BCBL {}\ \BBA {} Bailey, M\BPBI J.%
\end{APACrefauthors}%
\unskip\
\newblock
\APACrefYearMonthDay{2022}{January}{}.
\newblock
{\BBOQ}\APACrefatitle {Systematic review with meta-analysis of diagnostic test
  accuracy for {COVID-19} by mass spectrometry} {Systematic review with
  meta-analysis of diagnostic test accuracy for {COVID-19} by mass
  spectrometry}.{\BBCQ}
\newblock
\APACjournalVolNumPages{Metabolism}{126}{}{154922}.
\PrintBackRefs{\CurrentBib}

\bibitem [\protect \citeauthoryear {%
St{\aa}hle%
\ \BBA {} Wold%
}{%
St{\aa}hle%
\ \BBA {} Wold%
}{%
{\protect \APACyear {1987}}%
}]{%
staahle1987partial}
\APACinsertmetastar {%
staahle1987partial}%
\begin{APACrefauthors}%
St{\aa}hle, L.%
\BCBT {}\ \BBA {} Wold, S.%
\end{APACrefauthors}%
\unskip\
\newblock
\APACrefYearMonthDay{1987}{July}{}.
\newblock
{\BBOQ}\APACrefatitle {Partial least squares analysis with cross-validation for
  the two-class problem: {A Monte Carlo} study} {Partial least squares analysis
  with cross-validation for the two-class problem: {A Monte Carlo}
  study}.{\BBCQ}
\newblock
\APACjournalVolNumPages{Journal of Chemometrics}{1}{3}{185--196}.
\PrintBackRefs{\CurrentBib}

\bibitem [\protect \citeauthoryear {%
Tallarida%
\ \BBA {} Murray%
}{%
Tallarida%
\ \BBA {} Murray%
}{%
{\protect \APACyear {1987}}%
}]{%
tallarida1987area}
\APACinsertmetastar {%
tallarida1987area}%
\begin{APACrefauthors}%
Tallarida, R\BPBI J.%
\BCBT {}\ \BBA {} Murray, R\BPBI B.%
\end{APACrefauthors}%
\unskip\
\newblock
\APACrefYearMonthDay{1987}{}{}.
\newblock
{\BBOQ}\APACrefatitle {Area under a curve: trapezoidal and Simpson’s rules}
  {Area under a curve: trapezoidal and simpson’s rules}.{\BBCQ}
\newblock
\APACjournalVolNumPages{Manual of Pharmacologic Calculations}{}{}{77--81}.
\PrintBackRefs{\CurrentBib}

\bibitem [\protect \citeauthoryear {%
Thimoteo%
\ \protect \BOthers {.}}{%
Thimoteo%
\ \protect \BOthers {.}}{%
{\protect \APACyear {2022}}%
}]{%
thimoteo2022explainable}
\APACinsertmetastar {%
thimoteo2022explainable}%
\begin{APACrefauthors}%
Thimoteo, L\BPBI M.%
, Vellasco, M\BPBI M.%
, Amaral, J.%
, Figueiredo, K.%
, Yokoyama, C\BPBI L.%
\BCBL {}\ \BBA {} Marques, E.%
\end{APACrefauthors}%
\unskip\
\newblock
\APACrefYearMonthDay{2022}{}{}.
\newblock
{\BBOQ}\APACrefatitle {Explainable artificial intelligence for {COVID-19}
  diagnosis through blood test variables} {Explainable artificial intelligence
  for {COVID-19} diagnosis through blood test variables}.{\BBCQ}
\newblock
\APACjournalVolNumPages{Journal of Control, Automation and Electrical
  Systems}{33}{2}{625--644}.
\PrintBackRefs{\CurrentBib}

\bibitem [\protect \citeauthoryear {%
Tibshirani%
}{%
Tibshirani%
}{%
{\protect \APACyear {1996}}%
}]{%
tibshirani1996regression}
\APACinsertmetastar {%
tibshirani1996regression}%
\begin{APACrefauthors}%
Tibshirani, R.%
\end{APACrefauthors}%
\unskip\
\newblock
\APACrefYearMonthDay{1996}{January}{}.
\newblock
{\BBOQ}\APACrefatitle {Regression shrinkage and selection via the lasso}
  {Regression shrinkage and selection via the lasso}.{\BBCQ}
\newblock
\APACjournalVolNumPages{Journal of the Royal Statistical Society: Series B
  (Methodological)}{58}{1}{267--288}.
\PrintBackRefs{\CurrentBib}

\bibitem [\protect \citeauthoryear {%
Tideman%
\ \protect \BOthers {.}}{%
Tideman%
\ \protect \BOthers {.}}{%
{\protect \APACyear {2021}}%
}]{%
tideman2021automated}
\APACinsertmetastar {%
tideman2021automated}%
\begin{APACrefauthors}%
Tideman, L\BPBI E.%
, Migas, L\BPBI G.%
, Djambazova, K\BPBI V.%
, Patterson, N\BPBI H.%
, Caprioli, R\BPBI M.%
, Spraggins, J\BPBI M.%
\BCBL {}\ \BBA {} Van~de Plas, R.%
\end{APACrefauthors}%
\unskip\
\newblock
\APACrefYearMonthDay{2021}{}{}.
\newblock
{\BBOQ}\APACrefatitle {Automated biomarker candidate discovery in imaging mass
  spectrometry data through spatially localized Shapley additive explanations}
  {Automated biomarker candidate discovery in imaging mass spectrometry data
  through spatially localized shapley additive explanations}.{\BBCQ}
\newblock
\APACjournalVolNumPages{Analytica Chimica Acta}{1177}{}{338522}.
\PrintBackRefs{\CurrentBib}

\bibitem [\protect \citeauthoryear {%
Tran%
\ \protect \BOthers {.}}{%
Tran%
\ \protect \BOthers {.}}{%
{\protect \APACyear {2021}}%
}]{%
tran2021novel}
\APACinsertmetastar {%
tran2021novel}%
\begin{APACrefauthors}%
Tran, N\BPBI K.%
, Howard, T.%
, Walsh, R.%
, Pepper, J.%
, Loegering, J.%
, Phinney, B.%
, Salemi, M\BPBI R.%
\BCBL {}\ \BBA {} Rashidi, H\BPBI H.%
\end{APACrefauthors}%
\unskip\
\newblock
\APACrefYearMonthDay{2021}{April}{}.
\newblock
{\BBOQ}\APACrefatitle {Novel application of automated machine learning with
  {MALDI-ToF-MS} for rapid high-throughput screening of {COVID-19}: a proof of
  concept} {Novel application of automated machine learning with {MALDI-ToF-MS}
  for rapid high-throughput screening of {COVID-19}: a proof of
  concept}.{\BBCQ}
\newblock
\APACjournalVolNumPages{Scientific Reports}{11}{1}{1--10}.
\PrintBackRefs{\CurrentBib}

\bibitem [\protect \citeauthoryear {%
Ullah%
, Moon%
, Naeem%
\BCBL {}\ \BBA {} Jabbar%
}{%
Ullah%
\ \protect \BOthers {.}}{%
{\protect \APACyear {2022}}%
}]{%
ullah2022explainable}
\APACinsertmetastar {%
ullah2022explainable}%
\begin{APACrefauthors}%
Ullah, F.%
, Moon, J.%
, Naeem, H.%
\BCBL {}\ \BBA {} Jabbar, S.%
\end{APACrefauthors}%
\unskip\
\newblock
\APACrefYearMonthDay{2022}{}{}.
\newblock
{\BBOQ}\APACrefatitle {Explainable artificial intelligence approach in
  combating real-time surveillance of COVID19 pandemic from CT scan and X-ray
  images using ensemble model} {Explainable artificial intelligence approach in
  combating real-time surveillance of covid19 pandemic from ct scan and x-ray
  images using ensemble model}.{\BBCQ}
\newblock
\APACjournalVolNumPages{The Journal of Supercomputing}{}{}{1--26}.
\PrintBackRefs{\CurrentBib}

\bibitem [\protect \citeauthoryear {%
Valera%
\ \protect \BOthers {.}}{%
Valera%
\ \protect \BOthers {.}}{%
{\protect \APACyear {2021}}%
}]{%
valera2021covid}
\APACinsertmetastar {%
valera2021covid}%
\begin{APACrefauthors}%
Valera, E.%
, Jankelow, A.%
, Lim, J.%
, Kindratenko, V.%
, Ganguli, A.%
, White, K.%
, Kumar, J.%
\BCBL {}\ \BBA {} Bashir, R.%
\end{APACrefauthors}%
\unskip\
\newblock
\APACrefYearMonthDay{2021}{May}{}.
\newblock
{\BBOQ}\APACrefatitle {{COVID-19} point-of-care diagnostics: present and
  future} {{COVID-19} point-of-care diagnostics: present and future}.{\BBCQ}
\newblock
\APACjournalVolNumPages{ACS Nano}{15}{5}{7899--7906}.
\PrintBackRefs{\CurrentBib}

\bibitem [\protect \citeauthoryear {%
Vinod%
, Jeyavadhanam%
, Zungeru%
\BCBL {}\ \BBA {} Prabaharan%
}{%
Vinod%
\ \protect \BOthers {.}}{%
{\protect \APACyear {2021}}%
}]{%
vinod2021fully}
\APACinsertmetastar {%
vinod2021fully}%
\begin{APACrefauthors}%
Vinod, D\BPBI N.%
, Jeyavadhanam, B\BPBI R.%
, Zungeru, A\BPBI M.%
\BCBL {}\ \BBA {} Prabaharan, S.%
\end{APACrefauthors}%
\unskip\
\newblock
\APACrefYearMonthDay{2021}{}{}.
\newblock
{\BBOQ}\APACrefatitle {Fully automated unified prognosis of Covid-19 chest
  X-ray/CT scan images using Deep Covix-Net model} {Fully automated unified
  prognosis of covid-19 chest x-ray/ct scan images using deep covix-net
  model}.{\BBCQ}
\newblock
\APACjournalVolNumPages{Computers in Biology and Medicine}{136}{}{104729}.
\PrintBackRefs{\CurrentBib}

\bibitem [\protect \citeauthoryear {%
Virtanen%
\ \protect \BOthers {.}}{%
Virtanen%
\ \protect \BOthers {.}}{%
{\protect \APACyear {2020}}%
}]{%
virtanen2020scipy}
\APACinsertmetastar {%
virtanen2020scipy}%
\begin{APACrefauthors}%
Virtanen, P.%
, Gommers, R.%
, Oliphant, T\BPBI E.%
, Haberland, M.%
, Reddy, T.%
, Cournapeau, D.%
, Burovski, E.%
, Peterson, P.%
, Weckesser, W.%
, Bright, J.%
, van~der Walt, S\BPBI J.%
, Brett, M.%
, Wilson, J.%
, Millman, K\BPBI J.%
, Mayorov, N.%
, Nelson, A\BPBI R\BPBI J.%
, Jones, E.%
, Kern, R.%
, Larson, E.%
, Carey, C\BPBI J.%
, Polat, {\.I}.%
, Feng, Y.%
, Moore, E\BPBI W.%
, VanderPlas, J.%
, Laxalde, D.%
, Perktold, J.%
, Cimrman, R.%
, Henriksen, I.%
, Quintero, E\BPBI A.%
, Harris, C\BPBI R.%
, Archibald, A\BPBI M.%
, Ribeiro, A\BPBI H.%
, Pedregosa, F.%
, van Mulbregt, P.%
, Vijaykumar, A.%
, Bardelli, A\BPBI P.%
, Rothberg, A.%
, Hilboll, A.%
, Kloeckner, A.%
, Scopatz, A.%
, Lee, A.%
, Rokem, A.%
, Woods, C\BPBI N.%
, Fulton, C.%
, Masson, C.%
, H{\"a}ggstr{\"o}m, C.%
, Fitzgerald, C.%
, Nicholson, D\BPBI A.%
, Hagen, D\BPBI R.%
, Pasechnik, D\BPBI V.%
, Olivetti, E.%
, Martin, E.%
, Wieser, E.%
, Silva, F.%
, Lenders, F.%
, Wilhelm, F.%
, Young, G.%
, Price, G\BPBI A.%
, Ingold, G\BHBI L.%
, Allen, G\BPBI E.%
, Lee, G\BPBI R.%
, Audren, H.%
, Probst, I.%
, Dietrich, J\BPBI P.%
, Silterra, J.%
, Webber, J\BPBI T.%
, Slavi{\/vc}, J.%
, Nothman, J.%
, Buchner, J.%
, Kulick, J.%
, Sch{\"o}nberger, J\BPBI L.%
, de Miranda~Cardoso, J\BPBI V.%
, Reimer, J.%
, Harrington, J.%
, Rodr{\'i}guez, J\BPBI L\BPBI C.%
, Nunez-Iglesias, J.%
, Kuczynski, J.%
, Tritz, K.%
, Thoma, M.%
, Newville, M.%
, K{\"u}mmerer, M.%
, Bolingbroke, M.%
, Tartre, M.%
, Pak, M.%
, Smith, N\BPBI J.%
, Nowaczyk, N.%
, Shebanov, N.%
, Pavlyk, O.%
, Brodtkorb, P\BPBI A.%
, Lee, P.%
, McGibbon, R\BPBI T.%
, Feldbauer, R.%
, Lewis, S.%
, Tygier, S.%
, Sievert, S.%
, Vigna, S.%
, Peterson, S.%
, More, S.%
, Pudlik, T.%
, Oshima, T.%
, Pingel, T\BPBI J.%
, Robitaille, T\BPBI P.%
, Spura, T.%
, Jones, T\BPBI R.%
, Cera, T.%
, Leslie, T.%
, Zito, T.%
, Krauss, T.%
, Upadhyay, U.%
, Halchenko, Y\BPBI O.%
, V{\'a}zquez-Baeza, Y.%
\BCBL {}\ \BBA {} Contributors, S\BPBI .%
\end{APACrefauthors}%
\unskip\
\newblock
\APACrefYearMonthDay{2020}{}{}.
\newblock
{\BBOQ}\APACrefatitle {{SciPy} 1.0: fundamental algorithms for scientific
  computing in {Python}} {{SciPy} 1.0: fundamental algorithms for scientific
  computing in {Python}}.{\BBCQ}
\newblock
\APACjournalVolNumPages{Nature methods}{17}{3}{261--272}.
\PrintBackRefs{\CurrentBib}

\bibitem [\protect \citeauthoryear {%
Wang%
\ \protect \BOthers {.}}{%
Wang%
\ \protect \BOthers {.}}{%
{\protect \APACyear {2020}}%
}]{%
wang2020score}
\APACinsertmetastar {%
wang2020score}%
\begin{APACrefauthors}%
Wang, H.%
, Wang, Z.%
, Du, M.%
, Yang, F.%
, Zhang, Z.%
, Ding, S.%
, Mardziel, P.%
\BCBL {}\ \BBA {} Hu, X.%
\end{APACrefauthors}%
\unskip\
\newblock
\APACrefYearMonthDay{2020}{}{}.
\newblock
{\BBOQ}\APACrefatitle {Score-CAM: Score-weighted visual explanations for
  convolutional neural networks} {Score-cam: Score-weighted visual explanations
  for convolutional neural networks}.{\BBCQ}
\newblock
\BIn{} \APACrefbtitle {Proceedings of the IEEE/CVF conference on computer
  vision and pattern recognition workshops} {Proceedings of the ieee/cvf
  conference on computer vision and pattern recognition workshops}\ (\BPGS\
  24--25).
\PrintBackRefs{\CurrentBib}

\bibitem [\protect \citeauthoryear {%
Wen%
\ \protect \BOthers {.}}{%
Wen%
\ \protect \BOthers {.}}{%
{\protect \APACyear {2022}}%
}]{%
wen2022acsn}
\APACinsertmetastar {%
wen2022acsn}%
\begin{APACrefauthors}%
Wen, C.%
, Liu, S.%
, Liu, S.%
, Heidari, A\BPBI A.%
, Hijji, M.%
, Zarco, C.%
\BCBL {}\ \BBA {} Muhammad, K.%
\end{APACrefauthors}%
\unskip\
\newblock
\APACrefYearMonthDay{2022}{}{}.
\newblock
{\BBOQ}\APACrefatitle {{ACSN: Attention capsule sampling network for diagnosing
  COVID-19 based on chest CT scans}} {{ACSN: Attention capsule sampling network
  for diagnosing COVID-19 based on chest CT scans}}.{\BBCQ}
\newblock
\APACjournalVolNumPages{Computers in Biology and Medicine}{}{}{106338}.
\PrintBackRefs{\CurrentBib}

\bibitem [\protect \citeauthoryear {%
{(WHO) World Health Organization }%
}{%
{(WHO) World Health Organization }%
}{%
{\protect \APACyear {2023}}%
}]{%
whodash}
\APACinsertmetastar {%
whodash}%
\begin{APACrefauthors}%
{(WHO) World Health Organization }.%
\end{APACrefauthors}%
\unskip\
\newblock
\APACrefYearMonthDay{2023}{Accessed: March}{}.
\newblock
\APACrefbtitle {{WHO COVID-19 Dashboard}.} {{WHO COVID-19 Dashboard}.}
\newblock
\begin{APACrefURL} \url{https://covid19.who.int/} \end{APACrefURL}
\PrintBackRefs{\CurrentBib}

\bibitem [\protect \citeauthoryear {%
Wold%
, Esbensen%
\BCBL {}\ \BBA {} Geladi%
}{%
Wold%
\ \protect \BOthers {.}}{%
{\protect \APACyear {1987}}%
}]{%
wold1987principal}
\APACinsertmetastar {%
wold1987principal}%
\begin{APACrefauthors}%
Wold, S.%
, Esbensen, K.%
\BCBL {}\ \BBA {} Geladi, P.%
\end{APACrefauthors}%
\unskip\
\newblock
\APACrefYearMonthDay{1987}{August}{}.
\newblock
{\BBOQ}\APACrefatitle {Principal component analysis} {Principal component
  analysis}.{\BBCQ}
\newblock
\APACjournalVolNumPages{Chemometrics and Intelligent Laboratory
  Systems}{2}{1--3}{37--52}.
\PrintBackRefs{\CurrentBib}

\bibitem [\protect \citeauthoryear {%
Xie%
, Castro%
, Bell%
, Rubakhin%
\BCBL {}\ \BBA {} Sweedler%
}{%
Xie%
\ \protect \BOthers {.}}{%
{\protect \APACyear {2020}}%
}]{%
xie2020single}
\APACinsertmetastar {%
xie2020single}%
\begin{APACrefauthors}%
Xie, Y\BPBI R.%
, Castro, D\BPBI C.%
, Bell, S\BPBI E.%
, Rubakhin, S\BPBI S.%
\BCBL {}\ \BBA {} Sweedler, J\BPBI V.%
\end{APACrefauthors}%
\unskip\
\newblock
\APACrefYearMonthDay{2020}{}{}.
\newblock
{\BBOQ}\APACrefatitle {Single-cell classification using mass spectrometry
  through interpretable machine learning} {Single-cell classification using
  mass spectrometry through interpretable machine learning}.{\BBCQ}
\newblock
\APACjournalVolNumPages{Analytical chemistry}{92}{13}{9338--9347}.
\PrintBackRefs{\CurrentBib}

\bibitem [\protect \citeauthoryear {%
Xu%
\ \protect \BOthers {.}}{%
Xu%
\ \protect \BOthers {.}}{%
{\protect \APACyear {2023}}%
}]{%
xu2023improving}
\APACinsertmetastar {%
xu2023improving}%
\begin{APACrefauthors}%
Xu, Y.%
, Lam, H\BHBI K.%
, Jia, G.%
, Jiang, J.%
, Liao, J.%
\BCBL {}\ \BBA {} Bao, X.%
\end{APACrefauthors}%
\unskip\
\newblock
\APACrefYearMonthDay{2023}{}{}.
\newblock
{\BBOQ}\APACrefatitle {Improving COVID-19 CT classification of CNNs by learning
  parameter-efficient representation} {Improving covid-19 ct classification of
  cnns by learning parameter-efficient representation}.{\BBCQ}
\newblock
\APACjournalVolNumPages{Computers in Biology and Medicine}{152}{}{106417}.
\PrintBackRefs{\CurrentBib}

\bibitem [\protect \citeauthoryear {%
Yan%
\ \protect \BOthers {.}}{%
Yan%
\ \protect \BOthers {.}}{%
{\protect \APACyear {2021}}%
}]{%
yan2021rapid}
\APACinsertmetastar {%
yan2021rapid}%
\begin{APACrefauthors}%
Yan, L.%
, Yi, J.%
, Huang, C.%
, Zhang, J.%
, Fu, S.%
, Li, Z.%
, Lyu, Q.%
, Xu, Y.%
, Wang, K.%
, Yang, H.%
, Ma, Q.%
, Cui, X.%
, Qiao, L.%
, Sun, W.%
\BCBL {}\ \BBA {} Liao, P.%
\end{APACrefauthors}%
\unskip\
\newblock
\APACrefYearMonthDay{2021}{}{}.
\newblock
{\BBOQ}\APACrefatitle {Rapid detection of {COVID-19} using {MALDI-TOF-based}
  serum peptidome profiling} {Rapid detection of {COVID-19} using
  {MALDI-TOF-based} serum peptidome profiling}.{\BBCQ}
\newblock
\APACjournalVolNumPages{Analytical chemistry}{93}{11}{4782--4787}.
\PrintBackRefs{\CurrentBib}

\bibitem [\protect \citeauthoryear {%
Ye%
, Xia%
\BCBL {}\ \BBA {} Yang%
}{%
Ye%
\ \protect \BOthers {.}}{%
{\protect \APACyear {2021}}%
}]{%
ye2021explainable}
\APACinsertmetastar {%
ye2021explainable}%
\begin{APACrefauthors}%
Ye, Q.%
, Xia, J.%
\BCBL {}\ \BBA {} Yang, G.%
\end{APACrefauthors}%
\unskip\
\newblock
\APACrefYearMonthDay{2021}{}{}.
\newblock
{\BBOQ}\APACrefatitle {Explainable {AI for COVID-19 CT} classifiers: An initial
  comparison study} {Explainable {AI for COVID-19 CT} classifiers: An initial
  comparison study}.{\BBCQ}
\newblock
\BIn{} \APACrefbtitle {2021 IEEE 34th International Symposium on Computer-Based
  Medical Systems (CBMS)} {2021 ieee 34th international symposium on
  computer-based medical systems (cbms)}\ (\BPGS\ 521--526).
\PrintBackRefs{\CurrentBib}

\end{thebibliography}
\end{document}